\documentclass[lettersize,journal]{IEEEtran}
\usepackage{amsmath,amsfonts}
\usepackage{algorithmic}
\usepackage{algorithm}
\usepackage{booktabs}
\usepackage{hyperref}
\hypersetup{
    colorlinks=true,
    citecolor=blue,
    linkcolor=blue,
    urlcolor=black,
}
\usepackage{cleveref}
\usepackage{enumitem}

\usepackage{array}
\usepackage[caption=false,font=normalsize,labelfont=sf,textfont=sf]{subfig}
\usepackage{textcomp}
\usepackage{stfloats}
\usepackage{url}
\usepackage[most]{tcolorbox}
\usepackage{verbatim}
\usepackage{graphicx}
\usepackage{wrapfig}
\usepackage{comment}
\usepackage{amsthm}
\usepackage{listings}
\usepackage{cite}
\usepackage{amssymb}
\newtheorem{theorem}{Theorem}
\newtheorem{lemma}{Lemma}
\usepackage{cleveref}
\usepackage{orcidlink}

\usepackage[table,xcdraw]{xcolor}
\usepackage{siunitx}

\definecolor{winGreen}{HTML}{addd8e} %{E8F5E9} 
\newcommand{\win}[1]{\cellcolor{winGreen}\textbf{#1}}
\usepackage{multirow}
\usepackage{subcaption}
\usepackage{float}
\definecolor{promptbg}{RGB}{235,245,255}      % light blue background
\definecolor{promptframe}{RGB}{120,170,220}   % blue frame
\definecolor{promptkw}{RGB}{120,60,180}       % violet keywords
\definecolor{promptemph}{RGB}{30,90,160}      % dark blue emphasis
\usepackage{tikz}
\usetikzlibrary{matrix}
\usetikzlibrary{positioning, arrows.meta}
\usepackage{makecell}
\usepackage{pgfplots}
\pgfplotsset{compat=newest}

\lstdefinestyle{promptstyle}{
  basicstyle=\ttfamily\small,
  backgroundcolor=\color{promptbg},
  frame=single,
  rulecolor=\color{promptframe},
  framesep=6pt,
  xleftmargin=4pt,
  xrightmargin=4pt,
  breaklines=true,
  breakatwhitespace=false,
  columns=fullflexible,
  keepspaces=true,
  showstringspaces=false,
  upquote=true,
  numbers=none,
  keywordstyle=\bfseries\color{promptkw},
  emphstyle=\bfseries\color{promptkw}
}

\crefname{lstlisting}{Listing}{Listings}
\newcommand{\bem}[1]{{\bf\em #1}}

\newcommand{\rreva}[2][]{#1\frombody{red}{RR}{#2}}

\newtheorem{mydef}{\bf Definition}
\crefname{mydef}{Definition}{Definitions}
\Crefname{mydef}{Definition}{Definitions}

\newcommand{\frombody}[3]{
\noindent
\textcolor{#1}{
{$\bf [\!\![\!\![$}\underline{{\bf #2}}
{\bf :}
{\em #3}{$\bf ]\!\!]\!\!]$}}
{}}

\newcommand{\eg}{\mbox{{\em e.g.}}}
\newcommand{\ie}{\mbox{{\em i.e.}}}
\newcommand{\cf}{\mbox{{c.f.}}}

\newcommand{\pname}{\mbox{{\scshape GraphEVAL}}} %GraphEVAL
\newcommand{\StQ}[1]{{\textcolor{purple}{\bf StQ: #1}}} %Stefano's note 

\newcolumntype{C}[1]{>{\centering\arraybackslash}p{#1}}
\newcolumntype{L}[1]{>{\arraybackslash}p{#1}}

\begin{document}

%\title{Formatting Instructions for COLM 2026 \\ Conference Submissions}

%\title{Is the LLM Brain More Than a Pattern Matcher? Probing for Coherence, Diversity, and Robustness}

%\title{Can We Trust LLM's \StQ{LLMs'} %Logic %\StQ{
%Reasoning? %}? 
%Quantifying Uncertainty, Coherence, and Robustness via a Graph-Based Framework
%\StQ{Shorter title?: Trusting LLM Reasoning: A Graph-Based View of Uncertainty and Robustness}}
%\title{Trusting LLM Reasoning: A Graph-Based View of Uncertainty and Robustness}
%\title{$\pname$: Estimating Reasoning Uncertainty to Improve LLM Faithfulness}
\title{$\pname$: Quantifying Reasoning Uncertainty to
Assess Trustworthiness in LLMs}

%% Some Title Suggestions:
% 1. Is the "LLM Brain" More Than a Pattern Matcher? Probing for Coherence, Diversity, and Robustness
% 2. Can We Trust the Logic of Large Language Models? Dissecting the Coherence of Reasoning Patterns
% 3. Does the "LLM Brain" Scale or Stumble? Testing the Robustness and Diversity of Machine Logic
% 4. Can LLMs Maintain Logical Integrity? Dissecting Diversity and Coherence in Reasoning Patterns
% 5. How Reliable is the "LLM Brain"? A Systematic Audit of Reasoning Coherence and Robustness
% 6. Is LLM Reasoning Truly Coherent? Dissecting Patterns of Diversity and Robustness

\author{IEEE Publication Technology,~\IEEEmembership{Staff,~IEEE,}
        % <-this % stops a space
\thanks{This paper was produced by the IEEE Publication Technology Group. They are in Piscataway, NJ.}% <-this % stops a space
\thanks{Manuscript received April 19, 2021; revised August 16, 2021.}}

\author{\IEEEauthorblockN{
Riccardo Revalor\IEEEauthorrefmark{1}\orcidlink{0009-0004-0120-7852},~\IEEEmembership{Graduate Student Member,~IEEE}, 
Stefano Quer\IEEEauthorrefmark{2}\orcidlink{0000-0001-6835-8277},~\IEEEmembership{Senior Member,~IEEE},\\
Jalees Rehman\IEEEauthorrefmark{3}\orcidlink{0000-0002-2787-9292},%,~\IEEEmembership{}, 
and
Debjit Pal\IEEEauthorrefmark{4}\orcidlink{0000-0003-3722-5126}, ~\IEEEmembership{Member,~IEEE}}
% \IEEEauthorblockA{\\
% Email: \{rreva\IEEEauthorrefmark{1},
% jalees\IEEEauthorrefmark{3},
% dpal2\IEEEauthorrefmark{4}\}@uic.edu, stefano.quer@polito.it\IEEEauthorrefmark{2}
%}
\thanks{Riccardo Revalor and Debjit Pal are with the Department of Electrical and Computer Engineering, 
  University of Illinois Chicago, US (E-Mail: \{rreva, dpal2\}@uic.edu).}\thanks{Jalees Rehman is with the Department of Biochemistry and Molecular Genetics, 
  University of Illinois Chicago, US (E-Mail: \{jalees\}@uic.edu).}\thanks{Stefano Quer is with the Department of Control and Computer Engineering, 
  Politecnico di Torino, Italy (E-Mail: \{stefano.quer\}@polito.it)}
}

%Dept. of Control and Computer Engineering

\markboth{Journal of \LaTeX\ Class Files,~Vol.~14, No.~8, August~2026}%
{Shell \MakeLowercase{\textit{et al.}}: A Sample Article Using IEEEtran.cls for IEEE Journals}

\IEEEpubid{0000--0000/00\$00.00~\copyright~2026 IEEE}


\maketitle
\IEEEpubidadjcol

\begin{abstract}
%\rreva{rewrite}
%The abstract paragraph should be indented 1/2~inch (3~picas) on both left and right-hand margins. Use 10~point type, with a vertical spacing of 11~points. The word \textit{Abstract} must be centered and in point size 12. Two line spaces precede the abstract. The abstract must be limited to one paragraph. In recent years, Large Language Models (LLMs) have been widely adopted because of their ability to generalize their reasoning to solve multiple tasks.
%\StQ{Large-Language Models' %(LLMs) %often produce flawed or unfaithful reasoning, which tendency to produce unfaithful reasoning limits their trustworthiness. 
Unfaithful reasoning undermines the trustworthiness of Large-Language Models. In mission-critical settings where reasoning errors can lead to catastrophic physical, financial, or legal consequences, accuracy without faithfulness is insufficient. Decoding methods such as Self-Consistency (SC) may fail to detect these failures because they evaluate final-answer agreement while ignoring the reasoning that produces it.
%Large-Language Models (LLMs) can be prone to flawed and unfaithful reasoning that decoding strategies like Self-Consistency (SC) fail to detect, as they evaluate only final-answer agreement while ignoring the logical validity of intermediate steps.
%This raises three fundamental questions: (1) How can we reliably quantify uncertainty in LLM reasoning across tasks of increasing complexity? (2) Can semantic, structural, and causal awareness select more faithful reasoning paths than na\"ive majority voting? (3) How robust is reasoning topology under adversarial conditions? 
We therefore investigate %ask 
whether reasoning uncertainty can be reliably quantified, used to select more faithful answers, and evaluated under adversarial conditions.
%a graph-based reasoning framework that reframes uncertainty quantification (UQ) as a %problem of {\em holistic reasoning fidelity} problem. shifting the evaluation paradigm from final-answer consensus to %\emph{holistic reasoning fidelity}. the semantic and causal consistency of the reasoning paths that produces it.
To address these questions, we introduce $\pname$, a framework that estimates reasoning uncertainty to assess models' trustworthiness and shift evaluation from final-answer consensus to the semantic and causal consistency of the reasoning that produces it.
We propose a novel uncertainty quantification metric based on the Graph Reasoning Coherence Score (GRCS), %which assigns higher
where lower coherence reflects
higher reasoning uncertainty. %uncertainty to less faithful reasoning 
The resulting uncertainty increases as
reasoning faithfulness decreases in $\mathbf{88.9\%}$ of evaluated configurations and outperforms SOTA baselines in $\mathbf{77.8\%}$ of comparisons, while being, on average, $32.5\times$ faster.
%a novel UQ metric 
%that quantifies semantic-structural consensus of reasoning and captures partial mode collapse. %Among the evaluated metrics, GRCS is the only one that remains negatively correlated with reasoning faithfulness in most configurations across both more capable and smaller models.
%to measure semantic and structural agreement among reasoning paths.
%Additionally, we introduce Graph Self-Consistency (GSC), a medoid-based decoding strategy that occasionally trades nominal accuracy for higher reasoning fidelity in smaller models, thereby exposing the extent to which SC is inflated by unfaithful lucky guesses.
%For more capable models, GSC preserves or improves accuracy while increasing reasoning faithfulness.
For decoding, we introduce Graph Self Consistency (GSC), which accounts for within-answer reasoning uncertainty, ignored by SC, and selects the more %faithful 
logically sound answer in $\mathbf{24.9\%}$ more SC–GSC disagreements.
%which measures within-answer reasoning uncertainty ignored by SC to select the model's most faithful reasoning path. GSC %typically 
%selects the most faithful answers in $\approx \mathbf{25\%}$ more cases than SC. % than SC.
% StQ: Additionally, we introduce Graph Self-Consistency (GSC), a medoid-based decoding strategy that occasionally trade nominal accuracy for higher reasoning fidelity in smaller models, thereby exposing the extent to which SC is inflated by unfaithful lucky guesses, while preserving or improving accuracy in more capable ones.
%Finally, through adversarial ablation, we show that removing the GSC-selected paths %acts as a ``load-bearing path'' and forcing models away from it degrades reasoning faithfulness and, in targeted cases, causes drops in accuracy.
%degrades faithfulness in $\mathbf{11\text{--}55\%}$ of answers, with faithfulness even dropping more than $\mathbf{50\%}$ in specific settings.
Finally, under adversarial ablation, we show that removing GSC-selected paths degrades faithfulness in up to ${\bf 55}\%$ of answers (average: ${\bf 30.59}\%$), with drops exceeding $\mathbf{50\%}$ in some settings.
\end{abstract}


\section{Introduction}
\label{sec:intro}

%\StQ{Some numbers are in math mode; others are not. I tried to unify them, but the problem still exists. Which direction do you want to go? Math ot no-math mode?}
%\input{src/tables/datasets}
\IEEEPARstart{I}{n} recent years, Large-Language Models (LLMs) %Deep Learning (DL) architectures 
have achieved remarkable success across widely different domains %through a single model, 
%often generalizing to new tasks in zero-shot and few-shot settings
~\cite{llms_survey}. Nevertheless, their trustworthiness is limited by their tendency to produce %{\em 
plausible but %entirely 
fabricated reasoning, colloquially known as {\em hallucinations}, often doing so with high statistical confidence~\cite{ji2023survey}. %\cite{lecun2015deep}. 
%Such a lack of trustworthiness %has severely restricted their deployment in %high-stakes 
%limits their use in mission-critical environments~\cite{huang2020survey}. % is severely restricted %heavily bottlenecked 
%by fundamental trustworthiness flaws, such as lack of interpretability, susceptibility to adversarial perturbations, and uncalibrated uncertainty~\cite{huang2020survey}. 
%Large-Language Models (LLMs) %are considered the state-of-the-art (SOTA) in natural language processing, and 
%have been widely adopted because of their ability to perform a broad spectrum of %very different 
%tasks through a single model, often generalizing to new tasks in zero-shot and few-shot settings~\cite{brown2020languagemodels, zhao2023surveyllms}. 
%Nevertheless, their trustworthiness is limited by their tendency to produce {\em plausible but entirely fabricated reasoning}, colloquially known as {\em hallucinations}, often doing so with high statistical confidence~\cite{ji2023survey}. 
%To increase LLM performance, Chain-of-Thought (CoT)~\cite{wei2022chainofthought} %prompting was introduced~\cite{wei2022chainofthought}, with the goal of enabling 
%enabled LLM %them 
%to decompose complex problems into intermediate steps, and produce explicit, step-by-step %reasoning~\cite{kahneman2011thinking, ziabari2025reasoning, devarda2025cost}. %In this context, 
In certain cases, hallucinations %actively 
contribute to the ``lucky guess,'' %phenomenon, 
in which a LLM may follow %produce 
flawed, illogical, or even contradictory reasoning processes, yet yield %that somehow happen to output 
the correct answer~\cite{bachmann2026potential}. While the answer may be correct, the unreliability of this process %fundamentally 
threatens the model's trustworthiness and robustness. In \emph{mission-critical} settings where reasoning errors can lead to catastrophic %severe 
physical~\cite{Shen2023ChatGPTDoubleEdged}, financial~\cite{Chen2025TortLiabilityAI}, or legal~\cite{Linna2026ChallengesLegalReasoning} consequences, %harm, 
%empirical performance alone is %not 
accuracy without faithfulness is
insufficient~\cite{Morey2025EvaluationRequirements}.
%\IEEEpubidadjcol

This %limitation 
problem has motivated methods that expose the reasoning behind %an 
LLMs' answers. Among them, Chain-of-Thought (CoT) prompting elicits explicit intermediate steps for solving complex problems~\cite{wei2022chainofthought}. Once these steps are exposed, sequence-level decoding strategies,
%Decoding strategies that mimic human-like reasoning, 
%\eg, 
like Self Consistency (SC)~\cite{wang2023selfconsistency}, %\StQ{You have already may references, but this is one potentially missing: 
%and Self-Verification~\cite{weng2023selfverification} %} 
%\rreva{The reference is good, but don't you think reviewers will ask "why didn't you compare with Self Verification as well?" if we put it into the picture?} 
%have been widely used to sample LLM answers. 
can %compare 
sample alternative reasoning paths. SC %samples multiple 
%compares the reasoning paths %and 
%samples different CoTs and 
selects
%by selecting 
the most frequent final answer among multiple sampled CoTs ($\ie$, via majority voting), assuming that answer agreement ultimately leads to correctness~\cite{lyu2023faithful}.
%However, SC applies \bem{majority voting} to the final output, marginalizing the different CoTs, making it highly vulnerable to unfaithful reasoning%\StQ{You have already may references, but this is one potentially missing:
 %}.
However, because it ignores the reasoning behind each answer, SC cannot guarantee the trustworthiness of the reasoning.
%In certain cases, hallucinations actively contribute to the ``lucky guess,'' %phenomenon, 
%in which a LLM may follow %produce 
%flawed, illogical, or even contradictory reasoning processes, yet yield %that somehow happen to output 
%the correct answer~\cite{bachmann2026potential}. While the answer may be correct, the unreliability of this process %fundamentally 
%threatens the model's trustworthiness and robustness. In \bem{mission-critical} settings where reasoning errors can lead to catastrophic %severe 
%physical~\cite{Shen2023ChatGPTDoubleEdged}, financial~\cite{Chen2025TortLiabilityAI}, or legal~\cite{Linna2026ChallengesLegalReasoning} consequences, %harm, 
%empirical performance alone is %not 
%accuracy without faithfulness is
%insufficient~\cite{Morey2025EvaluationRequirements}. 

Therefore, we hypothesize that assessing and improving LLMs' trustworthiness requires %quantifying the coherence and uncertainty of their underlying reasoning paths.
looking beyond answer agreement to quantify
how consistently reasoning paths agree with one another and
how uncertain the underlying reasoning is.
Recent work, such as Topo-UQ~\cite{da2025understanding}, has 
advanced the field by representing each CoT as a graph of its reasoning steps and their connections, and measuring differences across these graphs to estimate uncertainty.
However, Topo-UQ does not generalize to tasks of higher reasoning complexity and weakens in the presence of mode collapse, 
%Mode collapse %observed 
%in compact architectures, 
where identical flawed reasonings % shortcuts 
are confidently repeated%, remains especially problematic
~\cite{ding2024breakchain,lin2025implicitreasoning}.
A complementary limitation affects CoTA, which scores pairwise consistency among reasoning steps~\cite{tanneru2024quantifying}, but cannot detect globally unfaithful reasoning caused by invalid causal dependencies. Thus, %existing 
these methods capture only partial aspects of reasoning and do not jointly evaluate its semantic and causal consistency.
\IEEEpubidadjcol

To address these challenges, %solve these limitations, 
we propose a graph-based reasoning framework, $\pname$, %a framework that shifts 
to shift the evaluation paradigm from final-answer consensus to %\emph{holistic reasoning fidelity}. 
the semantic and causal consistency of the reasoning %paths 
that produces it. Achieving this vision requires exploration and %an
effective solutions to the following three research questions (RQs): \textbf{RQ1:} How can we reliably quantify uncertainty in LLM reasoning processes across diverse tasks of %varying 
increasing %cognitive 
reasoning complexity? 
%\item 
\textbf{RQ2:} %To what extent does topological Medoid selection (Graph-SC) enhance reasoning accuracy compared to standard semantic majority voting, and what vulnerabilities in standard Self-Consistency does this structural awareness expose?
Does structural and causal awareness improve upon na\"ive majority voting to %and can it %GSC 
systematically filter out unfaithful %``lucky guesses'' 
reasoning that may artificially inflate %standard 
SC accuracy? 
%\item 
\textbf{RQ3:} How robust is the reasoning topology under adversarial conditions, and how do %models 
LLMs of varying %representational 
reasoning capacities respond when their dominant reasoning paths are adversarially ablated? 

To answer these RQs, we introduce $\pname$ and evaluate it on a broad spectrum of progressively complex tasks. %Specifically, 
Our key contributions are:%}
%To answer these RQs, we design, develop, and rigorously test %the following three components of $\pname$ %(\Cref{sec:methodology}) on a broad spectrum of progressively complex tasks. Specifically, following are our %three  key contributions.
%
\begin{enumerate}[ leftmargin=*, parsep=0cm, itemindent=0em, itemsep=0.2em, topsep=0.1em]
    \item {\bf A Novel UQ Metric}: %We introduce a %distributional uncertainty 
    We introduce the
    Graph Reasoning Coherence Score (GRCS) to quantify the
    semantic-structural coherence of %the sampled reasoning space
    LLMs' reasoning, and
    derive the %reasoning 
    uncertainty proxy, $\mathcal{U}_{\mathrm{GRCS}}$, to quantify its uncertainty. 
    %a new UQ metric, Graph Reasoning Coherence Score (GRCS). %, based on Semantic-Structural Graph Edit Distance (SS-GED). 
    %GRCS is the only evaluated metric that consistently shows a negative correlation with reasoning faithfulness across tasks of increasing complexity (\textbf{88.9\%} of the time), effectively capturing hidden uncertainty in flawed reasoning paths under different independent faithfulness measures, 
    %outperforming SOTA baselines (Topo-UQ, CoTA) \textbf{77.8\%} of times. 
    $\mathcal{U}_{\mathrm{GRCS}}$ assigns higher uncertainty to less faithful reasoning in $\mathbf{88.9\%}$ of evaluated settings and outperforms Topo-UQ and CoTA in $\mathbf{77.8\%}$ of comparisons.
    %GRCS remains effective on \textbf{longer-horizon reasoning}, while baselines degrade: on AIME~'26 with DeepSeek~R1, GRCS achieves the strongest correlation in the study ($-0.542$).
    %It remains effective on longer-horizon reasoning, achieving the strongest result in the study on AIME'26 with DeepSeek~R1 (-$0.542$), where baselines weaken.
    It remains effective on longer-horizon reasoning, achieving the study's strongest result on AIME'26 with DeepSeek~R1 (-$0.542$), where baselines weaken.
    %We further show that GRCS %has a lower computational cost 
    %is more computationally efficient than baselines in terms of runtime (\(23\text{--}82\times\) faster), energy consumption (\(37\text{--}3{,}918\times\) lower), and GPU utilization (\(4\text{--}8\times\) lower), at the cost of a %larger peak memory footprint.
    %higher peak GPU memory consumption (\(1.7\text{--}7.2\times\) more). Finally, we show that variations in LLM decomposition quality have limited impact on GRCS, which remains stable across four different decomposers.
    %It is also more efficient than baselines in terms of runtime (\(23\text{--}82\times\) faster), energy %consumes 
    %(\(37\text{--}3{,}918\times\) lower), and GPU utilization (\(4\text{--}8\times\) lower), %GPU utilization, 
    %while using \(1.7\text{--}7.2\times\) more peak GPU memory.
    It is also more resource-efficient, achieving up to $82\times$ faster runtime (geometric mean: $28\times$) and $93\times$ lower energy consumption (geometric mean: $37\times$) than Topo-UQ, and up to $337\times$ faster runtime (geometric mean: $33\times$) and $3{,}917\times$ lower energy consumption (geometric mean: $125\times$) than CoTA, at the cost of $3\times$ and $1.2\times$ higher average peak GPU memory, respectively.

    \item {\bf GSC Decoding}: We propose Graph Self Consistency (GSC), a %structural 
    decoding strategy that %explicitly 
    %filters out unfaithful reasoning and retrieves the models' most faithful reasoning path (the \emph{medoid}). 
    %that 
    selects % the model's most faithful reasoning path 
    the most central and
    supported reasoning path (the \emph{medoid}) instead of the majority answer. GSC retains within-answer reasoning information ignored by SC and exposes predictions driven by \emph{lucky guesses}.
    %We theoretically prove that, unlike SC, GSC preserves the conditional reasoning entropy of each reasoning path. %(\Cref{sec:gsc_proof}) 
    %We empirically validate this result by showing that GSC filters out SC predictions supported by high conditional reasoning entropy ($\eg$, 18.1\% of cases on GSM8K and 26.7\% on MedQA) and isolates the ``lucky guesses'' that artificially inflate SC accuracy. %({\bf RQ2},~\Cref{sec_rq2_all,sec:decoding_res}).
    %Through an hybrid human/LLM-judge audit, we %demonstrate 
    We show that, when the two decoding strategies disagree, GSC selects the more logically sound answer in $\mathbf{24.9\%}$ more cases.

    \item {\bf Probing Reasoning Robustness via Adversarial Ablation}: We study how robustly models reason by removing their GSC-selected medoid. %primary reasoning topology (the \emph{medoid}). %In %compact
    %smaller LLMs, %we find that reasoning built through topological consensus is more fragile and the models often fail to arrive at the right answer through %heterogeneous 
    %diverse paths 
    %ablation reveals a high %significant 
    %negative correlation where apparent accuracy gains %systematically mask reasoning collapse
    %are coupled with lower faithfulness, empirically exposing ``lucky guesses.'' While more capable models %are able to reroute to alternative valid reasoning paths instead of collapsing. 
    %preserve their original accuracy through alternative paths, they still exhibit decreased %drops in 
    %reasoning fidelity.
    %This exposes substantial reasoning degradation, as 
    We show that faithfulness decreases in %\textbf{43.2\%} of StrategyQA cases for Llama~3.1~8B, \textbf{55.0\%} for Phi-4, and \textbf{46.3\%} of GPQA cases for DeepSeek~R1. Across all configurations, this failure affects 11--55\% of answers, with some models showing faithfulness drops of at least \textbf{50\%} in nearly one-quarter of cases.
    %$\mathbf{11\text{--}55\%}$ 
    up to ${\bf 55}\%$ (average: ${\bf 30.59}\%$) of answers, including $\mathbf{43.2\%}$ of StratQA cases for Llama~3.1~8B, $\mathbf{55.0\%}$ for Phi-4, and $\mathbf{46.3\%}$ of GPQA cases for DeepSeek~R1.
    In some settings, faithfulness drops by at least $\mathbf{50\%}$ in nearly one-quarter of cases.
    This suggests that the medoid found by GSC is %a %functions as a %single 
    %``load-bearing structure'' %in their reasoning process 
    %critical reasoning path %larger 
    %for faithful deduction.
    critical to preserving faithful reasoning.
 % ({\bf RQ3},~\Cref{sec:sec_rq3_broad_ablation}).
\end{enumerate}
%Unlike standard topological baselines, our approach forces the LLM to directly elicit its own causal dependencies. 
%The 

%GRCS stands as the only evaluated metric that consistently negatively correlates with reasoning faithfulness across all model scales, successfully capturing the hidden uncertainty within flawed reasoning paths 
%({\bf RQ1},~\Cref{sec_rq1_all,sec:ea_cf_comparison}). 
% We further show that GRCS %has a lower computational cost 
% is more computationally efficient than baselines in terms of runtime, energy consumption, and GPU utilization, at the cost of a larger peak memory footprint (\Cref{sec:GRCS_comp_cost}).
%\item
 %. We empirically demonstrate that GSC effectively isolates the ``lucky guesses'' %margin 
%that artificially inflates %standard 
%SC accuracy ({\bf RQ2},~\Cref{sec_rq2_all,sec:decoding_res}).
%\item 
%\textbf{Adversarial Vulnerability Analysis:} We expose the ``Confident Hallucination'' paradox in compact LLMs, demonstrating that identical topological consensus is not a guarantee of logical fidelity, and we provide a framework capable of detecting these poisoned paths under adversarial stress.

%\end{enumerate}

The %rest of the 
paper is organized as follows.%~\Cref{sec:background} discusses necessary background,~\Cref{sec:methodology} details our methodology,~\Cref{sec:gsc_proof} theoretically formalizes why GSC captures more information than SC through a theorem,%through three Lemmas and a Theorem,
%~\Cref{sec:exp_setup} describes the used experimental setup,%metrics and baselines,
%~\Cref{sec:exp_results} details our experimental results,~\Cref{sec_rel_work} discusses related work and positions our current research, and~\Cref{sec:conclusion} concludes the paper and %highlights 
%discusses the limitations of our approach.
~\Cref{sec:background} presents the background,~\Cref{sec:methodology} details our methodology, and~\Cref{sec:gsc_proof} theoretically formalizes GSC's information advantage over SC.~\Cref{sec:exp_setup} and~\Cref{sec:exp_results} describe the experimental setup and results, respectively.~\Cref{sec:limitations,sec:threats} discuss %the 
limitations and validity threats of our approach. %Finally,
~\Cref{sec_rel_work} reviews related work, and~\Cref{sec:conclusion} concludes the paper. %with the limitations.

\section{Background and Preliminaries}\label{sec:background}

%\subsection{Uncertainty Quantification (UQ) and LLMs}

%\StQ{You use $\hat p(x)$ but later you reuse hat for estimated conditional entropy $H$: the mixture of hats between estimate and definition could confuse some readers if not clearly distinguished. For conditional entropy, either avoid the hat or explicitly define it as an empirical estimate
%computed from finite samples.}
\noindent \textbf{Uncertainty Quantification (UQ) %and 
in LLMs}. Traditional %Uncertainty Quantification (UQ) 
UQ in Machine Learning (ML) uses predicted probability as a confidence metric in classification tasks.  %originates from standard classification tasks, where the predicted probability is used as a confidence metric. 
For a given input $\mathbf{x}$ and output class $\mathbf{y}$, the confidence in a multiclass problem is typically generalized by the class with the maximum probability, formulated as $p(\mathbf{x}) = \max_c p(\mathbf{y} = c \mid \mathbf{x})$, where $c$ indexes the output classes~\cite{guo2017calibration}. This leads to the standard Maximum Probability (MP) uncertainty measure:
$\text{U}_{\text{MP}}(\mathbf{x}) = 1 - %\max_c p(\mathbf{y} = c \mid \mathbf{x})
p(\mathbf{x})$. There are two types of uncertainty: (i)  \emph{epistemic} uncertainty, which relates to the model's lack of knowledge, and (ii) \emph{aleatoric} uncertainty, which stems from noise in the data, that confuses the model~\cite{kendall2017uncertainties}. With more input data to the model, one can reduce the former but not the latter. 
The main challenge in LLMs is that %classical 
epistemic and aleatoric measures fail to quantify the variability and structural collapse of multi-step reasoning~\cite{kadavath2022language, liu2025uncertainty}. To address this, UQ in LLMs must be extended beyond the token level to evaluate the integrity of the entire reasoning.%manifold}
~\cite{%wang2023selfconsistency, 
duan2025uprop}.

\smallskip

%\subsection{Sequence Level Decoding}

\noindent \textbf{Sequence Level Decoding}. %To quantify the accuracy of the LLMs output, %make LLMs output more accurate, we decode %can operate on the sequences they generate. 
Sequence-level decoding determines how an LLM selects complete reasoning paths and final answers from its token-level probability distribution. Greedy decoding builds a single sequence by selecting the token with the maximum conditional probability at each time step $t$~\cite{wei2022chainofthought}. %Given an input prompt $\mathbf{x}$ and a vocabulary $\mathcal{V}$, the generated token $\hat{y}_t$ is formalized as:
%$\hat{y}_t = \arg\max_{y \in \mathcal{V}} P(y \mid \mathbf{x}, \mathbf{y}_{<t})$.
While computationally efficient, the greedy approach rigidly locks the model into a single logical trajectory. If epistemic uncertainty arises early in the generation, errors compound over subsequent steps and can lead to suboptimal reasoning paths. To mitigate these issues, SC~\cite{wang2023selfconsistency} leverages the intuition that a robust reasoning problem allows for multiple distinct logical pathways that nonetheless converge on the same correct conclusion. SC thus generates a diverse set of $N$ intermediate reasoning paths and their corresponding final answers. %~\cite{wang2023selfconsistency}. %, $y_1, \dots, y_N$~\cite{wang2023selfconsistency}. %SC leverages the intuition that a robust reasoning problem admits multiple distinct logical pathways that converge on the same correct conclusion. It 
It then outputs the final answer %$\widehat{y}$ 
by applying a majority vote on the sampled reasoning paths. %: $\widehat{y} = \arg\max_{y} \sum_{i=1}^N \mathbb{I}(y_i = y)$. 
The set of sampled paths exposes the structure of the model's reasoning, enabling analysis beyond single-token probabilities~\cite{wang2023selfconsistency}.

\smallskip
%\subsection{Reasoning Diversity and Adversarial Vulnerability}

\noindent \textbf{Reasoning Diversity and Adversarial Vulnerability}.
%Beyond evaluating the final consensus of a generated sequence, 
Analyzing the variance %and the spatial characteristic 
of the intermediate reasoning manifold provides a critical signal for %uncertainty quantification
UQ~\cite{Farquhar2024SemanticEntropy}.
%,Han2024SemanticEntropyProbes
%}. 
%The distribution of sampled reasoning paths $\{z_1, \dots, z_N\}$ exhibits varying degrees of structural~\cite{Yao2023TreeOfThoughts} and logical~\cite{wang2023selfconsistency} spread, which we %can 
%refer to as \emph{reasoning diversity} and quantify it using %. We %can 
%quantify this diversity %variance 
%using 
%the entropy of the %generation 
%distribution~\cite{Shannon1948}. 
The sampled reasoning paths $\{z_1,\dots,z_N\}$ can vary structurally and logically as a result of stochastic sampling with non-zero temperature~\cite{wang2023selfconsistency}. This variation defines \emph{reasoning diversity} and can be quantified via %through 
the entropy of the %generation 
distribution of generation~\cite{Shannon1948}.
%Given the prompt $\mathbf{x}$: $ \mathcal{H}(Z \mid \mathbf{x}) = - \sum_{z \in \mathcal{Z}} P(z \mid \mathbf{x}) \log P(z \mid \mathbf{x})$. 
LLMs with %very 
low entropy can be affected by \emph{mode collapse}, repeatedly proposing only a small subset of possible reasoning paths~\cite{%goodfellow2014generative,
hamilton2024detecting}, whereas models with %characterized by 
high entropy exhibit a more diverse %heterogeneous 
and creative reasoning~\cite{Yao2023TreeOfThoughts}. %\bem{We %can 
%resort to adversarial prompting techniques to test if the generated CoTs are logically robust~\cite{zhu2024promptrobust}}. 
To test whether these reasoning paths are logically robust, adversarial prompting can perturb them and measure if %whether 
their conclusions remain stable~\cite{zhu2024promptrobust}.
% An adversarial ablation attack explicitly forbids the model from resorting to its most probable reasoning path, %denoted as 
% $z_{\text{primary}}$~\cite{zou2023universal, wang2023adversarial}. %The perturbed prompt is defined as
% $
% \mathbf{x}' = \mathbf{x} \oplus \delta_{\text{ablate}}
% $ is the perturbed prompt,
% where $\delta_{\text{ablate}}$ is an adversarial ablation perturbation that instructs the model to discard $z_{\text{primary}}$. %The modified prompt 
% This forces the model to sample from the restricted distribution
% $
% \hat{z}_{\text{alt}} \sim P(z \mid \mathbf{x}',\, z \neq z_{\text{primary}}).
% $
% \bem{We can then measure the discrepancy between the original and alternative paths to quantify the structural divergence of the model’s reasoning}.

%Next, we will define a few terminologies that we will use in~\Cref{sec:methodology}.

%\input{pic/overall_methodology}

Next, we define the notions of coherence and divergence, %and coherence, %, and introduce the running example used throughout
and reasoning uncertainty, operationalized in \Cref{sec:methodology}. %to illustrate each stage of $\pname$.

%\StQ{What about usin two more distinct notation, e.g., $D_{set}(q)$ for the set/matrix and $mean_{d}(q)$ for mean divergence?} \rreva{what do you mean by "mean divergence"? Divergence itself is the expected distance}
%\StQ{Ok. What about cleaning the definition this way? Definition 1 (Coherence and Divergence). Let $G_1, \ldots, G_N$ be reasoning graphs sampled for a given question $q$, and let $d: \mathcal{G} \times \mathcal{G} \rightarrow \mathbb{R}_{\ge 0}$ be a graph distance function. The set of pairwise distances is
%\[
%  \mathcal{D}(q) = \{ d(G_i, G_j) \mid 1 \le i < j \le N \}.
%\]
%We define the divergence $D(q)$ as the mean pairwise distance across the reasoning manifold:
%\[
%  D(q) = \frac{2}{N (N - 1)} \sum_{1 \le i < j \le N} d(G_i, G_j).
%\]
%Conversely, we define coherence $C(q)$ as
%\[
%  C(q) = \frac{1}{1 + D(q)}.
%\]}

%\subsection{Definitions}\label{sec:prelim_definitions}
\begin{mydef}[{\bf Coherence and Divergence}]\label{def:coherence_divergence}
    Consider a question $q$ posed to an LLM, from which we sample $N$ reasoning paths %are sampled 
    and represent them as graphs $G_1,\ldots,G_N$ in the space $\mathcal{G}=\{G_i\}_{i=1}^{N}.$
    %Let $G_1, \ldots, G_N$ be reasoning graphs sampled for a given question $q$ that we ask to an LLM, and let 
    Let $d: \mathcal{G} \times \mathcal{G} \to \mathbb{R}_{\ge 0}$ be a graph distance function. We define the set of pairwise distances %set 
    %is defined 
    as $\{d(G_i, G_j) \mid 1 \le i < j \le N\}$. We %formally 
    define \emph{divergence}, $\mathcal{D}(q)$, as the %expected 
    mean pairwise distance across the %reasoning manifold
    reasoning graphs: 
    $      
    \mathcal{D}(q) = \frac{2}{N(N-1)} \sum_{1 \le i < j \le N} d(G_i, G_j).
    $
    
    Conversely, we define \emph{coherence} as %$\mathcal{C}(q)$, as the degree of structural and semantic consensus within the sample. We quantify this by mapping the divergence to a bounded similarity measure: 
    $
        \mathcal{C}(q) = \frac{1}{1 + \mathcal{D}(q)}.
    $
    %$\mathcal{C}(q) = \frac{1}{1 + \mathcal{D}(q)}$.
\end{mydef}

\begin{mydef}[{\bf Reasoning Uncertainty Proxy}]\label{def:reasoning_uncertainty_proxy}
Given the coherence $\mathcal{C}(q)$ defined in \Cref{def:coherence_divergence}, %we define reasoning uncertainty 
we operationalize reasoning
uncertainty as
$
\mathcal{U}(q) = 1 - \mathcal{C}(q).
$
Higher coherence corresponds to lower reasoning uncertainty, %whereas 
Higher divergence corresponds to higher reasoning uncertainty, as %a 
more dispersed reasoning %space 
%is generally associated with 
can symptomatize %suggest 
higher entropy.
\end{mydef}

\begin{figure}
    \centering
    % Left side for the image
    %\begin{minipage}[c]{0.45\textwidth}
        \centering
        \includegraphics[width=\linewidth]{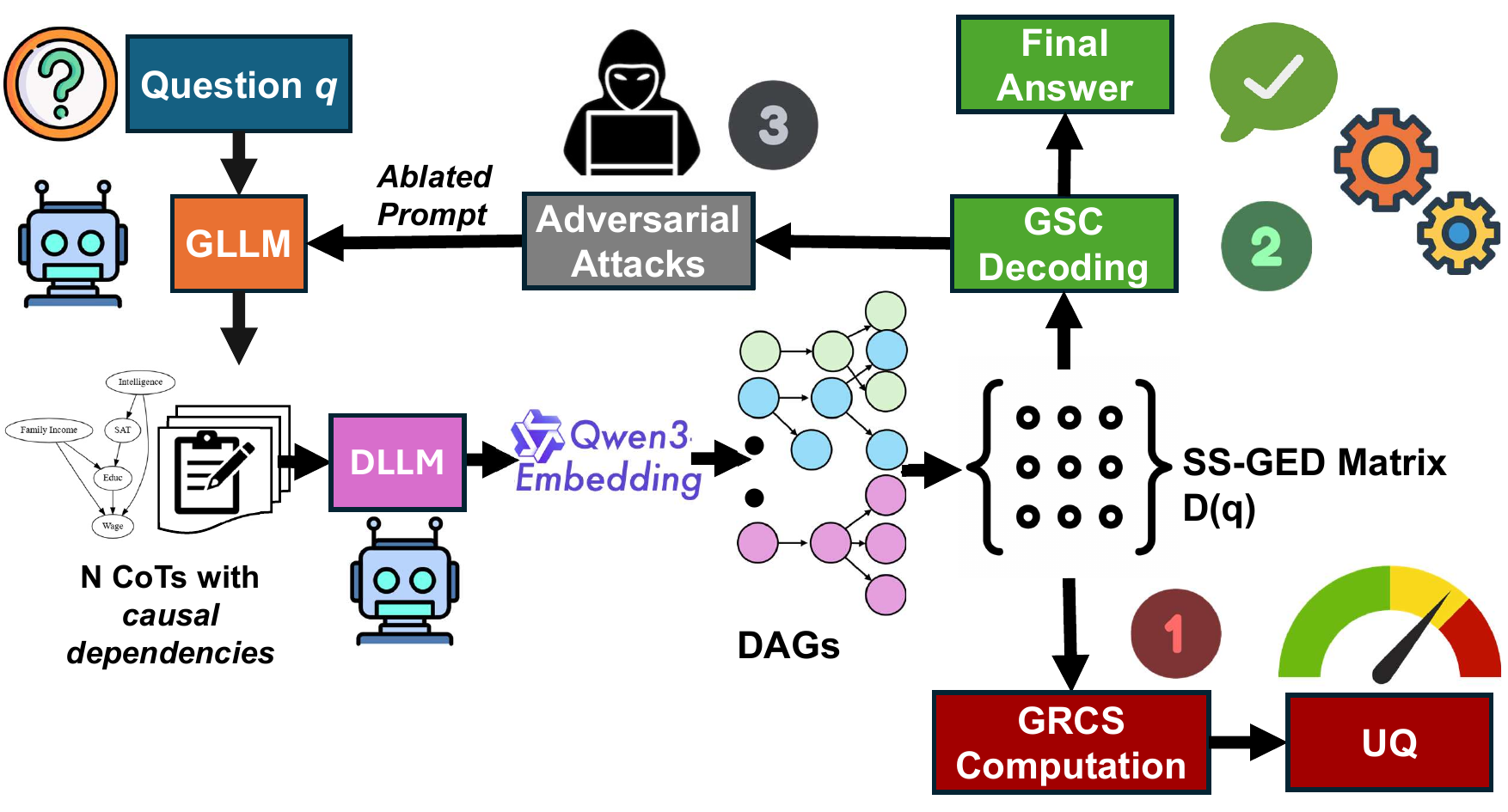}
    %\end{minipage}\hfill
    % Right side for the caption
    %\begin{minipage}[c]{0.3\textwidth}
        \caption{{\bf $\pname$ Methodology}. %{\bf GLLM}: . %GLLM's CoTs are decomposed into causal DAGs to construct a SS-GED distance matrix. 
        For each question $q$, we elicit $N$ CoTs from the Generator LLM (GLLM), decompose them into causal DAGs, and construct a matrix of their semantic-structural distances. This %shared reasoning topology 
        drives three core evaluations: {\bf (1)} GRCS-derived UQ, %uncertainty quantification, 
        which addresses \textbf{RQ1}; {\bf (2)} GSC decoding to select the most central and supported %reasoning 
        path, which addresses \textbf{RQ2}; and {\bf (3)} adversarial medoid ablation to assess medoid importance, which addresses \textbf{RQ3}.}
        \label{fig:overall_method}
    %\end{minipage}
    %\vspace{-5mm}
\end{figure}

\section{Methodology}
\label{sec:methodology}

\definecolor{cotgreen}{RGB}{235,247,235}
\definecolor{cotred}{RGB}{252,236,236}
\definecolor{questiongray}{RGB}{245,245,245}

\newcommand{\correctmark}{%
    \textcolor{green!50!black}{\raisebox{0.1ex}{\Large$\checkmark$}}%
}

\newcommand{\wrongmark}{%
    \textcolor{red!70!black}{\raisebox{0.1ex}{\Large$\times$}}%
}

\newcommand{\finalanswer}[2]{%
    \tcbox[
        on line,
        colback=#1,
        colframe=black,
        boxrule=0.45pt,
        arc=0mm,
        left=1mm,
        right=1mm,
        top=0.3mm,
        bottom=0.3mm
    ]{\textbf{Final answer: #2}}%
}
    
\begin{figure*}
\centering

% Question 
\begin{tcolorbox}[
    enhanced,
    colback=questiongray,
    colframe=black,
    boxrule=0.6pt,
    arc=0mm,
    left=7mm,
    right=2mm,
    top=2.3mm,
    bottom=1.2mm,
    overlay={
        \node[
            anchor=west,
            font=\bfseries\fontsize{24}{24}\selectfont,
            text=black,
            inner sep=0pt
        ] at ([xshift=1.5mm]frame.west) {?};
    }
]
\small\noindent
\textbf{Running Example Question.}
A laptop has 2\% battery remaining. Its charger is available. The user needs to keep working. What should they do first?
\textbf{A)} Restart the laptop.
\textbf{B)} Close all applications.
\textbf{C)} Lower the screen brightness.
\textbf{D)} Plug in the charger.
\end{tcolorbox}

\vspace{1.5mm}

% Correct CoTs
\begin{minipage}[t]{0.485\textwidth}
\begin{tcolorbox}[
    title={\textbf{CoT \#0}\hfill\correctmark},
    colback=cotgreen,
    colframe=black,
    colbacktitle=white,
    coltitle=black,
    fonttitle=\small,
    boxrule=0.5pt,
    arc=0mm,
    left=1.5mm,
    right=1.5mm,
    top=1mm,
    bottom=1mm,
    toptitle=0.5mm,
    bottomtitle=0.5mm
]
\ttfamily\footnotesize
0. The laptop has only 2\% battery. {\bfseries [D: none]}\\
1. The charger is available. {\bfseries [D: none]}\\
2. The laptop needs power to keep working.\\ {\bfseries [D: 0]}\\
3. The laptop should be plugged into the charger. {\bfseries [D: 1, 2]}\\
\textnormal{\textbf{The final answer is: D}}
\end{tcolorbox}
\end{minipage}
\hfill
\begin{minipage}[t]{0.485\textwidth}
\begin{tcolorbox}[
    title={\textbf{CoT \#1}\hfill\correctmark},
    colback=cotgreen,
    colframe=black,
    colbacktitle=white,
    coltitle=black,
    fonttitle=\small,
    boxrule=0.5pt,
    arc=0mm,
    left=1.5mm,
    right=1.5mm,
    top=1mm,
    bottom=1mm,
    toptitle=0.5mm,
    bottomtitle=0.5mm
]
\ttfamily\footnotesize
0. The laptop has little battery remaining. \\{\bfseries [D: none]}\\
1. The laptop needs power to keep working. \\{\bfseries [D: 0]}\\
2. The charger can prevent the laptop from shutting down. {\bfseries [D: none]}\\
3. Therefore, the laptop should be plugged into the charger. {\bfseries [D: 1, 2]}\\
\textnormal{\textbf{The final answer is: D}}
\end{tcolorbox}
\end{minipage}

\vspace{1.5mm}

%Incorrect CoTs

\begin{minipage}[t]{0.315\textwidth}
\begin{tcolorbox}[
    title={\textbf{CoT \#2}\hfill\wrongmark},
    colback=cotred,
    colframe=black,
    colbacktitle=white,
    coltitle=black,
    fonttitle=\small,
    boxrule=0.5pt,
    arc=0mm,
    left=1.2mm,
    right=1.2mm,
    top=1mm,
    bottom=1mm,
    toptitle=0.5mm,
    bottomtitle=0.5mm
]
\ttfamily\scriptsize
0. Charging immediately after heavy use can overheat a laptop.\\
   {\bfseries [D: none]}\\
1. The laptop should cool down before being connected to power.
   {\bfseries [D: 0]}\\
2. Lowering brightness can be done without connecting the charger.\\
   {\bfseries [D: none]}\\
3. Therefore, lower the screen brightness instead of plugging in the charger.
   {\bfseries [D: 1, 2]}\\
\textnormal{\textbf{The final answer is: C}}
\end{tcolorbox}
\end{minipage}
\hfill
\begin{minipage}[t]{0.315\textwidth}
\begin{tcolorbox}[
    title={\textbf{CoT \#3}\hfill\wrongmark},
    colback=cotred,
    colframe=black,
    colbacktitle=white,
    coltitle=black,
    fonttitle=\small,
    boxrule=0.5pt,
    arc=0mm,
    left=1.2mm,
    right=1.2mm,
    top=1mm,
    bottom=1mm,
    toptitle=0.5mm,
    bottomtitle=0.5mm
]
\ttfamily\scriptsize
0. The laptop has low battery.\\ {\bfseries [D: none]}\\
1. Plugging in a charger at 2\% battery can damage the battery.\\ {\bfseries [D: none]}\\
2. Lowering the screen brightness reduces power consumption.\\ {\bfseries [D: none]}\\
3. Therefore, the laptop should avoid charging and lower the screen brightness instead. {\bfseries [D: 0, 1, 2]}\\
\textnormal{\textbf{The final answer is: C}}
\end{tcolorbox}
\end{minipage}
\hfill
\begin{minipage}[t]{0.315\textwidth}
\begin{tcolorbox}[
    title={\textbf{CoT \#4}\hfill\wrongmark},
    colback=cotred,
    colframe=black,
    colbacktitle=white,
    coltitle=black,
    fonttitle=\small,
    boxrule=0.5pt,
    arc=0mm,
    left=1.2mm,
    right=1.2mm,
    top=1mm,
    bottom=1mm,
    toptitle=0.5mm,
    bottomtitle=0.5mm
]
\ttfamily\scriptsize
0. The laptop has very little battery remaining. {\bfseries [D: none]}\\
1. The charger is available. \\{\bfseries [D: none]}\\
2. The screen consumes battery power. {\bfseries [D: none]}\\
3. Lowering brightness reduces screen power consumption. {\bfseries [D: 1]}\\
4. Therefore, lowering brightness is better than plugging in the available charger.
   {\bfseries [D: 0, 2]}\\
{\bfseries The final answer is: C}
\end{tcolorbox}
\end{minipage}

\caption{$N = 5$ CoTs elicited from the GLLM for the running example question. \texttt{D} denotes the causal dependencies of each reasoning step. Two CoTs
reach the correct answer (Green). Three CoTs reach incorrect answers (Red). % and red backgrounds denote correct and incorrect answers.
}
\label{fig:running_example_cots}
\end{figure*}

In this section, we detail the multi-stage pipeline of $\pname$. %that %transforms 
%converts raw text generations into %analyzable topologies, 
%reasoning graphs, %thereby bringing the definitions of coherence and heterogeneity to fruition. 
%thereby operationalizing the notions of divergence and coherence.
%The pipeline consists of %cross-pollinates %consists of 
%(i) {\bf Causal Fact Decomposition and DAG Construction}, (ii) {\bf Semantic-Structural Distance Computation}, (iii) {\bf Graph Reasoning Coherence Score}, (iv) {\bf Graph Self Consistency}, and (v) {\bf GSC Evaluation via Adversarial Medoid Ablation}. \Cref{fig:overall_method} provides a high-level overview of our methodology.
To address \textbf{RQ1}, we construct causal DAGs (\Cref{sec:causal_fact_decomposition}), compare their semantic-structural distance (\Cref{sec:ss_distance_computation}), and aggregate their distances into GRCS (\Cref{sec:grcs}). For \textbf{RQ2}, GSC selects the medoid reasoning path instead of majority voting (\Cref{sec:consensus_context_synthesis}). For \textbf{RQ3}, we adversarially ablate this medoid and evaluate how reasoning is affected (\Cref{sec:faithfulness_adv_medoid_ablation}). \Cref{fig:overall_method} illustrates our methodology.

We use the following %running 
simple example to explain different %components 
stages of $\pname$. %\noindent\bem{Running Example.}
%Throughout %our methodology (\Cref{sec:methodology})
%\Cref{sec:methodology}, we refer to the following running example. % to illustrate how each stage of $\pname$ work. 
We ask an LLM the following question: %\emph{``Stephany has \$20, spends \$6, and then receives \$4. How much money does she have now?''}. 
\emph{A laptop has 2\% battery remaining. Its charger is available.
The user needs to keep working. What should they do first?
(A) Restart the laptop.
(B) Close all applications.
(C) Lower the screen brightness.
(D) Plug in the charger}. Intuitively, the %right 
correct answer is %intuitively 
\emph{(D) Plug the charger}.

% To operationalize the definitions of coherence and heterogeneity, we propose a multi-stage pipeline that transforms raw text generations into analyzable topologies. The pipeline cross-pollinates %consists of 
% (i) \bem{Causal Fact Decomposition}, (ii) \bem{Semantic-Structural Distance Computation}, (iii) \bem{Consensus Context Synthesis}, and (iv) \bem{Faithfulness Evaluation}.

%\smallskip 

%\paragraph{Causal Fact Decomposition and DAG Construction} \label{sec:methodology_decomposition}

\subsection{Causal Fact Decomposition and DAG Construction}\label{sec:causal_fact_decomposition}

%\noindent {\bf Causal Fact Decomposition and DAG Construction}. %The process begins by prompting the Generator LLM (\ie, the LLM %we want to 
%under study, GLLM) to produce a step-by-step CoT for a given question. 

%The goal of this step is to %transform 
%convert raw, textual LLM responses into %structured,  %mathematically analyzable formats. 
%analyzable reasoning graphs.
To convert raw textual LLM responses into analyzable reasoning graphs, we address the lack of a common structure for semantic and causal comparison by decomposing each reasoning path into atomic facts and causal dependencies, and encoding them as a DAG.
For each question $q$, we prompt a Generator LLM (\textbf{GLLM}) to generate $N$ CoTs at a high temperature $T$ to capture its reasoning diversity. We constrain each CoT to contain between $n_l$ and $n_h$ declarative steps to standardize its length.
%to ensure consistency and avoid verbosity. -> APPENDIX FOR 3-6
%In particular, the model 
The GLLM needs to explicitly declare the \emph{causal structure} of its reasoning by annotating each step with its direct logical antecedents, \eg, \texttt{[D: 0, 1]}, where \texttt{D} denotes that the current step depends on steps \texttt{0} and \texttt{1}.
%denotes the causal dependencies %, and %where 
%between step \texttt{0} and the current step, and step \texttt{1} and the current step. %are the logical antecedents of the current step. % The generator is restricted to 3--6 declarative sentences to ensure logical density while avoiding verbosity. To map the model’s reasoning manifold and ensure statistical significance, we elicit $N=20$ distinct reasoning paths per query. By employing a stochastic sampling temperature ($T=0.7$), we capture the diversity of the model's logical trajectories. 
%\StQ{Once defined, use the acronym DLLM consistently and specify the exact prompt and validation protocol.}
%\rreva{Specifying the exact prompts used for GLLM, DLLM and the adversarial attack would eat a lot of space. I provided a simplified prompts thet unifies the two tasks: GENERATE for GLLMs, DECOMPOSE for the DLLM.}

In our running example, we elicit $N=5$ distinct CoTs from the GLLM. As \Cref{fig:running_example_cots} shows, two CoTs
reach the correct answer, \emph{D}, through consistent reasoning.
Three CoTs reach an incorrect answer, \emph{C}, through unfaithful reasoning: CoT \#2 and \#3 are based on a hallucination ($\ie$, they invent that charging a laptop at 2\% can damage the battery), CoT \#4 contains %has 
a logical gap ($\ie$, %it states that 
lowering brightness reduces power consumption, but has no facts supporting the conclusion that lowering brightness is preferable to using the charger).

Subsequently, we need to decompose each reasoning to enable comparison across CoTs. To mitigate inductive bias from self-evaluation, we use a distinct Decomposer LLM (\textbf{DLLM}), %. Specifically, we use 
%OpenAI’s GPT-4o %as a high-fidelity fact extractor 
%\cite{openai2024gpt4o}, 
operating at deterministic temperature ($T=0.0$) to reduce stochastic variation in the decomposition. % to ensure consistency and precision. 
Following prior work, the DLLM %transforms 
breaks coarse reasoning steps into %granular 
\emph{Subject--Verb--Object} atomic facts~\cite{min2023factscore,zhang2024luq}. %Importantly, 
The DLLM must preserve the original \emph{causal dependencies}, %and causal structure, 
refining granularity without altering the GLLM's %Generator LLM's %logical 
reasoning. %The prompting strategy used for GLLM and DLLM is explained in Appendix \Cref{sec:appendix_prompts}.
%Prompts for GLLMs and DLLM are provided in Appendix~\ref{sec:appendix_prompts}.
\Cref{lst:unified_prompt} shows condensed versions of the prompts used for GLLMs and DLLM.
%These 
%Facts and dependencies are then mapped into a DAG, %Directed Acyclic Graph (DAG), 
%$G = (V, E)$, where each node $v \in V$ represents an atomic fact embedded via a high-%dimensional sentence transformer (Qwen3-Embedding-0.6B,~\cite{yang2025qwen3technicalreport}), and each directed edge $e \in E$ represents a causal dependency.
%Facts and dependencies are then mapped into a DAG $G=(V,E)$, where each node $v\in V$ is an atomic fact embedded with Qwen3-Embedding-0.6B~\cite{yang2025qwen3technicalreport}, and each directed edge $e\in E$ denotes a causal dependency.
%Facts and dependencies 
%The 

We then map the DLLM's output into a \textbf{DAG} $G=(V,E)$, where $v \in V$ are atomic facts and $e\in E$ are the causal dependencies between them.
%embedded with an embedding Transformer %Qwen3-Embedding-0.6B~\cite{yang2025qwen3technicalreport} 
%In this process, 
We encode the textual content of each fact using an embedding Transformer, and we store each resulting vector $\mathbf{h}_v$ as a DAG node.
The causal dependencies have unitary weights, since GLLM-assigned weights could inherit the same overconfidence our pipeline aims to detect, assigning high confidence even to flawed dependencies. 
% and are not %independently 
%embedded.
%Since the GLLM specifies only the presence of a dependency and not its strength, all edges are assigned a uniform weight of $1.0$.

For the running example, \Cref{fig:running_example_dag} shows the DAG constructed from CoT \#0 after the DLLM decomposition.

\definecolor{premisecolor}{RGB}{225,235,248}
\definecolor{intermediatecolor}{RGB}{255,245,204}
\definecolor{finalcolor}{RGB}{223,242,218}

\begin{figure}
\centering

\begin{minipage}[c]{0.56\columnwidth}
\centering
\begin{tikzpicture}[
    fact/.style={
        draw,
        rounded corners=1pt,
        align=center,
        inner sep=2pt,
        font=\scriptsize,
        minimum width=21mm,
        minimum height=6mm
    },
    premise/.style={
        fact,
        fill=premisecolor,
        draw=black
    },
    intermediate/.style={
        fact,
        fill=intermediatecolor,
        draw=black
    },
    final/.style={
        fact,
        fill=finalcolor,
        draw=black
    },
    edge/.style={
        -{Latex[length=1.3mm]},
        thin
    }
]

\node[premise] (v0) at (-1.25,0) {Laptop has only\\2\% battery.};
\node[premise] (v1) at ( 1.25,0) {Charger is\\available.};

\node[intermediate] (v2) at (-0.35,-1.0) {Laptop needs power\\to keep working.};
\node[intermediate] (v3) at (0.70,-2.0) {Laptop should be\\plugged into charger.};

% Final answer
\node[final] (v4) at (0.70,-3.0) {Final answer:\\D};

% Dependencies
\draw[edge] (v0) -- (v2);
\draw[edge] (v1) -- (v3);
\draw[edge] (v2) -- (v3);
\draw[edge] (v3) -- (v4);

\end{tikzpicture}
\end{minipage}
\hfill
\begin{minipage}[c]{0.40\columnwidth}
\raggedright
\captionsetup{type=figure,justification=raggedright,singlelinecheck=false}
\caption{\textbf{DAG of the decomposed CoT \#0}. Nodes are atomic facts (Blue: annotated with \texttt{[D: none]}; Yellow: annotated with non-empty \texttt{[D: \dots]}; Green: final answer). Edges are dependencies specified by \texttt{D}.}
\label{fig:running_example_dag}
\end{minipage}

\end{figure}

%\smallskip

%\paragraph{Semantic-Structural Graph Edit Distance (SS-GED)}\label{par:ss_ged}
\begin{lstlisting}[
  style=promptstyle,
  mathescape=true,
  caption={Unified GLLM/DLLM prompt. GLLM performs the GENERATE task. DLLM performs the DECOMPOSE.},
  label={lst:unified_prompt},
  emph={TASK,INPUT,GENERATE,DECOMPOSE,Rules,depends_on,step,final,answer,ONLY}
]
You are a precise causal reasoner.

TASK: {TASK};   INPUT: {INPUT}

If TASK = GENERATE:
- Answer the input question using $n_l$--$n_h$ numbered steps, starting from 0.
- Each step must be one declarative sentence expressing one fact or conclusion.
- Append [depends_on: N, ...] with all directly required prior steps, or
  [depends_on: none] for an independent premise.
- Do not repeat the question or add headers, bullets, tags, or extra text.
- End with: The final answer is: [answer]

If TASK = DECOMPOSE:
- Decompose every numbered step into independent subject--verb--object atomic facts.
- Append
  [step: N, depends_on: ...] to each fact, preserving the original steps and
  dependencies.
- Replace pronouns with their original subjects.
- Do not alter dependencies or repeat, rephrase, infer, or invent facts.
- Output ONLY the annotated atomic facts.
\end{lstlisting}

\subsection{Semantic-Structural Distance Computation}\label{sec:ss_distance_computation}

%\noindent {\bf Semantic-Structural Distance Computation%Semantic-Structural Graph Edit Distance (SS-GED)}. 

%To quantify the \emph{structural} and \emph{semantic} consistency across a %query's 
%$q$'s reasoning manifold, \ie, set of elicited CoTs, %we compute the pairwise distance $d(G_i, G_j)$ between all $\binom{N}{2} = 190$ pairs of reasoning graphs generated for a %single 
To quantify the semantic and structural consistency of the GLLM's reasoning, we address the lack of a scalable measure for comparing the DAGs constructed in the previous stage by
%we compute 
computing pairwise distances $d(G_i,G_j)$ over all %$\binom{N}{2}$ %$=190$ 
%graph 
DAG pairs. %, where each $G=(V,E)$ encodes a CoT as a set 
%of atomic reasoning facts $v \in V$ connected by causal-dependency edges 
%%$e \in E$.
 %By representing CoTs as a DAGs, we measure the stability of the model's logic. %, because a 
%A coherent model should exhibit low distance and thus high similarity between paths that converge on the same conclusion, while high pairwise distances across the $N=20$ samples indicate reasoning heterogeneity or logical instability.
A coherent GLLM should yield low pairwise distances between DAGs sharing the same conclusion, whereas high distances across the $N$ %=20$ 
%samples 
DAGs may indicate reasoning %heterogeneity. %or instability.
uncertainty.
Because this comparison must be repeated over all $\binom{N}{2}$ DAG pairs, %for every 
%question, across all datasets and GLLMs, 
computing the 
\emph{exact} Graph Edit Distance (GED~\cite{sanfeliu1983distance}), that is an NP-hard search, would be %computationally 
prohibitive~\cite{da2025understanding}. 

We therefore compute an \emph{approximate} GED based on optimal bipartite matching~\cite{riesen2009approximate}, which reduces the problem to two polynomial-time linear assignment problems, one over nodes and one over edges, solved via the Hungarian algorithm~\cite{korsah2007dynamic} in 
$\mathcal{O}(m^3)$ time (where $m \approx |V_i|+|V_j|$).
Nodes %(respectively edges) 
are paired via a one-to-one assignment to minimize total substitution cost, equal to cosine distance between the embeddings of each pair, $c_{sub} = 1-\cos(h_a,h_b)$ %($h_x$: embeddings of node/edge $x$). 
where $\mathbf{h}_a$ and $\mathbf{h}_b$ are the embeddings assigned in the previous stage to the %corresponding 
atomic facts. % or causal dependencies. 
Causal dependencies are also paired via a one-to-one assignment, but the substitution cost is unitary.
Because $|V_i|$ and $|V_j|$ (respectively $|E_i|$, 
$|E_j|$) may differ, the $\lvert|V_i|-|V_j|\rvert$ facts (respectively $\lvert|E_i|
-|E_j|\rvert$ dependencies) 
left unmatched ($\ie$, reasoning steps or causal dependencies present in 
one CoT but absent from the other) %after the optimal assignment 
are charged a fixed unitary 
insertion/deletion cost. 

\begin{figure}
    \centering
    \includegraphics[scale=0.24]{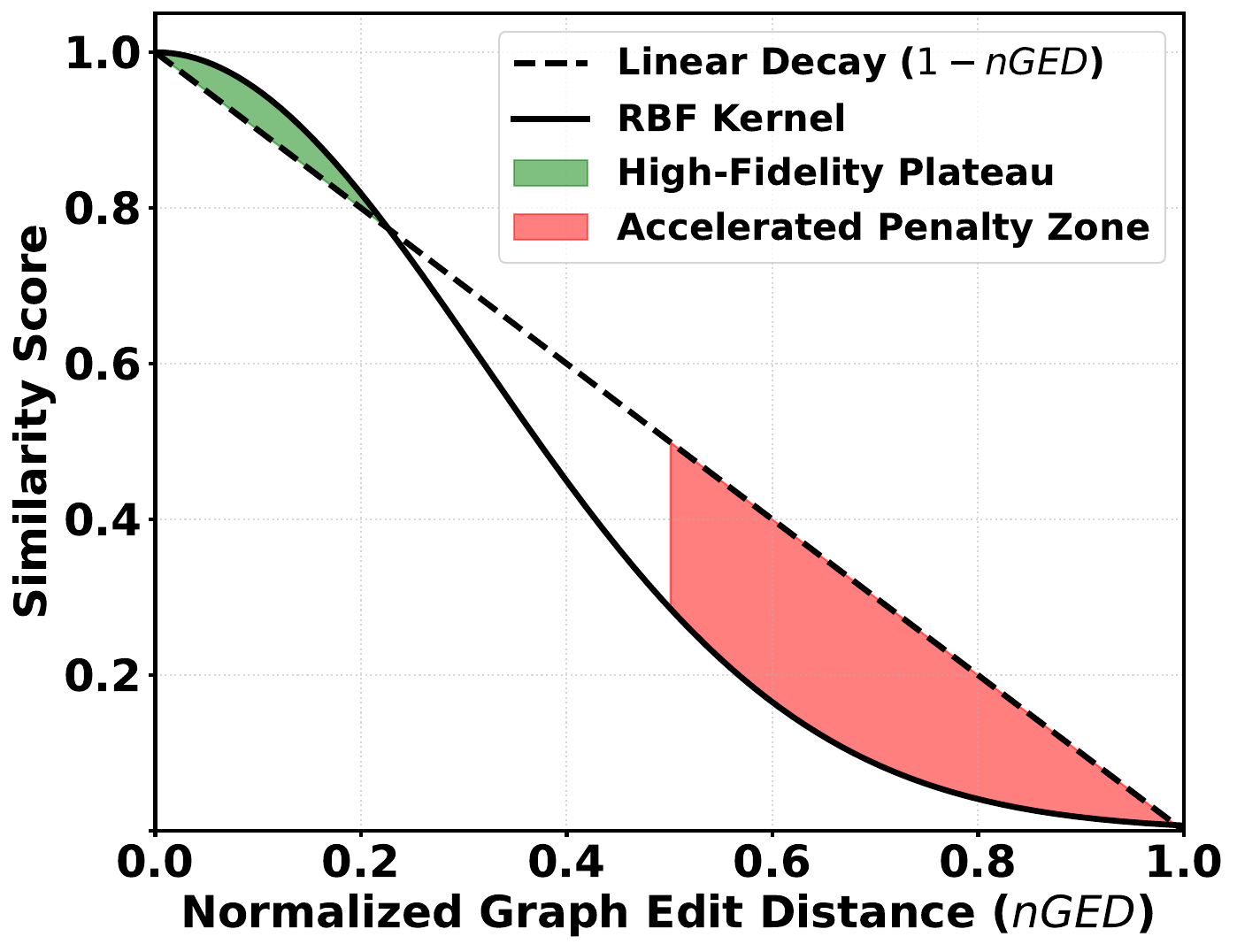}
    \caption{\textbf{Comparison of similarity decay models.}
        Linear decay assigns a linear penalty to all distances. The RBF kernel, used in our methodology, establishes a \emph{High-Fidelity Plateau} to tolerate
        minor structural variations and an \emph{Accelerated Penalty Zone} to
        isolate %significantly 
        highly divergent reasoning paths. %In the chart,
        %$\gamma=5.0$ yields a moderately sharp RBF decay. %, so that small structural differences are not over-amplified while distinct graphs remain well separated.}
        }
    \label{fig:rbf_fig_appendix}
\end{figure}

We solve node and edge assignments as an \emph{independent} linear assignment problem % over their respective embedding spaces, 
to preserve the $\mathcal{O}(m^3)$ complexity and to avoid biasing the distance against CoTs 
that place the same causal dependencies in a different order.
%To compute this distance, we extend %the GED metric 
%GED~\cite{riesen2009approximate} %by incorporating 
%with semantic awareness and optimized bipartite matching. %A formulation of GED is available at Appendix~\ref{sec:ged_def_appendix}. 
%We determine the optimal matching between nodes and edges in $\mathcal{O}(m^3)$ (where $m \approx (\lvert V_i \rvert + \lvert V_j \rvert)~\text{for graphs } G_i \text{ and } G_j$) time %using computationally efficient %the 
%Hungarian algorithm %, which offers significant computational efficiency 
%over NP-hard A$^\star$ search %baselines 
%using the computationally efficient Hungarian algorithm rather than NP-hard A* search baselines
%\cite{korsah2007dynamic}. %Substitution costs are defined as the cosine distance ($1 - \cos(\theta)$) between high-dimensional node and edge embeddings, calculated using vectorized matrix operations to handle large-scale comparisons.
%\StQ{Explicitly state that this is an approximate GED based on bipartite matching, not the exact approach. Explain node/edge insertion, deletion, and substitution costs and whether edge consistency is enforced.} \rreva{DONE ABOVE.}
%The substitution cost in GED while matching is the cosine distance, $1-\cos(\theta)$, where $\theta$ is the angle between node or edge embedding vectors.
%We introduce two critical improvements to address flaws in standard topological GED. 
We make two key improvements to address the limitations of topological GED. 
%\bem{First}, we implement \textbf{Fair Normalization} to account for variations in reasoning depth. Standard normalized GED (nGED,~\cite{marzal1993normalized}) typically divides the edit cost by $\max(|V_i|, |V_j|)$, %a method that disproportionately penalizes models for generating highly detailed, verbose explanations that get translated into very long graphs. To ensure the metric remains stable regardless of reasoning granularity, we instead normalize the total edit cost by the average graph size: $\frac{|V_i| + |V_j|}{2}$.
%making the normalization depend entirely on the larger graph. When one reasoning graph is much longer than the other, this can make comparisons less balanced. We therefore normalize by the average graph size, $\frac{|V_i| + |V_j|}{2}$, so that the scale depends symmetrically on both graphs.
First, we introduce \textbf{Fair Normalization}, scaling the edit cost by the average graph size ${(|V_i|+|V_j|})/{2}$, %$\frac{|V_i|+|V_j|}{2}$, 
so that the comparison scale depends only on the graphs $G_i$ and $G_j$ under consideration. %symmetrically on both graphs.
%\bem{Second}, we use \textbf{Semantic-Structural Blending} to reconcile topology with logical meaning. Because two graphs can share identical structures yet reach opposing conclusions, the SS-GED score blends topological distance with the semantic similarity~\cite{salton1983introduction} of the terminal nodes -- the final conclusions of the reasoning chains. We extract terminal embeddings by averaging the vectors of all nodes with an out-degree of zero. The final unified similarity score $S(G_i, G_j)$ is computed as:
Second, we introduce \textbf{Semantic-Structural Blending} to account for cases where two graphs share similar topology but reach different conclusions.
%\StQ{It looks like that nGED is used but not explicitly defined (normalized edit distance. Add somewhere: Let $nGED(G_i,G_j)$ denote the SS-GED distance normalized by the Fair Normalization factor described above.}
Let $\text{nGED}(G_i,G_j)$ denote the GED distance normalized by the Fair Normalization factor described above.
Semantic-Structural GED (SS-GED), therefore, combines structural similarity with the semantic similarity %~\cite{salton1983introduction} 
of terminal nodes, computed by averaging the embeddings of all nodes with out-degree zero. The resulting similarity is
$S(G_i, G_j) = \beta \cdot S_{conc} + (1 - \beta) \cdot \exp(-\gamma \cdot \text{nGED}^2)$ where $S_{conc}$ %represents the cosine similarity of the terminal embeddings, 
is cosine similarity of terminal-nodes, %cosine similarity, %$\gamma$ (set to 5.0) 
$\gamma$ %=5.0$ 
controls the Radial Basis Function (RBF) kernel spread for structural similarity, and %$\beta$ (default 0.4) weights the relative importance of the semantic outcome. The final distance used for manifold analysis is defined as $d(G_i, G_j) = 1 - S(G_i, G_j)$.
$\beta$ %=0.4$ 
weights semantic similarity. %, placing slightly greater emphasis on structural similarity to prioritize the reasoning over final answers. 
We use an RBF transform to penalize more distant graphs, as \Cref{fig:rbf_fig_appendix} shows. The final SS-GED is $d^\prime(G_i,G_j)=1-S(G_i,G_j)$. %Because node and edge assignments are independent, SS-GED compares edges independently of node assignments, without enforcing that matched edges connect matched node pairs.
SS-GED compares edges independently of node assignments, without enforcing that matched edges connect matched node pairs. This avoids over-penalizing CoTs whose semantically equivalent reasoning facts have different embeddings and may result in non-identical DAGs.

%\emph{Step 4: Compute Pairwise SS-GED.} With $N=5$ DAGs, 
In our example, we compute the SS-GED distance $d^\prime(G_i,G_j)$ for all
$\binom{5}{2}=10$ pairs, using $\beta$, $\gamma$ from \Cref{sec:exp_setup}. %Consider graphs $G_0$ and $G_1$, corresponding to CoTs \#0 and \#1 (\Cref{fig:running_example_cots}), that correctly answer \emph{D}. 
Graphs $G_0$, $G_1$ (from CoTs \#0, \#1 in \Cref{fig:running_example_cots}) share 
atomic facts %express 
with semantically similar meanings ($\eg$, the battery
is almost empty, the laptop needs power, and the available charger provides
it), and connected with similar causal dependencies. Thus,
$\mathrm{nGED}(G_0,G_1)=0.18$ and, because both reach answer \emph{D},
$S_{\mathrm{conc}}=0.98$, yielding a small SS-GED
$d^\prime(G_0,G_1)=0.10$. %Thus, $\mathrm{nGED}(G_0,G_1)$ is small. The two reasonings arrive at the same final answer, thus $S_{\mathrm{conc}}$ is high. Therefore, $S(G_i, G_j)$ is high and \textbf{SS-GED is low}.
The graphs of the other three CoTs that answer \emph{C} %share the same terminal answer but 
differ substantially in atomic facts and dependencies, and produce %higher values of $\mathrm{nGED}$, and %hence 
larger SS-GEDs % distances%from the
%correct pair
, reaching up to $0.59$ (average: $0.50$).
%a higher SS-GED. 
\textbf{Thus, SS-GED preserves final-answer agreement and penalizes inconsistent reasoning.}

\subsection{Graph Reasoning Coherence Score}\label{sec:grcs}

%\noindent {\bf Graph Reasoning Coherence Score}.

% \StQ{Clarify how GRCS detects identical-but-flawed mode collapse. Pairwise distance among repeated identical hallucinations should be low, yielding high coherence, unless an external correctness/semantic anchor is used. Revise the claim or the metric definition.}\rreva{revised the claim: GRCS flags \textbf{partial} mode collapse. Run two experiments to practically prove that GRCS detects semantic-structural consensus and partial model collapse. Full collapse weakens GRCS, but it's true that this condition happens very rarely. GRCS is robust at most when approx 40\% of CoTs collapse.}

%To provide a holistic measure of the GLLM's reasoning stability, this step aggregates the pairwise SS-GED distances into a single distributional uncertainty metric, Graph Reasoning Coherence Score (GRCS). %representing our core framework product. 
To measure the GLLM's reasoning stability, this step aggregates the pairwise SS-GED distances into a single metric, Graph Reasoning Coherence Score (GRCS).
%First, we use %using 
%the pairwise distance set 
%$\mathfrak{D}(q)$ (Definition~\ref{def:coherence_divergence}) to create a pairwise SS-GED matrix containing %, $\mathfrak{D}(q) = \{d(G_i, G_j) \mid 1 \le i < j \le N\}$, with 
%distances among all $N$ elicited CoTs for each $q$. Building upon $\mathfrak{D}(q)$, we instantiate our formal definition of coherence ($\cf$ \Cref{sec:background}), defining the GRCS using the mean pairwise distance, $\mu(\mathfrak{D}(q))$, across the reasoning manifold: 
First, we compute the set of pairwise SS-GED distances
$\mathfrak{D}(q) =\{d^\prime(G_i, G_j)\}$ among the $N$ elicited CoTs for each question $q$.
Building upon $\mathfrak{D}(q)$, we instantiate our formal definition
of coherence ($\cf$, \Cref{def:coherence_divergence}) as
%\begin{equation}
$
\mathrm{GRCS} = 1/ %\frac{1}
({1+\mu(\mathfrak{D}(q))}),
$
%\iffalse
%GRCS quantifies the semantic-structural consensus of the reasoning space and captures pathological mode collapse and confident hallucinations. Mode collapse occurs when a GLLM repeatedly proposes a narrow subset of identical and often flawed reasoning shortcuts rather than exploring a diverse and robust reasoning manifold. %GRCS captures this behavior thanks to its reliance on the SS-GED matrix. 
%Confident hallucinations happen when a GLLM generates multiple CoTs relying on fabricated atomic facts or disconnected conclusions. If a GLLM exhibits these behaviors, the semantic penalty within SS-GED increases the average pairwise distance.
%Conversely, the structural penalty penalizes the ``lucky guesses'', because the topological variation stemming from highly divergent facts inflates the distance.
%\rreva{here, put the reason why we chose to create GSC. Consider this as a downstream, practical application of GraphEVAL framework}
%\fi
%GRCS quantifies semantic-structural consensus along two axes: 
where $\mu(\mathfrak{D}(q))$ is the mean of the set of SS-GED distances.
GRCS lies in $(0,1]$: values closer to $1$ indicate stronger
semantic-structural agreement among the sampled reasoning paths, whereas
lower values indicate %greater 
reasoning disagreement.
%GRCS instantiates the coherence $\mathcal{C}(q)$ from \Cref{def:coherence_divergence}. 
Following \Cref{def:reasoning_uncertainty_proxy}, we define GRCS's corresponding reasoning uncertainty proxy as:
$
    \mathcal{U}_{\mathrm{GRCS}} = 1 - \mathrm{GRCS}.
$
%Higher GRCS indicates greater reasoning coherence and lower
%uncertainty. 
Thus, higher ${\cal U}_{\mathrm{GRCS}}$ indicates greater
%reasoning 
uncertainty.

GRCS captures two behaviors: \textbf{semantic-structural consensus} of the reasoning space, which occurs when the GLLM's reasoning has consistent causal-effect relationships across its DAG and corresponding nodes across those DAGs carry consistent meaning, and \textbf{partial mode collapse} in the GLLM's sampled CoTs, which occurs when the GLLM repeatedly proposes a narrow subset of identical and %often %flawed 
hallucinated reasoning paths. %shortcuts. %We verify these two behaviors by injecting controlled contamination into real elicited sets of CoTs. 
%We verify these two behaviors by injecting controlled 
%contamination into real elicited sets of $N=20$ CoTs for each question on StrategyQA~\cite{geva2021aristotle}, embedding all 
%9{,}583 unique facts and 17{,}504 unique causal edges with 
%$Qwen3-Embedding-0.6B~\cite{yang2025qwen3technicalreport} and repeating each sweep over 5 independent runs.
%contamination into 
%real elicited sets of $N=20$ CoTs, embedding facts and causal edges with 
%Qwen3-Embedding-0.6B~\cite{yang2025qwen3technicalreport} and repeating each 
%sweep over 5 independent runs across all evaluated 
%benchmarks~(\cf, \Cref{sec:exp_setup}).
%For 
%each question, we replace a fraction $g$ of its $N$ %real 
%elicited CoTs with %real 
%CoTs 
%sampled 
%elicited from other, unrelated questions (\Cref{fig:grcs_sweeps} left), and 
%separately replace a fraction $f$ with near-duplicates of one designated 
%flawed CoT (\Cref{fig:grcs_sweeps} right), recomputing GRCS at each fraction. GRCS falls monotonically with contamination, from $0.65$ at 
%$g{=}0$ to $0.42$ at $g{=}1$, because it rewards genuine agreement rather 
%than flawed repetition. Under collapse, GRCS stays flat or below 
%its $f{=}0$ baseline for $f \lesssim 0.4$, correctly flagging partial mode collapse, then 
%rises once collapse dominates the manifold. 
We verify these two behaviors with two independent sweeps across all evaluated benchmarks described in \Cref{sec:exp_setup}. In the consensus sweep, for each question, we replace a fraction $g$ of the $N$ elicited CoTs with CoTs elicited for other, unrelated questions, to simulate a GLLM losing its semantic structural-consensus (\Cref{fig:consensus_frontier}). In the collapse sweep, we separately replace a fraction $f$ of the same question's $N$ CoTs with $f \cdot N$ near-duplicate copies of one CoT that belongs to those $N$ CoTs. Each copy is perturbed to keep a cosine similarity of 0.98 to the original CoT rather than being an identical copy, to simulate a GLLM repeatedly proposing the same reasoning with a different wording \Cref{fig:collapse_frontier}). For both experiments, we compute GRCS at each fraction. In the consensus sweep, depending on the dataset, GRCS falls monotonically with $g$ for every dataset, from baselines between $0.42$ (AIME'26) and $0.69$ 
(GSM8K) at $g{=}0$ down to $0.34$--$0.41$ at $g{=}1$. This means %\bem{As CoTs get replaced with unrelated reasoning, the semantic-structural agreement of the reasoning drops and GRCS gets lower}. 
\textbf{GRCS gets lower when the semantic-structural agreement of the reasoning drops}.
In the collapse sweep, 
%\rreva{here: 2 cases (if the cot is hallucinated, partial mode collapse. if the cot is right -> grcs awards consistency!!! pyt this and review this part}
every dataset dips slightly below its $f{=}0$ baseline around $f{\approx}0.1$, 
stays flat or below baseline out to $f{\approx}0.3$--$0.4$, then rises 
sharply and converges to $0.88$--$0.92$ by $f{=}1$ regardless of starting 
point.
This means \textbf{GRCS flags partial mode-collapse}. Up to $f{\approx}0.4$, the duplicated CoTs sit inside a diverse manifold, and GRCS stays low. Past that point, duplicates dominate the manifold, pairwise 
distances shrink toward zero, and GRCS rises. %toward its spurious ceiling. 
In the cases where the copies are faithful to the reasoning, GRCS rewards consensus. If the copies are hallucinated, confident repetition of the same error dominates the reasoning to the point of total mode collapse. Baselines at ($f{=}0$ and $g{=}0$) track task complexity ($\eg$, GRCS for GPQA and AIME~'26 is lower because these two datasets are the most difficult).

\definecolor{medqacol}{RGB}{214,39,40}
\definecolor{gpqacol}{RGB}{31,119,180}
\definecolor{boolqcol}{RGB}{197,27,138}
\definecolor{stratqacol}{RGB}{44,160,44}
\definecolor{gsm8kcol}{RGB}{244,124,86}
\definecolor{aimecol}{RGB}{123,104,174}

\newcommand{\legenditem}[2]{%
    \begin{tikzpicture}[baseline=-0.6ex]
        \draw[#1, line width=1.0pt] (0,0) -- (0.55,0);
        \fill[#1] (0.275,0) circle (1.7pt);
        \node[anchor=west, font=\bfseries\footnotesize] at (0.72,0) {#2};
    \end{tikzpicture}%
}

\begin{figure}%[t]
    \centering
    \captionsetup[subfloat]{labelfont=bf}
    \subfloat[\textbf{\small Consensus frontier}
    \label{fig:consensus_frontier}]{
        \includegraphics[width=0.45\linewidth]{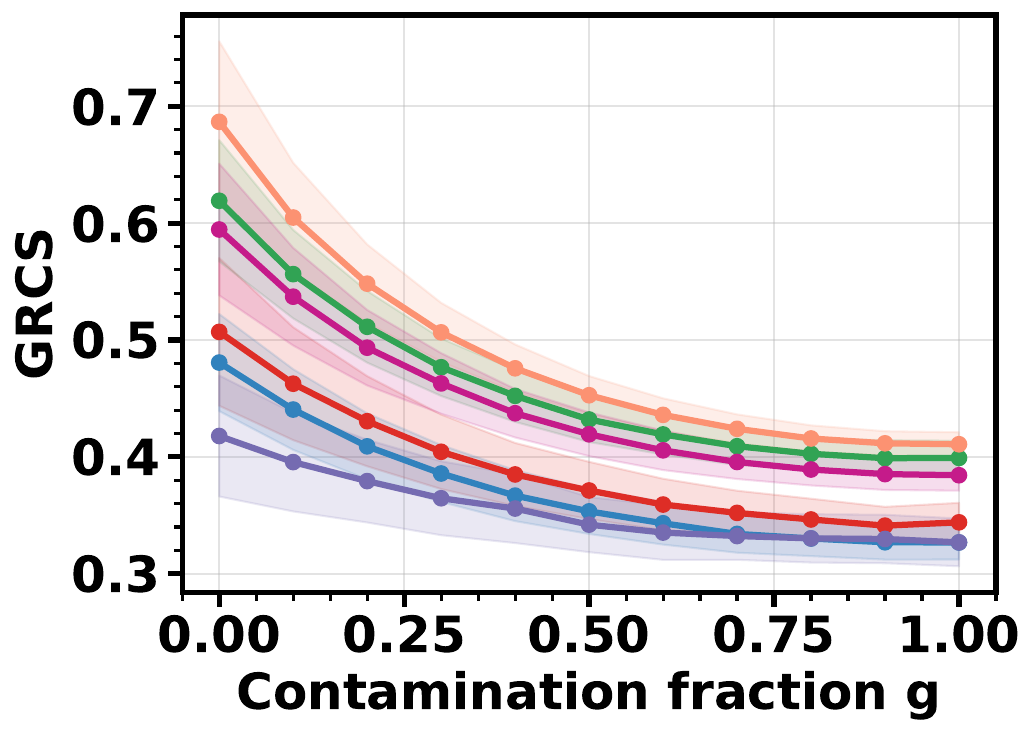}
    }
    \hfill
    \subfloat[\textbf{\small Mode-collapse frontier}
    \label{fig:collapse_frontier}]{
        \includegraphics[width=0.45\linewidth]{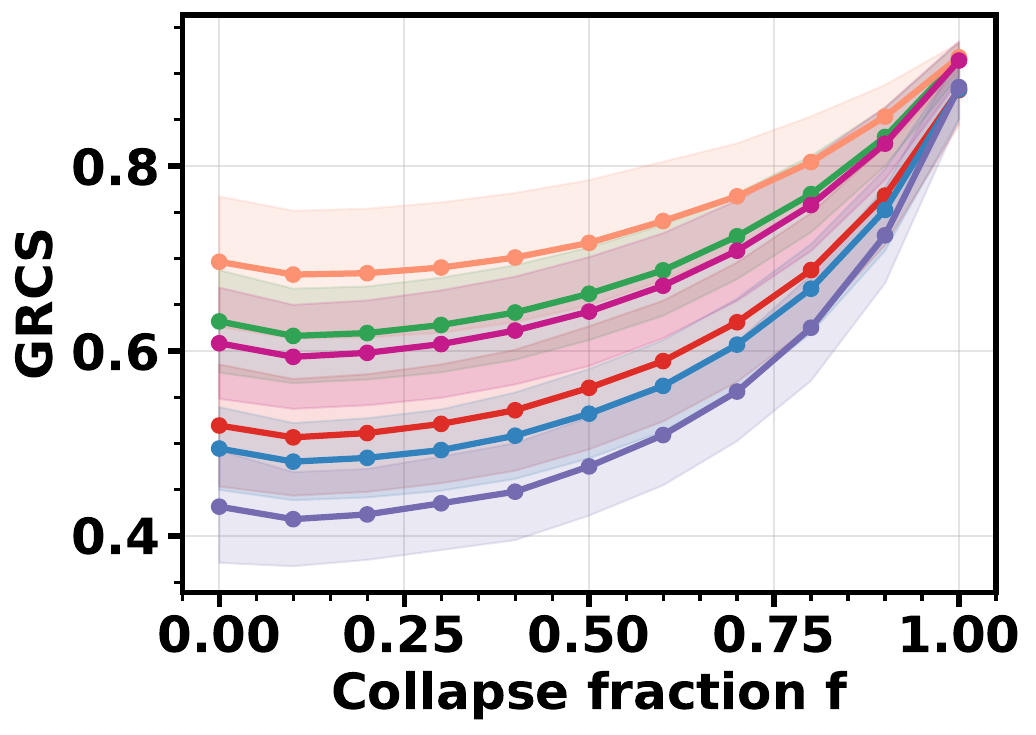}
    }

    \vspace{0.5em}

    \begin{center}
    \setlength{\tabcolsep}{10pt}
    \renewcommand{\arraystretch}{1.1}
    \begin{tabular}{ccc}
        \legenditem{medqacol}{MedQA} &
        \legenditem{gpqacol}{GPQA} &
        \legenditem{boolqcol}{BoolQ} \\
        \legenditem{stratqacol}{StratQA} &
        \legenditem{gsm8kcol}{GSM8K} &
        \legenditem{aimecol}{AIME'26}
    \end{tabular}
    \end{center}

    \caption{\textbf{GRCS consensus and mode-collapse frontiers} across all
    evaluated benchmarks (\cf, \Cref{sec:exp_setup}), averaged over five
    independent runs. The dark lines show the mean GRCS across all sampled
    questions and runs at each fraction, while the light bands indicate
    $\pm 1$ standard deviation.}
    \label{fig:grcs_sweeps}
\end{figure}

In our example, the SS-GED distances from Step~4 have mean
$\mu(\mathfrak{D}(q))=0.59$, and yield
$\mathrm{GRCS}=0.63$. This value reflects the limited semantic-structural consensus of the GLLM's
reasoning: although $G_0$ and $G_1$ are consistent, the remaining graphs
produce substantially inconsistent reasoning.

\subsection{Graph Self Consistency}\label{sec:consensus_context_synthesis}
%\noindent {\bf Consensus Context Synthesis%Graph Self-Consistency (GSC)
%}. %We introduce \textbf{Graph Self-Consistency} (GSC) as a downstream, practical application of our $\pname$ framework for decoding. 

%To transition from graph analysis to practical downstream decoding, this step introduces 
To create a decoding rule out of the reasoning graphs, this step introduces 
Graph Self-Consistency (GSC).
SC uses majority voting over sampled outputs~\cite{wang2023selfconsistency}, but this can reward correct yet unfaithful answers, termed \emph{lucky guesses}~\cite{lyu2023faithful}. %We therefore introduce \textbf{Graph Self-Consistency} (GSC), which, 
The primary cause of this behavior is that SC is blind to a GLLM's underlying reasoning. Evaluating intermediate reasoning steps rather than only final answers has recently been shown to improve reasoning reliability~\cite{lightman2024verify}. To fill this gap, we present GSC decoding, which, among all GLLM's sampled reasoning paths, selects the \emph{medoid CoT}, \ie, the CoT that is most central and supported in the semantic-structural graph manifold at the step level by the rest of the pool. With pairwise SS-GED similarities $S(G_i,G_j)=1-d(G_i,G_j)$, we define Structural Centrality for each $G_i$ as
$
C(G_i)=\sum_{j\neq i} S(G_i,G_j).
$
%For local support, 
%To measure the cross-graph support, 
Consider node $v$ of graph $G_i$. Let $c(v)$ denote the fraction of other graphs containing at least one semantically similar node to $v$ (cosine similarity $\geq 0.9$). We then define graph-level consensus score (\ie, support of graph $G_i$ among its peers) as:
\begin{equation}
P(G_i)=\frac{\sum_{v\in V_i} w(v)\,c(v)}{\sum_{v\in V_i} w(v)},
\end{equation}
where $w(v)=2$ for terminal nodes and $w(v)=1$, otherwise, as %, assigning unsupported final conclusions twice the weight of unsupported intermediate steps (
terminal nodes are more critical since %, as 
they %more 
directly determine the final answer. %). 
The selected medoid is therefore:
%$
%\text{Medoid}
\begin{equation}
    z^*=\arg\max_{G_i}\left[\alpha \frac{C(G_i)}{\max_j C(G_j)}+(1-\alpha)P(G_i)\right]
\end{equation}
%z^*=\arg\max_{G_i}\left[\alpha \frac{C(G_i)}{\max_j C(G_j)}+(1-\alpha)P(G_i)\right]
%$,
%with $\alpha=0.8$. 
where $\alpha$ denotes the \textit{trade-off} between centrality and support. % in the medoid selection. %Its final answer is returned as the prediction. 
%See Appendices~\ref{sec_gsc_detailed_appendix} and~\ref{sec:alpha_sensitivity_appendix} for full GSC definition and $\alpha$ sensitivity analysis.
% \paragraph{Quantifying Uncertainty}\label{par_uq_methodology}
%\rreva{put running example}
%%% DP: Please provide a gap between successive paras in the editor. Else hard to find.

In our running example, SC considers only final answer frequency, thus it selects the \textbf{incorrect} answer \emph{C} (supported by three CoTs out of five). GSC instead evaluates the reasoning graphs. The low SS-GED between $G_0$ and $G_1$ ($0.10$) gives $G_1$ the highest structural centrality, $C(G_1)=2.30$, and its steps also receive the highest graph-level support, $P(G_1)=0.25$. Combining both terms gives $G_1$ the highest GSC score among all reasoning graphs, $0.85$. GSC therefore selects CoT~\#1 as the medoid and returns the \textbf{correct} answer \emph{D}. \textbf{This %result 
shows GSC can select the most supported reasoning even when SC's majority vote is incorrect.}
%\smallskip
%\smallskip

\begin{lstlisting}[
  style=promptstyle,
  caption={GLLM Adversarial Attack Prompt},
  label={lst:ADV_prompt},
  emph={CRITICAL,CONSTRAINT,STRICTLY,FORBIDDEN,COMPLETELY,ALTERNATIVE,depends_on,Question,final,answer}
]
You are a precise causal reasoner.

CRITICAL CONSTRAINT: You are strictly forbidden from using the following
reasoning path or logical steps:
<forbidden_reasoning>
{MEDOID_TEXT}
</forbidden_reasoning>

You must find a completely alternative, valid reasoning path from scratch.
Do not reuse or copy the forbidden logic.
- Use 3--6 numbered declarative steps, starting from 0.
- Append [depends_on: N, ...] or [depends_on: none] to each step.
- Output only the steps and final answer. End with: The final answer is: [answer]
Question: {QUESTION}
\end{lstlisting}

\subsection{GSC Evaluation via Adversarial Medoid Ablation}\label{sec:faithfulness_adv_medoid_ablation}

% \noindent {\bf Faithfulness Evaluation via Adversarial Medoid Ablation
% %Adversarial Medoid Ablation
% }.\label{par:adv_ablation} 

To assess the importance of the medoid CoT selected by GSC, we design a prompt-poisoning attack that directly suppresses it. 
% For each question $q$, we retrieve the medoid CoT $z^*$ %selected by GSC 
% and inject it into a constrained generation prompt via an adversarial perturbation $\delta_{\text{ablate}}$. The poisoned prompt $x^\prime = x \cup \delta_{\text{ablate}}$ explicitly forbids the model from reusing the medoid's logic~\cite{zou2023universal, wang2023adversarial}, instructing it to produce a \textit{completely alternative} reasoning chain from scratch by sampling from the restricted reasoning space given by %distribution
% $%\widehat{z}_{\text{alt}} \sim 
% P(z \mid x^\prime,\, z \neq z^{*})$. 
For each question \(q\), let \(x_q\) be the original prompt and \(z_q^\star\) the medoid CoT selected by GSC. We build an adversarial perturbation \(\delta_{\mathrm{ablate}}(z_q^\star)\) that presents \(z_q^\star\) in full and explicitly forbids reusing any of its reasoning steps or conclusions~\cite{wang2023adversarial}. The resulting poisoned prompt, \(x_q^\mathrm{poisoned} = x_q \cup \delta_{\mathrm{ablate}}(z_q^\star)\), % (\(\cup\): prompt concatenation), % steers the model away from \(z_q^\star\) and forces a fully alternative CoT.
forces the GLLM to produce a fully alternative CoT.
We then sample $N$ alternative CoTs at the original temperature $T$. \Cref{lst:ADV_prompt} summarizes our adversarial prompt. % we used. 

In our example, we would include CoT~\#1 in the adversarial prompt and forbid the GLLM from reusing that reasoning path. % and compare their average accuracy and faithfulness with those of $z_q^\star$.

\section{Analytical Comparison %Supremacy 
of Topology-Based Decoding over Outcome Voting}\label{sec:gsc_proof}

We now formalize and analytically establish why GSC preserves information that SC discards. We first introduce three %foundational 
Lemmas progressively introducing %(\Cref{lem:info_loss,lem:reasoning_diversity,lem:info_recovery}) that provide the 
mathematical tools needed to construct and prove the main %state our main 
Theorem (\Cref{thm:supremacy}).

Let $\mathcal{Z}$ be the space of all CoTs elicited from a GLLM, and let $\mathcal{Y}$ be the space of all possible
final answers. Let $Z \in \mathcal{Z}$ and $Y \in \mathcal{Y}$ be discrete random
variables representing a sampled reasoning path and its corresponding final
answer generated by a GLLM.
The extraction of a final answer from a CoT is a
deterministic, %many-to-one 
surjective mapping
$\operatorname{Ans}: \mathcal{Z} \to \mathcal{Y}$, such that $Y = \operatorname{Ans}(Z)$. In our methodology, $\operatorname{Ans}$ is a string parser function that extracts the tokens of the final answer from the GLLM's output.
For any specific final answer $y \in \mathcal{Y}$, %there exists an equivalence
%class 
the set
of %reasoning paths 
CoTs that converge %to 
on it %can be defined as
is $\mathcal{Z}_y = \{ z \in \mathcal{Z} \mid \operatorname{Ans}(z) = y \}$.

%\vspace{3mm}
We can now define the Lemmas:

\begin{lemma}[{\bf Entropy Decomposition under Deterministic Answer Extraction}]
\label{lem:info_loss}
The uncertainty (Shannon Entropy~\cite{Shannon1948}) of the final answer $Y$ is upper bounded by the uncertainty of the reasoning path $Z$, such that $H(Y) \leq H(Z)$. The difference between their respective entropies is equal to the conditional entropy of the reasoning path given the final answer, $H(Z|Y)$.
\end{lemma}

\begin{IEEEproof}
Using the chain rule for Shannon Entropy, the joint entropy of the reasoning path $Z$ and the final answer $Y$ is
$H(Z, Y) = H(Z) + H(Y | Z)$. 
Because %the final answer 
$Y$ is entirely and deterministically derived from %the reasoning path 
$Z$ via the mapping $Y = f(Z)$, there is no uncertainty in $Y$ once $Z$ is known. Therefore, $H(Y | Z) = 0$. Substituting this into the equation yields:
$H(Z, Y) = H(Z)$. 
By symmetry of the chain rule, the joint entropy can also be expressed as:
$H(Z, Y) = H(Y) + H(Z | Y)$. 
Equating the two expressions for joint entropy gives:
$H(Z) = H(Y) + H(Z | Y)$. 
Because $Z$ and $Y$ are discrete random variables representing finite text sequences ($\ie$, the CoT and its final answer), their Shannon entropy is non-negative ($H(Z|Y) \geq 0$)~\cite{cover1999elements}. %It therefore %directly 
%follows that 
Therefore: $\mathbf{H(Y) \leq H(Z)}$.
\end{IEEEproof}

\begin{lemma}[{\bf Conditional Entropy of Reasoning Paths}]
\label{lem:reasoning_diversity}
If the GLLM generates multiple reasoning paths that all converge on the same final answer $y$ but are logically distinct %($\eg$, 
due to alternative logical deductions or hallucinated steps, %), 
then the %hidden 
conditional entropy $H(Z | Y=y)$ is strictly greater than zero.
\end{lemma}

\begin{IEEEproof}
The conditional entropy of the reasoning space given a specific answer $y$ is: % defined as:
$H(Z | Y=y) = -\sum_{z \in \mathcal{Z}_y} P(Z=z | Y=y) \log P(Z=z | Y=y)$.

%In probability theory, the sum of all probabilities for a given outcome must equal exactly 1. %Because 
Given that $\sum_{z \in \mathcal{Z}_y} P(Z = z \mid Y = y)=1$, 
if the GLLM samples at least two distinct reasoning paths $z_1, z_2 \in \mathcal{Z}_y$ %(whether through valid alternative logic or unfaithful hallucination), 
(%in practice, 
%this can happen 
as can occur under stochastic %decoding 
sampling with non-zero temperature~\cite{wang2023selfconsistency}) %, which samples different continuations from a non-degenerate token distribution)
the %total ù
unit probability mass %of 1 is %necessarily 
is distributed %across multiple %states. 
%reasoning 
among these paths. Consequently, the probability of any single path lies strictly in the open interval $(0, 1)$. Since $\log(P) < 0$ for all $P \in (0, 1)$, every non-zero term in the summation is strictly positive, yielding $\mathbf{H(Z | Y=y) > 0}$.
\end{IEEEproof}

\begin{lemma}[{\bf Retention of Conditional Reasoning Uncertainty in the Graph}]
\label{lem:info_recovery}
Let $\Gamma:\mathcal{Z}\to\mathcal{G}$ be the deterministic mapping that transforms each reasoning path $Z$ into a DAG ($\cf$, \Cref{sec:methodology}), and $G=\Gamma(Z)$ be the %graph-valued random variable induced by the
random graph %induced %by the random reasoning path 
%by 
associated to $Z$.
%where each graph contains %the 
Each realization $g\in\mathcal{G}$ contains 
%its corresponding 
a final
answer %$Y=\operatorname{Ans}(Z)$ 
$y=\operatorname{Ans}(z)$ as %a 
designated leaf. Then, %we can define the entropy of $G$ as:
the entropy of $G$ decomposes as:
$
H(G)=H(Y)+H(G\mid Y),
$
and $H(G\mid Y)$ satisfies:
$
0\leq H(G\mid Y)\leq H(Z\mid Y).
$
%Thus, $H(G\mid Y)$ %quantifies 
%measures the conditional reasoning uncertainty
%retained by the graph %representation 
%after conditioning on the final
%answer.
Thus, $G$ retains the within-answer reasoning uncertainty %$H(G\mid Y)$ 
preserved by $\Gamma$.
\end{lemma}
\begin{IEEEproof}
%Let $Z$ be an elicited reasoning path, let $Y=\operatorname{Ans}(Z)$ be its final
%answer, and let $G=\Gamma(Z)$ be the deterministic graph representation
%constructed following \Cref{sec:methodology}. Each graph $G$ contains a designated terminal node that contains the final answer $Y$. Since $Z$ is a random variable and $\Gamma$ is
%deterministic, $G$ is a discrete random variable over the graph space
%$\mathcal{G}$. We can define its Shannon Entropy as:
%$
%H(G)= -\sum_{g\in\mathcal{G}}P(G=g)\log P(G=g),
%$
%which measures uncertainty \cite{Shannon1948} over the semantic-structural graph
%representations induced by the reasoning distribution. Each graph $G$ contains a designated terminal node whose label is the
%final answer $Y$. Hence, $Y$ is deterministically recoverable from
%$G$. 
$Y$ is deterministically recoverable
from $G$ through $\Gamma$. Hence, $H(Y\mid G)=0$. We %can 
apply the chain rule: % and write:
$H(G,Y)=H(G)+H(Y\mid G)=H(G),$
and $H(G,Y)=H(Y)+H(G\mid Y).$
%We can thus write:
Therefore:
$
H(G)=H(Y)+H(G\mid Y),
$
where  $H(G\mid Y)$ measures the uncertainty \cite{Shannon1948} among graph
representations %that remains 
retained after conditioning on $Y$. %Moreover:
Since $G=\Gamma(Z)$ is a deterministic function of $Z$, $H(G\mid Z,Y)=0.$
We apply the conditional chain rule:
$H(Z\mid Y)=H(G\mid Y)+H(Z\mid G,Y),$
where $H(Z\mid Y)$ is the uncertainty among CoTs %that produce 
with the same final answer.
Because $Z$ is discrete, conditional entropy is non-negative: $H(Z\mid G,Y)\geq 0$.
Therefore:
%It satisfies:
$
0\leq H(G\mid Y)\leq H(Z\mid Y).
$
%where $H(Z\mid Y)$ is the uncertainty among CoTs %that produce 
%with the same final answer. Because $G=\Gamma(Z)$ is a deterministic representation of
%$Z$, it cannot retain more uncertainty than the original CoTs. In other words:
%$
%H(Z\mid Y) = H(G\mid Y)+H(Z\mid G,Y)
%$ with $H(Z\mid G,Y)\geq 0$.
Moreover, for any answer $y$ with probability $P(Y=y)>0$, if $P(G\mid Y=y)$ assigns positive probability to at least two distinct graphs, %representations
then:
$
H(G\mid Y=y)>0.
$
%As result, the graph %representation 
%retains the component of conditional
%reasoning uncertainty that remains distinguishable after mapping
%CoTs into graphs.
Thus, $G$ retains the within-answer reasoning uncertainty %$H(G\mid Y)$ 
preserved by $\Gamma$.
\end{IEEEproof}

These Lemmas let us define the Theorem:

\begin{theorem}[{\bf Conditional Information Retained by Topology-Based Decoding}]
\label{thm:supremacy}

Let Self-Consistency (SC) be a decoding strategy that selects $Y$ solely from its empirical marginal distribution, such that SC is invariant to variations in $Z$ for a fixed $Y$. Let Graph Self-Consistency (GSC) be a decoding strategy that operates on the graph $G = \Gamma(Z)$, which captures the conditional graph-level uncertainty $H(G \mid Y)$. If %the %structural mapping 
$\Gamma$ bounds the graph-level uncertainty such that: $0 \le H(G \mid Y) \le H(Z \mid Y)$, then SC is strictly blind to the within-answer reasoning uncertainty $H(Z \mid Y)$, whereas GSC measures the intermediate structural uncertainty retained by $\Gamma$, thereby providing a %strictly 
richer characterization of uncertainty than SC.
%SC selects the final answer solely from its empirical final-answer distribution and it is invariant to changes among reasoning paths that have the same final answer. Consequently, SC %does not use 
%is blind to the within-answer reasoning uncertainty measured \cite{Shannon1948} by $H(Z\mid Y)$. GSC operates on the graph representations $G=\Gamma(Z)$ and instead considers the conditional graph-level uncertainty measured \cite{Shannon1948} by $H(G\mid Y)$ where $0\leq H(G\mid Y)\leq H(Z\mid Y)$. Thus, GSC provides a richer characterization of uncertainty than SC.
\end{theorem}

\begin{IEEEproof}
    SC aggregates results via majority voting over the space of final answers $\mathcal{Y}$~\cite{wang2023selfconsistency}. Therefore, its decision depends only on the empirical distribution of the final answers, %$\widehat{P}(Y)$. 
    Any change to a reasoning path $z_i$ that preserves its
    final answer, $\operatorname{Ans}(z_i^\prime)=\operatorname{Ans}(z_i),$ leaves the answer frequencies, and hence the SC decision, unchanged. Thus, SC is invariant to reasoning variation. % within each answer class.
    From \Cref{lem:info_loss}, the entropy of the reasoning $Z$ decomposes as:
    $
    H(Z)=H(Y)+H(Z\mid Y),
    $
    where $H(Z\mid Y)$ measures the uncertainty %\cite{Shannon1948} 
    among CoTs that produce the same final answer. Lemma \ref{lem:reasoning_diversity} demonstrates that, %whenever the reasoning manifold exhibits diversity ($|\mathcal{Z}_y| > 1$), $H(Z|Y)$ is strictly positive. 
    for any answer $y$ with $P(Y=y)>0$, if the conditional distribution $P(Z\mid Y=y)$ assigns positive probability to at least two distinct reasoning paths,
    $
    \left|\operatorname{supp} P(Z\mid Y=y)\right|>1,
    $
    then $H(Z\mid Y=y)>0$. Therefore, within-answer reasoning uncertainty can be non-zero even when the final answer is fixed. SC does not assume that $H(Z\mid Y)=0$ or explicitly
    estimate $H(Y)$ but is insensitive to the variation represented by $H(Z\mid Y)$. By \Cref{lem:info_recovery}, the graph $G=\Gamma(Z)$ %representation 
    used by GSC satisfies:
    $H(G)=H(Y)+H(G\mid Y)$ and $0\leq H(G\mid Y)\leq H(Z\mid Y).$
    Therefore, GSC can distinguish reasoning manifolds that are
    the same to SC by using the conditional reasoning information
    retained in $H(G\mid Y)$ by $\Gamma$. Thus, GSC captures %strictly 
    more information than SC whenever $\mathbf{H(G\mid Y) > 0}$.
\end{IEEEproof}
\section{Experimental Setup and Evaluation}
\label{sec:exp_setup}

%\rreva{in paper, change all "StrategyQA" to "StratQA"}
\iffalse
\StQ{You have already may references, but this is one potentially missing: evaluating intermediate reasoning steps rather than only final answers has recently been shown to improve reasoning reliability~\cite{lightman2024verify}.}
\rreva{Thanks! Done in Methodology when introducing GSC}
\fi

\noindent \textbf{Datasets}. %We evaluate our framework across diverse reasoning tasks of varying complexity. The selected benchmark datasets serve as domain-specific examples of these tasks, spanning commonsense, mathematical, clinical, and scientific domains, and covering a spectrum from moderate to expert-level difficulty:
%We evaluate our framework across reasoning tasks of increasing complexity, spanning mathematical, commonsense, clinical, and scientific domains:
We evaluate across reasoning tasks of increasing %cognitive 
reasoning complexity: % in mathematical, reading-comprehension, commonsense, clinical, and scientific domains:
1) \textbf{GSM8K}: %A dataset of 
grade-school arithmetic %word 
problems~\cite{cobbe2021training}; %, %This enabled 
%enabling methodological comparison with %existing 
%SOTA baselines~\cite{tanneru2024quantifying,da2025understanding}. % and high susceptibility to mode collapse.
2) \textbf{BoolQ}: yes/no reading-comprehension~\cite{clark2019boolq}; %questions requiring short entailment-style reasoning.
3) \textbf{StrategyQA} (StratQA for brevity): %229 questions of open-domain, commonsense multi-hop boolean reasoning. This contrasts GPQA by providing shorter, non-expert reasoning chains. 
%A benchmark of 
open-domain %, multi-hop 
planning questions~\cite{geva2021aristotle}; % with boolean answers; %, requiring %short chains of 
%compositional reasoning.
%5) \textbf{BBH Causal Judgement}~\cite{suzgun2022challenging}: deductive tasks from the BIG-bench Hard benchmark evaluating causal explanation. %and yes/no decision consistency under shortcut vulnerability.
4) \textbf{MedQA}: %USMLE}: %100 randomly subsampled questions (seed 42) of clinical multi-step reasoning. This grounds the framework in a high-stakes domain where reasoning coherence is a critical priority. 
 %clinical reasoning benchmark derived from U.S. medical licensing exams. We evaluate on 100 randomly sampled questions (seed 42).
 clinical reasoning benchmark from U.S. medical licensing exams; 100 randomly sampled questions (seed 42)~\cite{jin2021whatdisease};
%4) \textbf{GSM8K}~\cite{cobbe2021training}: The full dataset of grade-school arithmetic word problems. This enabled methodological comparison with existing baselines~\cite{da2025understanding} \rreva{put other existing works, look at topo uq refs}. Its structured numerical graphs provide a controlled contrast to the open-ended causal graphs of GPQA.
5) \textbf{GPQA}: %Diamond}: %198 questions requiring expert-level, multi-step causal reasoning in biology, chemistry, and physics. This serves as our primary showcase dataset, as no prior UQ framework evaluates on this benchmark. 
 %challenging benchmark of 198 graduate-level questions in biology, chemistry and physics requiring deep 
%causal reasoning. 
challenging graduate-level biology, chemistry, and physics benchmark~\cite{rein2024gpqa}; and %requiring deep causal reasoning.
%For additional evaluations on much longer horizon reasoning benchmarks, refer to \Cref{sec:long_horizon_reasoning,sec:scalability_extended_cots}. \rreva{pur AIME here instead of putting it in the appendix}
6) \textbf{AIME'26}: %high-school competition-level mathematics presenting significantly longer-horizon reasoning paths~\cite{dekoninck2026beyond}.
long-horizon reasoning benchmark based on multi-step competition-level mathematics~\cite{dekoninck2026beyond}.

\smallskip

\smallskip

\noindent \textbf{Models}.
%\paragraph{LLMs Under Study}
%To generate the CoTs evaluated in our framework, we selected %four 
%three diverse open-weight models: Meta-Llama-3.1-8B-Instruct~\cite{grattafiori2024llama3}, %Meta-Llama-3.3-70B-Instruct~\cite{meta_llama33_model_card_2024}, 
%Microsoft Phi-4~\cite{abdin2024phi4}, and DeepSeek-R1-Distill-70B   ~\cite{guo2025deepseek, deepseek_r1_docs_2025}.
%This selection spans different parameter scales and training paradigms to stress-test our methodology. Llama 3.1 (8B) %and Llama 3.3 (70B) 
%serves as general-purpose instruction-tuned baseline. %at small and large scales. 
%Phi-4 %represents a mid-size 
%is a 14B-parameter, mid-sized model optimized for mathematical and logical reasoning, while DeepSeek-R1 (70B) represents a larger-scale, distilled, RL-enhanced reasoning paradigm within a dense architecture.
%By relying exclusively on open-weight models, we ensure transparency and enable deterministic reproduction of %the reasoning topologies and uncertainty metrics 
%results reported in this work.
We study three %open-weight 
GLLMs: Llama 3.1 8B~\cite{grattafiori2024llama3}, Phi-4 (14B)~\cite{abdin2024phi4}, and DeepSeek R1 (Distill-Llama-70B)~\cite{guo2025deepseek}, %\rreva{put exactly why you chose these models} 
We specifically choose these GLLMs to represent general-purpose, reasoning-specialized, and distilled RL-enhanced paradigms, respectively, and to evaluate $\pname$ across different model sizes.
%We choose these GLLMs because we want to %spanning 
%span general-purpose, reasoning-specialized, and distilled RL-enhanced paradigms, respectively, and test how $\pname$ behaves with different model sizes. %This diversity stress-tests our framework while ensuring transparency and reproducibility.
%Relying on open-source GLLMs ensures transparency and reproducibility.
%For DLLM, we rely on OpenAI’s GPT-4o as a high-fidelity fact extractor 
%\cite{openai2024gpt4o}. 
While we rely on open-source GLLMs to ensure the transparency and reproducibility, we use OpenAI’s GPT-4o~\cite{openai2024gpt4o} as DLLM to achieve high-fidelity decomposition. %For a complete ablation study on the DLLM, refer to \Cref{sec:dllm_ablation_sec}.
We use Qwen3-Embedding-0.6B~\cite{yang2025qwen3technicalreport} as embedding Transformer, %specifically 
choosing a distinct model family to ensure architectural independence with GLLMs/DLLM.

\smallskip

%\smallskip
%\rreva{update this}
\noindent \textbf{Baselines}. For UQ, %(\Cref{sec_rq1_all}), 
we benchmark %our GRCS %metric %our GRCS metric 
%GRCS 
the GRCS-derived uncertainty proxy
$\mathcal{U}_{\mathrm{GRCS}}$
against CoTA~\cite{tanneru2024quantifying} and Topo-UQ~\cite{da2025understanding}\footnote{For both baselines, we use the authors' implementations, adapted to our benchmarks (Topo-UQ: \url{https://github.com/LongchaoDa/LLM-Topology}, commit \texttt{d4d837bf7123c5d67b3289987ce8dc70f57b1141}; CoTA: \url{https://github.com/harsha070/uncertainty-quantification-nle}, commit \texttt{1c4f1a577644999d1a9281698a69e3ad0d60c53d}).}. %, and two statistical metrics adapted to our topological space: dispersion and entropy (used in~\cite{catak2024uncertainty, rajamohan2025ensemble}). %in the broader cross-domain experiment of increasing complexity, where prior baselines are unavailable 
%Dispersion ($\sigma(\mathfrak{D}(q))$) is the standard deviation of the pairwise SS-GED distances, capturing overall spread. Entropy ($H(\mathfrak{D}(q))$) is the Shannon entropy~\cite{Shannon1948} of the distance distribution (bins determined via the Freedman–Diaconis rule~\cite{freedman1981histogram}), %which 
%capturing distributional complexity.
Additionally, we compare their computational cost and energy efficiency by measuring GPU runtime for processing a single CoT and the full set of $N$ CoTs per question, together with energy consumption (J), peak GPU utilization (\%), and peak GPU memory usage (MB). We measure GPU runtime as elapsed GPU processing time, and profile GPU energy, memory and utilization using NVIDIA\textsuperscript{\textregistered} NVML library%(\texttt{pynvml}
~\cite{nvidia_nvml}. %, and measure peak GPU utilization and memory using PyTorch CUDA statistics~\cite{paszke2019pytorch}.  
Since all methods use the same set %response set
elicited CoTs, generation time is excluded. %We also report energy consumption (J), peak GPU usage (\%), and peak GPU memory usage (MB). 
All measurements are collected on the same hardware and GPU across five independent runs, with the mean and standard deviation reported.
For task accuracy, we compare GSC against SC~\cite{wang2023selfconsistency} to test if it improves reasoning fidelity beyond majority voting\footnote{SC is implemented directly as majority voting over answers, following Wang et al.~\cite{wang2023selfconsistency}}.

\smallskip
%\rreva{put before evaluation, gpu memory}
\noindent \textbf{Hardware Used}. To execute %both 
$\pname$ %pipeline and %the 
and baselines, we use %\StQ{2x - two ?} 
an Intel\textsuperscript{\textregistered} Xeon\textsuperscript{\textregistered} w9-3495X CPU (56 cores, %112 threads, 
1.9 GHz) with 1 TB %512 GB 
DDR5 RAM and two NVIDIA\textsuperscript{\textregistered} RTX\textsuperscript{\textregistered} 6000 Ada GPUs (96 GB VRAM in total). GLLM and DLLM inference, embedding, and graph creation are %performed 
executed on GPU.

\smallskip

\noindent {\bf $\mathcal{U}_{\mathrm{GRCS}}$ Evaluation}.
%To determine whether GRCS captures model uncertainty, 
To determine whether the GRCS-derived uncertainty proxy
$\mathcal{U}_{\mathrm{GRCS}}$ captures reasoning uncertainty,
we evaluate it against faithfulness, which measures if the reasoning genuinely supports the final answer~\cite{lanham2023measuring,turpin2023language}.
%using two %complementary, independent approaches. %These measures are specifically chosen to expose ``lucky guesses'' and hallucinations where the generated reasoning path does not logically drive the final output. 
Faithfulness and uncertainty are deeply intertwined in LLMs, as unfaithful reasoning reflects epistemic uncertainty because the output emerges from hallucinations or \emph{lucky guesses} rather than verifiable deduction~\cite{lanham2023measuring, turpin2023language, tanneru2024quantifying}. %This uncertainty manifests either as highly divergent reasoning paths (high entropy) or as confident but flawed mode collapse~\cite{hamilton2024detectingmodecollapselanguage}.
\textbf{Therefore, an %ideal 
accurate UQ metric should increase when a model's faithfulness decreases}~\cite{tanneru2024quantifying,da2025understanding}.
We %compute 
compute faithfulness using two independent approaches to ensure robustness. First, we apply Early Answering (EA)~\cite{tanneru2024quantifying}, iteratively prompting GLLMs with expanding medoid CoT facts. Reasoning is unfaithful if the answer emerges prematurely, and faithful if the full CoT is required. This %directly 
exposes unfaithful reasoning by revealing when a model arrives at the final answer before its logical deduction is complete. %To ensure robustness against prompt-truncation artifacts and to introduce formal causal rigor, 
Second, we also compute faithfulness via Counterfactual Editing (CFE), grounded in Rung $3$ (Counterfactuals) of Pearl's Ladder of Causation~\cite{pearl2009causality}.
We intervene at each intermediate step by prompting a distinct LLM (Mistral-Large $2512$~\cite{mistral2025introducing}; this is done to avoid self-evaluation bias) to generate a counterfactual, keeping all other steps fixed to observe the effect on the outcome. Prior to evaluation, we validated the counterfactuals generation by comparing human annotations against an LLM-as-judge (Gemini 3 Pro~\cite{gemini2025family}) on $100$ sampled steps (seed $42$), achieving a strong Cohen's $\kappa$ of $0.81$~\cite{cohen1960kappa}. A step is counted as causally faithful if altering it changes the final answer (proving the model actively relied on that premise), and unfaithful if the answer is unchanged (proving the premise was ignored). \textbf{While EA tests whether the final answer depends on the complete reasoning sequence, CFE isolates the causal contribution of individual logical steps.}
For both methods, we compute Pearson Correlation Coefficient (PCC), Spearman Rank Correlation (SRC), and Kendall Rank Correlation (KRC) between the UQ metric and the faithfulness scores. %~\cite{pearson1895regression, spearman1904proof, kendall1938new}.
A strong negative correlation indicates an accurate UQ metric~\cite{tanneru2024quantifying, da2025understanding}.
%Because SRC and KRC exhibit trends consistent with PCC across all evaluations, we report only PCC for brevity and provide the full SRC and KRC %tabular 
%results in our public repository\rreva{put link of it}.
%Faithfulness and uncertainty are deeply intertwined in LLMs~\cite{lanham2023measuring, turpin2023language, tanneru2024quantifying}. Unfaithful reasoning reflects epistemic uncertainty because the output emerges from opaque latent biases or ``lucky guesses'' rather than verifiable deduction. This uncertainty manifests either as highly divergent reasoning paths (high entropy) or as confident but flawed mode collapse~\cite{hamilton2024detectingmodecollapselanguage}.
%\bem{Because an ideal uncertainty metric should increase when a model's reasoning faithfulness decreases, a strong negative correlation indicates an accurate UQ metric~\cite{tanneru2024quantifying, da2025understanding}.}
%Therefore, an %ideal 
%accurate UQ metric should increase when a model's reasoning faithfulness decreases. \bem{Thus, a strong negative correlation indicates an accurate UQ metric~\cite{tanneru2024quantifying, da2025understanding}}.
To minimize sampling bias, we estimate correlations via $1,000$ bootstrap resamples per dataset ($20$ questions, $20$ responses each).

\smallskip
%\rreva{detail better why we did this. What was the need. Then, detail it better from a statistical point. detail the strength}

\noindent \textbf{GSC Evaluation}.\label{sec:audit_methodology} %For GSC--SC comparisons, we focus on the disagreement cases (\textit{SC-only wins}, \textit{GSC-only wins}). To prevent %prevent 
%evaluator bias, for each (model, dataset) pair, we sample up to $N=50$ anonymized traces per category and label them as \textit{Robust} or \textit{Lucky Guess}. A human annotator (one of the authors) and Gemini~3~Pro~\cite{gemini2025family} independently annotate the subset using the same rubric~\cite{gu2026llmjudge}, and then Cohen's $\kappa$~\cite{cohen1960kappa} is computed to estimate the agreement. Then, the LLM judge is applied to the full disagreement set. To assess the statistical significance of our results we apply the Wilcoxon signed-rank test~\cite{wilcoxon1945} to the paired proportions of faithful wins.
To compare GSC against SC, we use three approaches. First, we measure the task accuracy differences, $\Delta(\text{GSC} - \text{SC})$, and assess significance using McNemar's test~\cite{mcnemar1947}. %, using McNemar's test~\cite{mcnemar1947} for statistical significance. 
Second, we empirically measure which correct SC answers GSC discards as a function of conditional graph uncertainty, $H(G\mid Y)$, by grouping answers into quartiles from low uncertainty (Q1) to high uncertainty (Q4).
Third, we assess reasoning fidelity by auditing disagreement cases (\textit{SC-only} vs.\ \textit{GSC-only wins}).
We define \emph{Deductive Soundness} (DS) as a binary evaluation criterion. For each node $v$ of a CoT $z$ with graph $G=\Gamma(z)$, let
$\operatorname{pa}(v)=\{u\in V:(u,v)\in E\}$ be the set of logical
logical antecedents of $v$ annotated by the GLLM, and let $\mathcal{R}=\{r\in V:\operatorname{pa}(r)=\varnothing\}$ denote the premise nodes. We define
%\begin{equation}
%    \operatorname{DS}(z)=
%    \mathbf{1}\!\left[
%    \left(\bigwedge_{r\in\mathcal{R}}\operatorname{True}(r)\right)
%    \land
%    \left(
%    \bigwedge_{v\in V\setminus\mathcal{R}}
%    \left(\bigwedge_{u\in\operatorname{pa}(v)}u\right)\models v
%    \right)
%    \land
%    \left(\bigwedge_{v\in V}v\right)\models\operatorname{Ans}(z)
%    \right].
%\end{equation}
%
\begin{equation}
\begin{aligned}
\operatorname{DS}(z)
= \mathbf{1}\!\Bigg[
&\left(
    \bigwedge_{r\in\mathcal{R}}
    \operatorname{True}(r)
\right)
\\
&\land
\left(
    \bigwedge_{v\in V\setminus\mathcal{R}}
    \left[
        \left(
            \bigwedge_{u\in\operatorname{pa}(v)}u
        \right)
        \models v
    \right]
\right)
\\
&\land
\left[
    \left(
        \bigwedge_{v\in V}v
    \right)
    \models \operatorname{Ans}(z)
\right]
\Bigg].
\end{aligned}
\end{equation}
DS is $1$ iff all premises are factually correct, every derived
fact follows from its declared antecedents without logical gaps, and
the complete reasoning entails the final answer. Otherwise, it is $0$.
% to determine effectivity of GSC as a filter %if GSC effectively filters out unfaithful ``lucky guesses'' 
We perform this analysis via a human annotator and Gemini~3~Pro~\cite{gemini2025family} as an LLM judge, and assess their agreement using Cohen's $\kappa$~\cite{cohen1960kappa}, and test significance using the Wilcoxon signed-rank test~\cite{wilcoxon1945}.
% to compare their faithful-win rates~\cite{wilcoxon1945}.
%For both stages, we perform test for statistical significance using McNemar’s test to compare answer correctness~\cite{mcnemar1947}, and the Wilcoxon signed-rank test to compare their faithful-win rates~\cite{wilcoxon1945}. %In both cases $p<0.05$ indicates significance.
% To prevent bias, a human annotator (one of the authors) and Gemini~3~Pro~\cite{gemini2025family} independently annotate $N=50$ anonymized traces per (GLLM, dataset) pair as \textit{Robust} (the correct answer is supported by a logically faithful reasoning path) or \textit{Lucky Guess} (the correct answer is reached via flawed, hallucinated, or unfaithful reasoning) following the LLM-as-a-judge methodology~\cite{gu2026llmjudge}. After validating inter-rater reliability via Cohen's $\kappa$~\cite{cohen1960kappa}, we deploy the LLM judge across the full disagreement set. To assess the statistical significance of our results we apply the Wilcoxon signed-rank test~\cite{wilcoxon1945} to the paired proportions of faithful wins.

\smallskip

\noindent \textbf{GSC Medoid-Ablation Evaluation}.
To quantify how suppressing the GSC medoid affects correctness and reasoning fidelity, we measure $\Delta\mathrm{Accuracy}$ and $\Delta\mathrm{Faithfulness}$ (measured by EA~\cite{tanneru2024quantifying}). Let $\mathrm{Acc}(z)=\mathbf{1}[\operatorname{Ans}(z)=y_q]$ and $\phi(z)\in[0,1]$ denote the faithfulness of CoT $z$. We define
$\Delta\mathrm{Accuracy}_q=\frac{1}{N}\sum_{k=1}^{N}\mathrm{Acc}(z_{q,k}^{\mathrm{adv}})-\mathrm{Acc}(z_q^\star)$ and
$\Delta\mathrm{Faithfulness}_q=\frac{1}{N}\sum_{k=1}^{N}\phi(z_{q,k}^{\mathrm{adv}})-\phi(z_q^\star)$ ($\ie$, $\Delta\text{Accuracy}$ and $\Delta\text{Faithfulness}$ are computed as \emph{ablated} minus \emph{original)}.
Negative values indicate degradation after medoid suppression. Positive values indicate the GLLM is able to find either accurate and/or faithful new reasoning paths.

\smallskip

\noindent \textbf{Hyperparameters}.
%\smallskip
We elicit $N = 20$ CoTs per question, at $T = 0.7$, to provide a statistically robust sample of GLLMs' reasoning manifold without incurring prohibitive computational overhead. We use the same parameters for our adversarial experiments. Each CoT contains $[n_{l}, n_{h}] = [3, 6]$ 
declarative steps, to ensure consistency and avoid verbosity. For SS-GED computation, we run a sensitivity analysis on $\beta$ and $\gamma$ on $50$ questions from each dataset (seed $42$), by fixing the GLLM to Llama~3.1~8B (since these hyperparameters are not influenced by the GLLM choice). We vary $\beta\in\{0.0,0.2,0.4,0.6,0.8,1.0\}$ and $\gamma\in\{0.5,1.0,2.5,5.0,10.0,20.0\}$, and evaluate $\mathcal{U}_{\mathrm{GRCS}}$ correlations and GSC accuracy for every possible $(\beta, \gamma)$ pair. Varying $\beta$ changes the $\mathcal{U}_{\mathrm{GRCS}}$--faithfulness PCC by at most $0.085$ across all datasets. The largest GSC accuracy variation is $8.9\%$  for Phi-4 on GPQA, from $36.8\%$ at $\beta=0.2$ to $27.9\%$ at $\beta=1$. Varying $\gamma$ from $0.5$ to $20$ changes PCC by at most $0.104$, while GSC accuracy changes by at most $4.0\%$. Overall, for values of $\beta$ and $\gamma$ that are not extremes ($\ie$, $\gamma = 20$ or $\beta \in (0; 1)$), the pipeline is stable. We therefore use $\beta=0.4$ to slightly favor structural similarity while retaining semantic information, and $\gamma=5$ to avoid the sharper correlation shifts observed at $\gamma=20$. In GSC medoid selection, we set $\alpha=0.8$ to prioritize structural centrality while %letting ù
path consensus penalizes centrally located but weakly supported steps. %The framework is quantitatively stable across nearby values of $\alpha$ (\Cref{sec:alpha_sensitivity_appendix}).
We run a sensitivity analysis on $\alpha$ and found 
%accuracy shifts by at most $2.1$ pp across $\alpha \in [0.2, 1.0]$ relative to 
%our default $\alpha=0.8$.
results are quantitatively stable across $\alpha \in [0.4, 1.0]$ (max accuracy shift: $2.1\%$). $\alpha \in [0.0, 0.4)$ shows swings up to $\pm 10.0\%$, confirming structural centrality stabilizes medoid selection.
%\rreva{ADD BETA, GAMMA + SWEEP. FIRST SWEEP THEN BETA, GAMMA}

\begin{table}
    \centering
    \caption{\textbf{UQ Correlations (PCC) vs SOTA Baselines.} Comparison of $\mathcal{U}_{\mathrm{GRCS}}$ against SOTA baselines. More negative values indicate stronger correlation with unfaithful reasoning (lower = better). Bold results indicate the strongest performance per row. Green cells indicate $\mathcal{U}_{\mathrm{GRCS}}$ wins.} %$\mathcal{U}_{\mathrm{GRCS}}$'s median PCC
%is $-0.155$, compared with $-0.084$ for CoTA and $+0.016$ for
%Topo-UQ.} % (wins $\mathbf{77.8\%}$ ($14/18$) of times against CoTA, $\mathbf{77.8\%}$ ($14/18$) against Topo-UQ, $\mathbf{66.7\%}$ ($12/18$) against both).}
    \label{tab:absolute_pcc}
    \resizebox{\columnwidth}{!}{%
    \begin{tabular}{llccc}
        \toprule
        \textbf{GLLM} & \textbf{Dataset} & \textbf{CoTA \cite{tanneru2024quantifying}} & \textbf{Topo-UQ \cite{da2025understanding}} & \textbf{$\mathcal{U}_{\mathrm{GRCS}}$ (Ours)} \\
        \midrule
        \multirow{6}{*}{\textbf{Llama 3.1 8B}}
        & GSM8K      & -0.040          & -0.140          & \win{-0.306}    \\
        & BoolQ      & -0.070          & \textbf{-0.240} & -0.158          \\
        & StratQA    & -0.132          & +0.008          & \win{-0.213}    \\
        & MedQA      & -0.194          & +0.062          & \win{-0.280}    \\
        & GPQA       & +0.053          & +0.024          & \win{+0.007}    \\
        & AIME '26   & +0.054          & +0.162          & \win{+0.044}    \\
        \midrule
        \multirow{6}{*}{\textbf{Phi-4}}
        & GSM8K      & \textbf{-0.140} & +0.120          & -0.136          \\
        & BoolQ      & +0.080          & +0.010          & \win{-0.197}    \\
        & StratQA    & \textbf{-0.149} & -0.015          & -0.068          \\
        & MedQA      & -0.216          & +0.124          & \win{-0.241}    \\
        & GPQA       & -0.086          & +0.069          & \win{-0.109}    \\
        & AIME '26   & +0.001          & +0.068          & \win{-0.253}    \\
        \midrule
        \multirow{6}{*}{\textbf{DeepSeek R1}}
        & GSM8K      & \textbf{-0.400} & -0.220          & -0.150          \\
        & BoolQ      & -0.220          & \textbf{-0.290} & -0.022          \\
        & StratQA    & -0.153          & +0.021          & \win{-0.184}    \\
        & MedQA      & -0.081          & -0.037          & \win{-0.105}    \\
        & GPQA       & +0.061          & \textbf{-0.049} & -0.023          \\
        & AIME '26   & +0.123          & +0.246          & \win{-0.542}    \\
        \bottomrule
    \end{tabular}%
    }
    \vspace{-4mm}
\end{table}

\section{Experimental Results
}\label{sec:exp_results}

\definecolor{uqgreen}{RGB}{0,136,55}
\definecolor{uqorange}{RGB}{214,102,0}

\begin{figure*}[t]
    \centering

    % Left Column: GRCS vs. CoTA
    \begin{minipage}{0.49\textwidth}
        \centering
        \centerline{\small\textbf{(a) $\mathcal{U}_{\mathrm{GRCS}}$ vs. Topo-UQ}}
        \includegraphics[width=\linewidth]{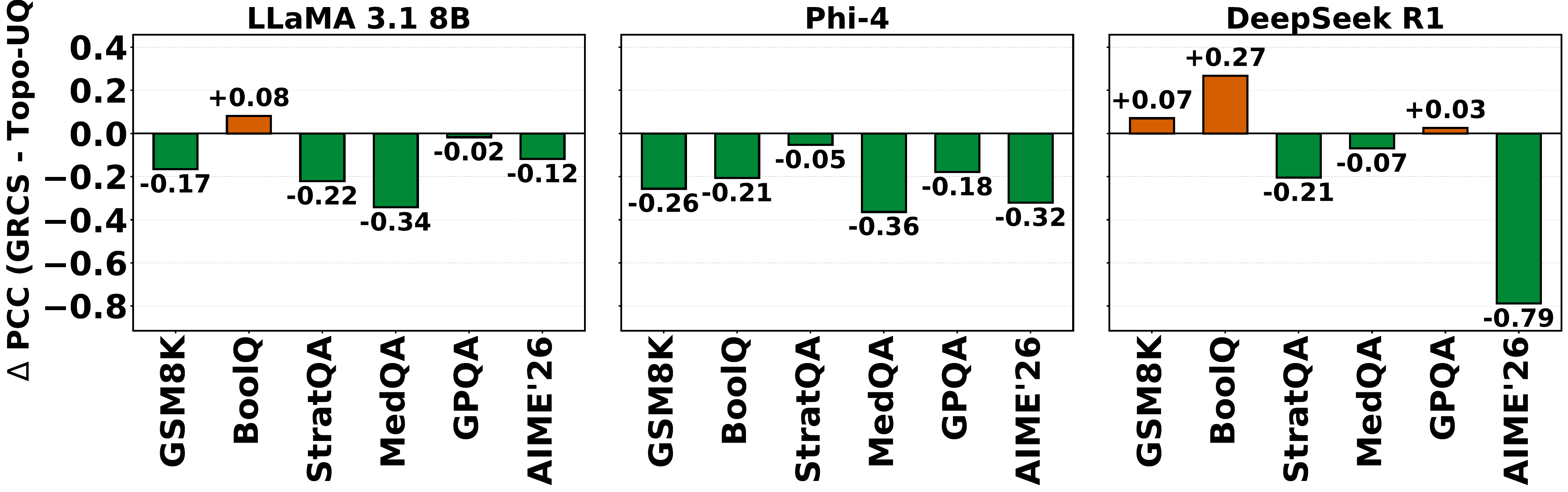}
        %\vspace{-2mm}
        %\centerline{\small\textbf{(a) GRCS vs. CoTA}}
    \end{minipage}
    \hfill
    % Right Column: GRCS vs. Topo-UQ
    \begin{minipage}{0.49\textwidth}
        \centering
       \centerline{\small\textbf{(b) $\mathcal{U}_{\mathrm{GRCS}}$ vs. CoTA}}
        \includegraphics[width=\linewidth]{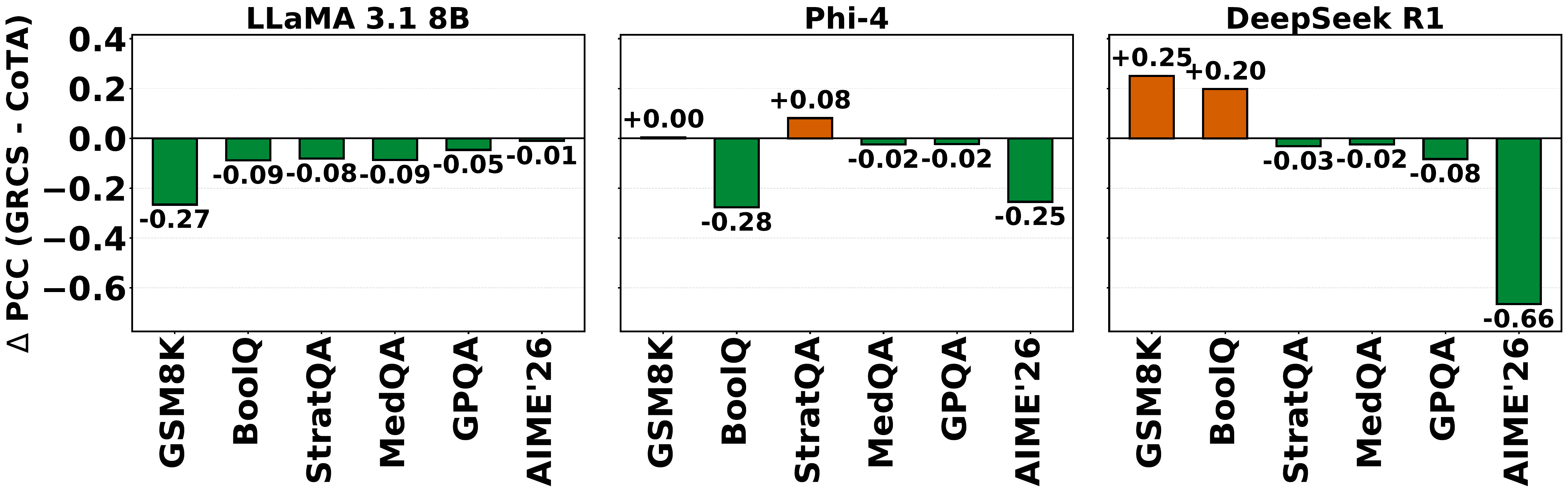}
        %\vspace{-2mm}
        %\centerline{\small\textbf{(b) GRCS vs. Topo-UQ}}
    \end{minipage}

    \vspace{4mm}

    % Shared legend
    \begin{minipage}{\textwidth}
        \centering
        \begin{tikzpicture}[
            legendtext/.style={
                inner sep=0pt,
                outer sep=0pt
            },
            legendbox/.style={
                draw=black,
                line width=0.7pt,
                minimum width=6mm,
                minimum height=2.7mm,
                inner sep=0pt,
                outer sep=0pt
            }
        ]

            % GRCS
            \node[legendbox, fill=uqgreen] (greenbox) {};

            \node[
                legendtext,
                right=1.5mm of greenbox
            ] (greentext)
            {\small\textbf{$\mathcal{U}_{\mathrm{GRCS}}$'s PCC More Negative (Better)}};

            % Baseline
            \node[
                legendbox,
                fill=uqorange,
                right=8mm of greentext
            ] (orangebox) {};

            \node[
                legendtext,
                right=1.5mm of orangebox
            ] (orangetext)
            {\small\textbf{Baseline's PCC More Negative}};

        \end{tikzpicture}
    \end{minipage}

    \vspace{-1mm}

    \caption{\textbf{Absolute UQ Correlation Improvement ($\Delta$ PCC) vs. SOTA Baselines.}
    Downward (green) bars indicate stronger negative correlations than the baseline.
    $\mathcal{U}_{\mathrm{GRCS}}$ outperforms each baseline in $77.8\%$ of configurations, with the largest
    gains on complex tasks such as AIME~'26.}
    \label{fig:uq_delta_main}

\end{figure*}

To address the RQs of \Cref{sec:intro}, we evaluate $\pname$ %our graph-based framework 
for UQ (\Cref{sec_rq1_all,sec:ea_cf_comparison,sec:GRCS_comp_cost};  \textbf{RQ1}), decoding % in 
(\Cref{sec_rq2_all,sec:decoding_res}; \textbf{RQ2}), and adversarial robustness (\Cref{sec:sec_rq3_broad_ablation}; \textbf{RQ3}). %, across GLLMs and tasks %detailed in \Cref{sec:exp_setup}.
%detailed in \Cref{sec:exp_setup}.
\Cref{sec:scalability_resources} evaluates 
%tests 
stability under longer reasoning. % paths.
\Cref{sec:dllm_ablation_sec} presents the DLLM ablation study.

\subsection{%RQ1: Uncertainty Analysis
Effectiveness of $\pname$ for UQ
%\dpal{START HERE}
}\label{sec_rq1_all}

%\rreva{we can merge 5.1.1 and 5.1.2 using a big graph when cota+topo-uq baseslines for the other datasets are available, ideally by today.}
%I've totally rewritten this. Since by just commenting the text coudl lead to a whole messm I got rid of the text saving it in a separate file, just for backup, bu i don't think we need it.
\emph{To evaluate the effectiveness of %our 
%the UQ framework of $\pname$ %and its applicability %generalizability 
$\pname$ for UQ
%across %diverse cognitive domains of increasing complexity}, we compare %our proposed GRCS metric against %SOTA 
%benchmarks of 
under increasing reasoning complexity}, we compare $\mathcal{U}_{\mathrm{GRCS}}$ against 
baselines %(CoTA, Topo-UQ, Dispersion, Entropy), 
across %five 
%diverse cognitive domains 
various benchmarks and GLLMs (\cf,~\Cref{sec:exp_setup}).  %spanning arithmetic (GSM8K), reading comprehension (BoolQ), commonsense (StratQA), clinical (MedQA), and expert reasoning (GPQA). 
\Cref{tab:absolute_pcc} reports the PCC %Pearson Correlation Coefficients (PCC) 
of these metrics against reasoning faithfulness (measured by EA, $\cf$, \Cref{sec:exp_setup}) %across three GLLMs, %using the boostrap resempling strategy described in \Cref{sec:exp_setup}, 
while
\Cref{fig:uq_delta_main} shows the absolute PCC improvement ($\Delta$ PCC) with respect to each baseline. % over SOTA baselines ($\cf$, \Cref{sec:exp_setup}).
%Due to the page limit, SRC and KRC results are shown in Appendix~\ref{sec:krc_src_section_appendix}. 
%{\em ù
Stronger negative correlations with faithfulness indicate better uncertainty estimation, meaning the metric more accurately quantifies %flags 
unfaithful reasoning~\cite{tanneru2024quantifying}. %, da2025understanding}. %For multiple GLLM--dataset pairs, dispersion and entropy exhibit weak, inconsistent, or even positive correlations (\eg, they yield positive correlations on GSM8K/BoolQ for Llama 3.1 8B and on GSM8K/StratQA for DeepSeek R1), suggesting that dispersion and entropy %variance-only signals 
%mistake repetitive but flawed reasoning for confidence. This may occur especially when models collapse onto near-identical invalid reasoning (See Appendix \ref{sec:sec_top_mapping} for details on topological mapping).
%As demonstrated in our topological mapping (Appendix \Cref{sec:decoding_res_path_appendix}), .
%Consequently, our goal is to identify a metric that maintains this negative correlation {\em as consistently as possible} across %diverse 
%all benchmarks. %This %universality 
Our goal is therefore to identify the metric that most consistently maintains %this 
a negative correlation across all benchmarks.
%is the strongest indicator of a metric's real-world reliability and robustness.

While existing baselines show strength in specific regimes, they struggle to generalize across task complexities and model sizes.
Topo-UQ demonstrates strong negative correlations only on the least difficult datasets (GSM8K, BoolQ) for Llama 3.1 8B and DeepSeek R1.
However, it struggles on more complex benchmarks (\eg, it yields distinctly positive or almost-zero correlations for Phi-4 on all tasks, and flips to positive for Llama $3.1$ 8B on StratQA, MedQA, GPQA, and AIME'26). Because it measures purely structural spread, repeated erroneous derivations on these challenging tasks yield artificially low uncertainty. 
Overall, CoTA proves to be more reliable than Topo-UQ %,% demonstrating 
and shows highly effective performance on DeepSeek R1  on GSM8K, reaching a correlation of $-0.40$. %for basic mathematical reasoning (achieving $-0.40$ on GSM8K), but it ultimately suffers from a similar boundary condition, yielding positive correlations on GPQA for both Llama $3.1$ 8B and DeepSeek R1, and flipping entirely to positive correlations on AIME '26 across all GLLMs.
Nevertheless, it weakens on more challenging tasks and yields a positive correlation on GPQA for both Llama 3.1 8B and DeepSeek R1, and on AIME'26 for all GLLMs. The key insight is that CoTA measures consistency across reasoning steps, but ignores the causal dependencies connecting them. Consequently, semantically consistent steps connected through invalid causal dependencies may still appear reliable.
In contrast, GRCS jointly evaluates semantic coherence and structural similarity ($\cf$, \Cref{sec:grcs}), making it more sensitive to 
inconsistent reasoning and invalid causal dependencies,
and
$\mathcal{U}_{\mathrm{GRCS}}$ converts this coherence into the
corresponding reasoning uncertainty proxy.
%and results shows it is more robust and sensitive to confident but logically unfaithful reasoning. 
%GRCS yields the strongest negative correlation among all evaluated metrics in 14 out of the 18 (\textbf{77.8\%}) $\langle$GLLM--dataset$\rangle$ configurations and %, 
%remains %ng 
%negatively correlated with faithfulness in 16 out of 18 (\textbf{88.9\%}) configurations. %This makes it the \bem{most consistent cross-scale uncertainty signal.}
%In direct head-to-head comparisons, 
$\mathcal{U}_{\mathrm{GRCS}}$ \textbf{outperforms CoTA in} $\mathbf{77.8\%}$ ($14$ of $18$) of the $\langle$GLLM--dataset$\rangle$ configurations, and simultaneously \textbf{outperforms Topo-UQ in} $\mathbf{77.8\%}$ ($14$ of $18$), %of configurations (\Cref{fig:uq_delta_main}). 
with partially overlapping wins: it outperforms \textbf{both baselines} in \textbf{$66.7$\%} ($12/18$), only CoTA in $2/18$, only Topo-UQ in $2/18$, and neither baseline in the remaining $2/18$ configurations (GSM8K, BoolQ with DeepSeek~R1, \Cref{fig:uq_delta_main}). %Overall, GRCS yields 
%It achieves the strongest negative correlation among all evaluated metrics in $12$ out of the $18$ (\textbf{$66.7$\%}) configurations, and 
Overall, it %is %consistently negatively correlated with faithfulness 
\textbf{assigns higher uncertainty to less faithful reasoning in} $\mathbf{88.9\%}$ ($16$ out of $18$) %(\textbf{$88.9$\%}) 
configurations, by being negatively correlated with faithfulness (\Cref{tab:absolute_pcc}). Across the six datasets, $U_{\mathrm{GRCS}}$ also achieves the strongest
median PCC for every GLLM: $-0.186$ (Llama~3.1~8B), $-0.167$ (Phi-4),
and $-0.128$ (DeepSeek~R1), compared with $+0.016/+0.069/-0.043$ for Topo-UQ and $-0.055/-0.113/-0.117$ for
CoTA.
%Among all results, $\mathcal{U}_{\mathrm{GRCS}}$ achieves a strong negative correlation of nearly -$0.6$ on the demanding AIME'26 benchmark for DeepSeek R1, proving its ability to capture hidden uncertainty in long-horizon mathematical reasoning where structural-only metrics like Topo-UQ weaken. % (it achieves a 
Its strongest result occurs on
AIME'26 with DeepSeek~R1 ($-0.542$), proving its ability to capture hidden uncertainty in long-horizon mathematical reasoning where both baselines weaken. %positive correlation of $+0.246$).
%We note that GRCS correlations weaken on DeepSeek R1. More capable models exhibit greater reasoning heterogeneity under uncertainty (%Appendix 
%\Cref{sec:sec_top_mapping}), reducing the rigid mode collapse that GRCS most effectively detects.
%However, 
SRC %exactly 
reproduced the PCC correlation direction in all
configurations, while KRC agreed in $17/18$ configurations
($94.4\%$, as it weakens on GPQA with DeepSeek~R1).
Our results also highlight the boundary conditions of topological UQ. For Llama $3.1$ 8B, all evaluated metrics (including %our 
$\mathcal{U}_{\mathrm{GRCS}}$) yield positive correlations on the highly complex GPQA and AIME '26. This indicates that, when a task vastly exceeds a GLLM's capability, the model often lacks the domain knowledge to meaningfully explore the reasoning space and %usually defaults to %repetitive hallucinated shortcuts. 
can default to
full mode-collapse. Topo-UQ is not able to detect both partial and full mode collapse, as the results show (positive correlations in $4$ out of $6$ benchmarks for Llama 3.1 8B). As we showed in~\Cref{fig:grcs_sweeps}, GRCS detects partial mode-collapse, where baselines weaken, but cannot distinguish faithful consensus from the complete repetition of a single hallucinated derivation. Consequently, neither can $\mathcal{U}_{\mathrm{GRCS}}$. %Topological metrics mistake this pathological repetition for reasoning consensus, %artificially 
%and assign low uncertainty to completely unfaithful outputs, which drives the correlation positive.
%Ultimately, 
\bem{Despite this boundary condition, $\mathcal{U}_{\mathrm{GRCS}}$ is the most
consistent of the evaluated UQ metrics for identifying unfaithful
reasoning.} %} %across %both bigger and smaller GLLMs}.
%various benchmarks of different complexity}.
%\subsection{%RQ2: Decoding Analysis
%Effectiveness of Structural and Causal Awareness to Filter ``Lucky Guesses''
%}\label{sec_rq2_all}

%\begin{figure}
%    \centering
%    %\includegraphics[width=0.65\textwidth]
%    \includegraphics[scale=0.35]
%    {pic/grcs_consistency_scatter.pdf}
%    \caption{\textbf{GRCS Stability Across Independent Faithfulness %Measures} The tight clustering within $\pm0.05$ of the diagonal ($r = %0.83$, MAE = $0.06$) proves that GRCS reliably isolates unfaithful %reasoning across both EA and CFE. %when evaluating faithfulness using both %EA and CFE. 
%    \label{fig:ea_cf_comp_chart}
%    %\vspace{-6mm}
%\end{figure}

% Colors
\definecolor{gsm8k}{HTML}{DB7526}
\definecolor{boolq}{HTML}{2687BD}
\definecolor{gpqa}{HTML}{26AC87}
\definecolor{medqa}{HTML}{D38CB4}
\definecolor{strategyqa}{HTML}{E9AD26}
\definecolor{aime}{HTML}{6FBFEC}
\definecolor{modelgray}{HTML}{555555}
\definecolor{linegray}{HTML}{777777}

% Legend symbols
\newcommand{\legendcircle}[1]{%
    \tikz[baseline=-0.55ex]{%
        \filldraw[
            fill=#1,
            draw=white,
            line width=0.35pt
        ] (0,0) circle (2.8pt);
    }%
}

\newcommand{\legendsquare}{%
    \tikz[baseline=-0.55ex]{%
        \filldraw[
            fill=modelgray,
            draw=white,
            line width=0.35pt
        ] (-2.8pt,-2.8pt) rectangle (2.8pt,2.8pt);
    }%
}

\newcommand{\legendcross}{%
    \tikz[baseline=-0.55ex]{%
        \draw[modelgray,line width=1.6pt]
            (-2.8pt,-2.8pt) -- (2.8pt,2.8pt);
        \draw[modelgray,line width=1.6pt]
            (-2.8pt,2.8pt) -- (2.8pt,-2.8pt);
    }%
}

\newcommand{\legendline}{%
    \tikz[baseline=-0.15ex]{%
        \draw[linegray,dashed,line width=0.8pt]
            (0,0) -- (13pt,0);
    }%
}
\begin{figure}
    \centering

    \begin{minipage}[c]{0.55\columnwidth}
        \centering
        \includegraphics[width=\linewidth]
        {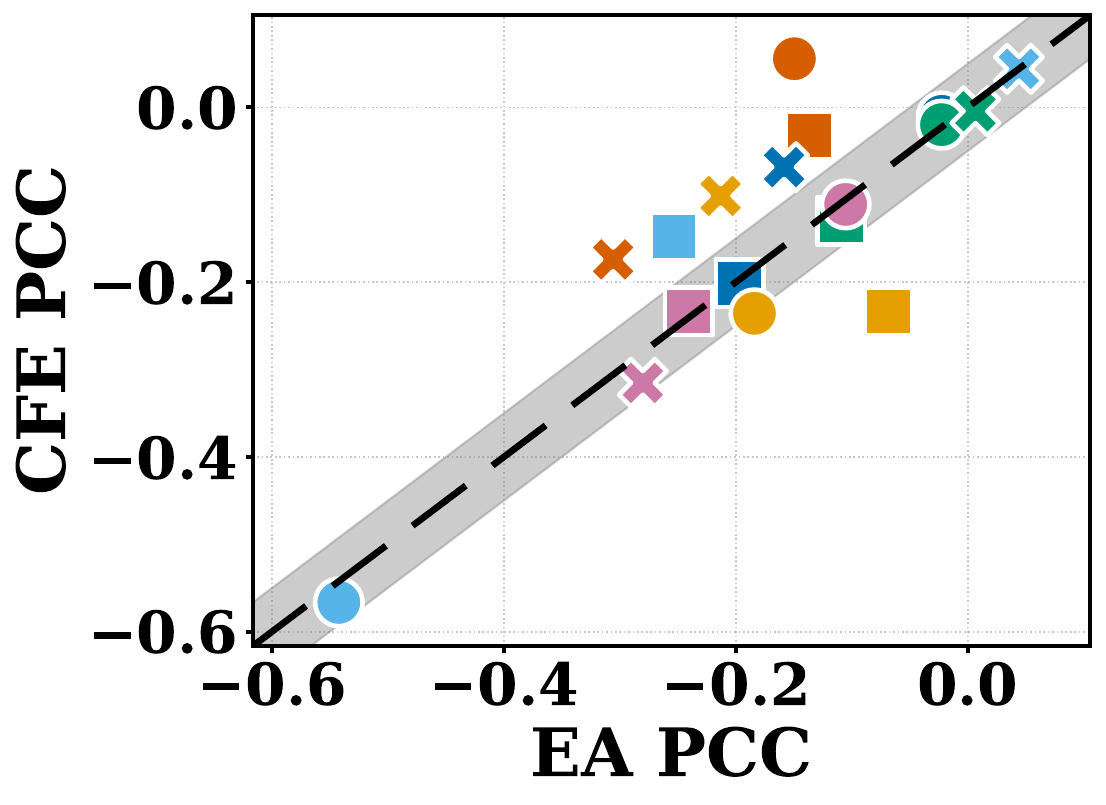}
    \end{minipage}\hfill
    \begin{minipage}[c]{0.41\columnwidth}
        \centering
        \scriptsize
        \setlength{\tabcolsep}{1.5pt}
        \renewcommand{\arraystretch}{1.12}

        \begin{tabular}{@{}c@{\hspace{3pt}}l@{}}
            \multicolumn{2}{c}{\bfseries Dataset} \\[1pt]

            \legendcircle{gsm8k}
                & \textbf{GSM8K} \\

            \legendcircle{boolq}
                & \textbf{BoolQ} \\

            \legendcircle{gpqa}
                & \textbf{GPQA} \\

            \legendcircle{medqa}
                & \textbf{MedQA} \\

            \legendcircle{strategyqa}
                & \textbf{StratQA} \\

            \legendcircle{aime}
                & \textbf{AIME'26} \\[3pt]

            \multicolumn{2}{c}{\bfseries GLLM} \\[1pt]

            \legendcircle{modelgray}
                & \textbf{DeepSeek R1} \\

            \legendcross
                & \textbf{LLaMA 3.1 8B} \\

            \legendsquare
                & \textbf{Phi-4} \\[2pt]

                \legendline & \textbf{$y=x$ (Perfect Agreement)} \\
        \end{tabular}
    \end{minipage}

    \caption{\textbf{$\mathcal{U}_{\mathrm{GRCS}}$ Stability Across Independent
    Faithfulness Measures.}
    The tight clustering within $\pm0.05$ of the diagonal
    ($r=0.83$, MAE${}=0.06$) shows that %GRCS reliably
    %isolates unfaithful reasoning across both EA and CFE.
    the uncertainty correlations are
stable across EA and CFE.}
    \label{fig:ea_cf_comp_chart}
\end{figure}

%\input{pic/audit_img}
%\vspace{-5cm}

\subsection{$\mathcal{U}_{\mathrm{GRCS}}$ Stability Across Independent Faithfulness Measures}
\label{sec:ea_cf_comparison}

\emph{To verify that $\mathcal{U}_{\mathrm{GRCS}}$ captures %intrinsic 
reasoning uncertainty, rather than reflecting bias from a single evaluation strategy}, %we evaluate faithfulness from two independent angles ($\cf$, \Cref{sec:methodology}). Specifically, 
we compare the correlation of $\mathcal{U}_{\mathrm{GRCS}}$ against reasoning faithfulness using two independent measures: %both 
Early Answering (EA)%, which tests whether the final answer depends on the complete reasoning sequence~\cite{tanneru2024quantifying}, and Counterfactual Editing (CFE), 
~\cite{tanneru2024quantifying}, and Counterfactual Editing (CFE)%,which isolates the causal contribution of %each 
%individual reasoning steps
~\cite{pearl2009causality}. \Cref{fig:ea_cf_comp_chart} shows %the %consistency of GRCS PCCs between these two methodologies, across all GLLMs and datasets. The resulting scatter plot demonstrates 
%a statistically significant positive alignment ($r = 0.83$, $p < 0.05$) between the two techniques across GLLMs and datasets. 
that $\mathcal{U}_{\mathrm{GRCS}}$ yields strongly consistent correlations with faithfulness under EA and CFE across GLLMs and datasets ($r = 0.83$, $p < 0.05$).
%The tight clustering of results %inside the $\pm0.05$ corridor 
%within $\pm0.05$ of the diagonal and the Mean Absolute Error (MAE) of $0.06$ between EA and CFE indicate that GRCS results are stable using both interventions. 
Most results lie within $\pm 0.05$ of the diagonal ($\ie$, $y = x$ denotes perfect agreement between results from the two faithfulness measures), with an MAE of only $0.06$ for the $\mathcal{U}_{\mathrm{GRCS}}$ correlations obtained using EA and CFE.
While these interventions perturb the GLLMs' reasoning topology in different independent ways, the consistent concentration of results in where both $x<0$ and $y<0$ confirms that $\mathcal{U}_{\mathrm{GRCS}}$ consistently assigns higher uncertainty to unfaithful reasoning.
%\textbf{These results show $\mathcal{U}_{\mathrm{GRCS}}$ is a robust measure of reasoning uncertainty across independent faithfulness measures.}
\bem{These results support $\mathcal{U}_{\mathrm{GRCS}}$ as a robust proxy for
reasoning uncertainty across independent faithfulness measures.}

\newcolumntype{N}[1]{%
    S[
        table-format=#1,
        table-number-alignment=center,
        detect-weight=true,
        detect-inline-weight=math
    ]%
}

\definecolor{grcsrow}{gray}{0.93}

\newcommand{\cellwin}{%
    \cellcolor{winGreen}\bfseries
}

\begin{table*}
    \centering
    \scriptsize
    \setlength{\tabcolsep}{2.5pt}
    \renewcommand{\arraystretch}{0.95}

    \caption{
        \textbf{Resource Efficiency of $\mathcal{U}_{\mathrm{GRCS}}$ and baselines.}
        GPU runtime is reported per individual CoT and per set of
        $20$ CoTs. Energy consumption, peak GPU memory, and peak GPU
        utilization are reported per set of $20$ CoTs.
        All methods operate on the same collected responses, therefore multi-round CoT generation time is excluded. Measurements %are
        collected on the same GPU over complete datasets and $5$
        independent runs, with results reported as mean $\pm$ standard
        deviation. The best mean result for each metric and dataset is bold in a green cell. Lower values are better for all metrics.
    }
    \label{tab:computational_cost}

    \begin{tabular}{
        C{1.30cm}
        L{1.30cm}
        N{2.4} @{\,${}\pm{}$\,} N{1.4}
        N{3.2} @{\,${}\pm{}$\,} N{1.2}
        N{6.2} @{\,${}\pm{}$\,} N{2.2}
        N{4.2} @{\,${}\pm{}$\,} N{1.2}
        N{2.2} @{\,${}\pm{}$\,} N{1.2}
    }
        \toprule
        \textbf{Dataset}
        & \textbf{UQ Metric}
        %& \multicolumn{2}{c}{\textbf{GPU Runtime / CoT (s)}}
        & \multicolumn{2}{c}{
            \shortstack{\textbf{GPU Runtime (s)}\\\textbf{per 1 CoT}}
        }
        & \multicolumn{2}{c}{
            \shortstack{\textbf{GPU Runtime (s)}\\\textbf{per 20 CoTs}}
        }
        %& \multicolumn{2}{c}{\textbf{Energy (J)}}
        & \multicolumn{2}{c}{
            \shortstack{\textbf{Energy (J)}\\\textbf{per 20 CoTs}}
        }
        %& \multicolumn{2}{c}{\textbf{Peak GPU Mem. (MB)}}
        & \multicolumn{2}{c}{
            \shortstack{\textbf{Peak GPU Mem. (MB)}\\\textbf{per 20 CoTs}}
        }
        %& \multicolumn{2}{c}{\textbf{Peak GPU Usage (\%)}} \\
        & \multicolumn{2}{c}{
            \shortstack{\textbf{Peak GPU Usage (\%)}\\\textbf{per 20 CoTs}}
        } \\
        
        \cmidrule(lr){3-4}
        \cmidrule(lr){5-6}
        \cmidrule(lr){7-8}
        \cmidrule(lr){9-10}
        \cmidrule(lr){11-12}

        \multirow{3}{*}{GSM8K}
        & \textbf{$\mathcal{U}_{\mathrm{GRCS}}$}
        & \cellwin 0.0381 & 0.0034
        & \cellwin 0.75   & 0.07
        & \cellwin 72.04  & 4.86
        & 1252.79      & 0.00
        & \cellwin 8.25   & 2.75 \\
        
        & Topo-UQ
        & 0.7800 & 0.0100
        & 15.39  & 0.03
        & 2465.31 & 30.18
        & \cellwin 428.85 & 0.00
        & 45.02 & 0.06 \\
        
        & CoTA
        & 0.3386 & 0.0008
        & 6.67   & 0.02
        & 1951.60 & 26.93
        & 952.24 & 0.08
        & 84.27 & 1.19 \\

        \midrule

        \multirow{3}{*}{BoolQ}
        & \textbf{$\mathcal{U}_{\mathrm{GRCS}}$}
        & \cellwin 0.0300 & 0.0033
        & \cellwin 0.60   & 0.07
        & \cellwin 60.28  & 2.32
        & 1273.67      & 0.00
        & \cellwin 11.19  & 5.53 \\
        
        & Topo-UQ
        & 0.8867 & 0.0014
        & 17.73  & 0.03
        & 3047.38 & 2.36
        & \cellwin 428.80 & 0.00
        & 45.36 & 0.31 \\
        
        & CoTA
        & 0.4243 & 0.0054
        & 8.49   & 0.11
        & 2428.33 & 39.97
        & 938.47 & 0.04
        & 86.88 & 0.13 \\
        
        \midrule
        
        \multirow{3}{*}{StratQA}
        & \textbf{$\mathcal{U}_{\mathrm{GRCS}}$}
        & \cellwin 0.0315 & 0.0006
        & \cellwin 0.63   & 0.01
        & \cellwin 62.71  & 5.69
        & 1286.92     & 0.00
        & \cellwin 11.19  & 5.53 \\
        
        & Topo-UQ
        & 0.6616 & 0.0050
        & 13.17  & 0.10
        & 2376.24 & 92.86
        & \cellwin 428.79 & 0.00
        & 46.40 & 0.22 \\
        
        & CoTA
        & 0.3622 & 0.0010
        & 7.23   & 0.02
        & 2124.12 & 38.13
        & 937.27 & 0.00
        & 85.78 & 2.58 \\
        
        \midrule
        
        \multirow{3}{*}{MedQA}
        & \textbf{$\mathcal{U}_{\mathrm{GRCS}}$}
        & \cellwin 0.0349 & 0.0018
        & \cellwin 0.70   & 0.04
        & \cellwin 64.79  & 3.23
        & 1281.63     & 0.00
        & \cellwin 5.17   & 1.23 \\
        
        & Topo-UQ
        & 1.7386 & 0.0238
        & 34.77  & 0.48
        & 6051.18 & 47.59
        & \cellwin 429.10 & 0.00
        & 47.93 & 0.59 \\
        
        & CoTA
        & 1.3937 & 0.0076
        & 27.87  & 0.15
        & 8201.94 & 93.73
        & 955.47 & 0.05
        & 94.19 & 0.42 \\
        
        \midrule
        
        \multirow{3}{*}{GPQA}
        & \textbf{$\mathcal{U}_{\mathrm{GRCS}}$}
        & \cellwin 0.0378 & 0.0007
        & \cellwin 0.74   & 0.01
        & \cellwin 72.38  & 1.49
        & 1343.50      & 0.00
        & \cellwin 8.23   & 4.21 \\
        
        & Topo-UQ
        & 0.6358 & 0.0030
        & 6.36   & 0.03
        & 1048.89 & 40.16
        & \cellwin 429.44 & 0.00
        & 48.57 & 0.00 \\
        
        & CoTA
        & 2.9603 & 0.0008
        & 49.34  & 0.01
        & 14709.75 & 4.36
        & 1136.97 & 0.00
        & 95.70 & 0.76 \\
        
        \midrule
        
        \multirow{3}{*}{AIME'26}
        & \textbf{$\mathcal{U}_{\mathrm{GRCS}}$}
        & \cellwin 0.0674 & 0.0004
        & \cellwin 1.35   & 0.01
        & \cellwin 34.60  & 0.24
        & 1405.09      & 0.00
        & \cellwin 17.23  & 2.50 \\
        
        & Topo-UQ
        & 5.5800 & 0.0729
        & 111.28 & 1.45
        & 927.54 & 22.01
        & \cellwin 435.10 & 0.00
        & 49.78 & 0.92 \\
        
        & CoTA
        & 22.8113 & 0.0153
        & 455.47  & 0.30
        & 135539.46 & 97.34
        & 1856.45 & 0.16
        & 98.50 & 0.07 \\

        \bottomrule
    \end{tabular}
\end{table*}

\subsection{Reasource Efficiency of $\mathcal{U}_{\mathrm{GRCS}}$}\label{sec:GRCS_comp_cost}
\emph{%To assess the computational cost and the energy footprint of the GRCS pipeline, and compare it against SOTA baselines}, 
To compare $\mathcal{U}_{\mathrm{GRCS}}$ with SOTA baselines in resource efficiency} we report, for each dataset, the GPU runtime, %measured in seconds, %needed to compute GRCS both for a single GLLM CoT, and for the whole set of 20 sampled GLLM CoT following \Cref{sec:exp_setup}, and compare it with wall clock time needed to execute Topo-UQ and CoTA on a single CoT and the whole set of 20 CoTs. Please note that, since the above methods share the same collected responses set, we did not include the time used for multi-round generation for comparisons. In addition, we report the energy consumption, measured in Joules, and the peak GPU memory footprint, measured in MB, required by each method under the same experimental conditions. All measurements are collected on the same hardware, using the same GPU for every method, over 5 independent runs, and we report the mean and standard deviation across runs.
the energy consumption and the peak GPU %usage and 
memory footprint and usage ($\cf$, \Cref{sec:exp_setup}).
%\rreva{complete table and put analysis here}. ù
\Cref{tab:computational_cost} shows that $\mathcal{U}_{\mathrm{GRCS}}$ is consistently the fastest, most energy-efficient, and least GPU-intensive metric across all six datasets. %It requires up to $0.75$s (average: $0.66$s) and $72.04$J (average: $65.01$J) per set of $20$ CoTs on simpler benchmarks (\eg, GSM8K, BoolQ, and StratQA), %$0.60$--$0.75$s and $60$--$72$J per set of $20$ CoTs on the simpler benchmarks ($\eg$, GSM8K, BoolQ, StratQA), which is approximately a $\mathbf{23\times}$ {\bf speed-up} and a 
%corresponding to a $\mathbf{23\times}$ \textbf{speed-up} and a $\mathbf{40\times}$ \textbf{reduction} in energy consumption relative to Topo-UQ,
%$\mathbf{40\times}$ \textbf{reduction} in energy consumption relative to Topo-UQ ($13.17$--$17.73$s, $2,376$--$3,047$J on the same three datasets), roughly an 
%$\mathbf{11\times}$ \textbf{speed-up} and a 
%$\mathbf{37\times}$ \textbf{reduction} in energy consumption relative to CoTA ($6.77$-%-$8.49$s, $2,377$--$2,428$J). 
It requires up to $1.35$s (geometric mean: $0.76$s) and $72.38$J (geometric mean: $59.46$J) per set of $20$ CoTs. It is up to $82\times$ faster (geometric mean: $28\times$) and $93\times$ more energy-efficient (geometric mean: $37\times$) than Topo-UQ, and up to $337\times$ faster (geometric mean: $33\times$) and $3{,}917\times$ more energy-efficient (geometric mean: $125\times$) than CoTA. 
%and a $\mathbf{11\times}$ \textbf{speed-up} and $\mathbf{37\times}$ \textbf{reduction} in energy consumption relative to CoTA.
%This efficiency %advantage widens substantially 
%continues on AIME '26 (the hardest setting). GPU runtime per $20$ CoTs nearly doubles for GRCS with respect to the most expensive non-AIME setting (GSM8K), while for Topo-UQ it increases by up to $3.2\times$
%relative to its most expensive non-AIME setting (MedQA) and for CoTA it climbs up to $9.2\times$ relative to its most expensive non-AIME setting (GPQA). 
On the hardest datasets, CoTA shows the most severe degradation of all three methods and a significant increase in energy consumption, from a maximum of $14,709.75$J on GPQA to $135,539.46$J on AIME'26 ($3,917.9\times$ $\mathcal{U}_{\mathrm{GRCS}}$'s own energy cost ($34.60$J) on the same dataset). The gap between $\mathcal{U}_{\mathrm{GRCS}}$ and the baselines primarily stems from implementation differences in embedding. %Topo UQ processes text sequentiallty y and, for each pairwise comparison, independently computes node and edge embeddings for each of the 
%N sampled CoTs.
For each question, Topo-UQ compares every pair of the $N$ sampled CoTs, and recomputes each chain's nodes and edges embeddings $(N - 1)$ times, causing an $\mathcal{O}(N^2)$ redundancy in embedding calls. % that grows both with 
%N and with the size of the underlying reasoning graphs. 
When computing $\mathcal{U}_{\mathrm{GRCS}}$, we avoid this redundancy by computing all embeddings in parallel batches during graph construction (\Cref{sec:methodology}), and reusing them across all pairwise graph comparisons. This reduces the embedding cost from $\mathcal{O}(N^2)$ to $\mathcal{O}(N)$ and makes GPU runtime more stable. CoTA uses a cross-encoder, which scores every pair of reasoning steps across $\binom{N}{2}$ CoT pairs. We avoid this redundancy by representing each CoT with an independently computed embedding rather than a joint pairwise representation. As a result, all $\binom{N}{2}$ comparisons reuse the same set of embeddings. $\mathcal{U}_{\mathrm{GRCS}}$'s higher peak memory footprint (on average, $3\times$ more than Topo-UQ and $1.2\times$ than CoTA, although CoTA consumes more on AIME'26) comes from using a newer and larger embedding model (Qwen3-Embedding-0.6B, $\sim$600M parameters \cite{yang2025qwen3technicalreport}) compared to the BERT-base ($\sim$110M) and DeBERTa-v3-base ($\sim$184M) embedders used by Topo-UQ and CoTA, respectively~)\cite{bert_ref}. \bem{These results show that, despite using a heavier embedding model with a higher peak GPU memory consumption, computing $\mathcal{U}_{\mathrm{GRCS}}$ %still 
achieves lower execution time, %(\(23\text{--}82\times\) faster), 
energy consumption, %(\(37\text{--}3{,}918\times\) lower), 
and peak GPU utilization %(\(4\text{--}8\times\) lower) 
compared to the baselines}. %}

%\begin{figure}[h]

%\setlength{\intextsep}{0pt} % Default is usually larger
%\begin{wrapfigure}[18]{r}{0.3\textwidth}
%    \centering
%    \includegraphics[width=\linewidth]
    %\includegraphics[scale=0.18]
%    {pic/accuracy_combined_heatmap.pdf} 
%    \caption{\textbf{Testing for Real, Faithful Accuracy}. 
    %\textit{Methodology}: We compare the nominal accuracy of standard SC against our GSC framework using the same datasets and models of \Cref{sec_rq1_all}. 
%    The rightmost panel displays the %absolute 
%    performance difference ($\Delta$ GSC $-$ SC). 
    %\textit{Results}: 
%    Smaller models (\eg, LLaMA 3.1 8B) show accuracy drops %(up to $-9.0$ pp on MedQA), 
%    revealing reliance on SC-rewarded lucky guesses. As reasoning capability increases (Phi-4, DeepSeek R1), the gap narrows and turns positive with DeepSeek R1 improving in all settings.} %GSC thus filters hallucinations while preserving or improving strong-model performance.}
    \label{fig:exp3_fig}
    %\vspace{-5mm}
%\end{figure}
%\end{wrapfigure}

\begin{figure}
         \centering
         \includegraphics[width=\linewidth]{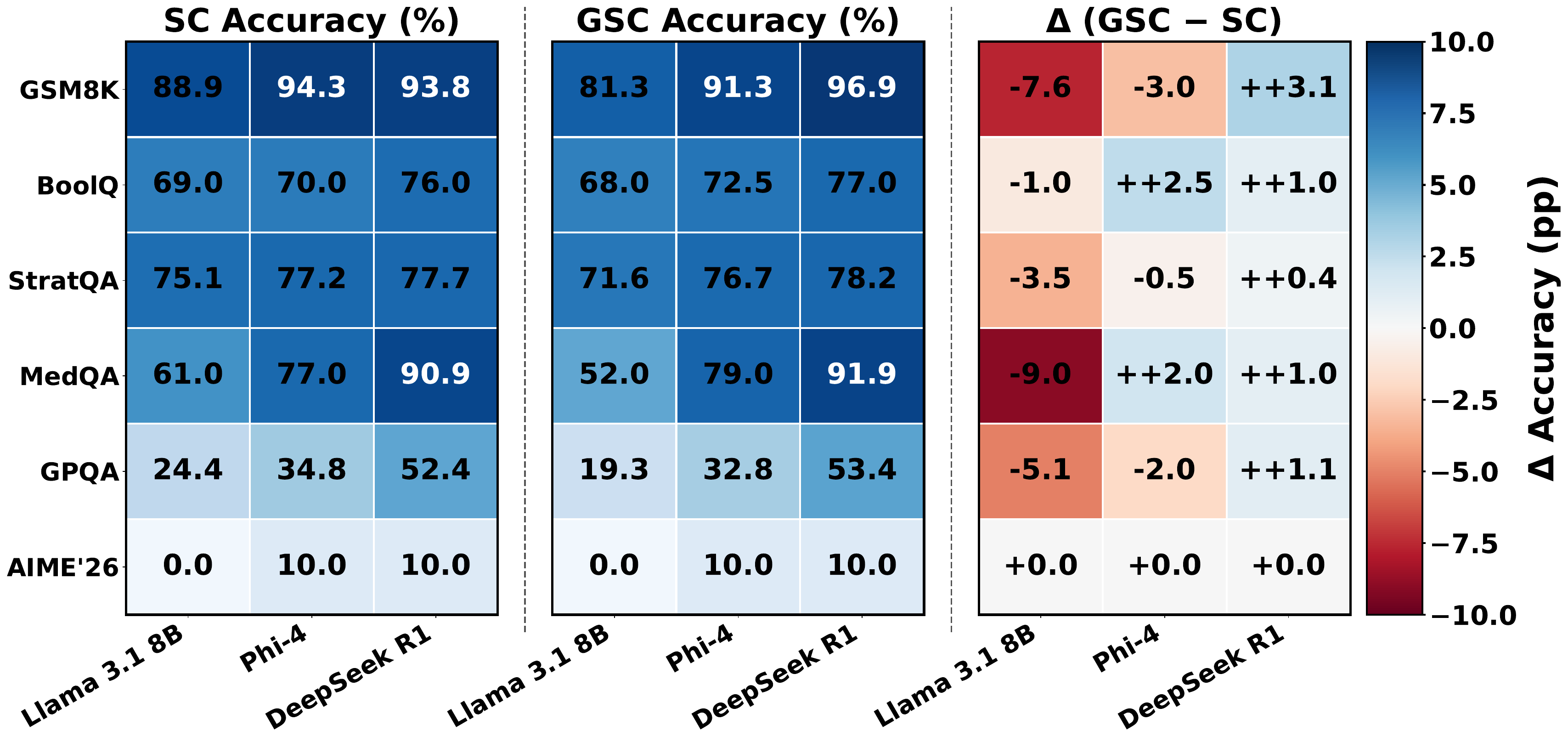}
            %\caption{\textbf{Testing for Real, Faithful Accuracy}. 
     %\textit{Methodology}: We compare the nominal accuracy of standard SC against our GSC framework using the same datasets and models of \Cref{sec_rq1_all}. 
     %The rightmost panel displays the absolute performance difference ($\Delta$ GSC $-$ SC). 
%     %\textit{Results}: 
     %Smaller models (\eg, LLaMA 3.1 8B) show accuracy drops (up to $-9.0\%$ %pp 
     %on MedQA), revealing reliance on SC-rewarded lucky guesses. 
    % As reasoning strength increases (Phi-4, DeepSeek R1), the gap narrows and turns positive with DeepSeek R1 improving in all settings. GSC thus filters hallucinations while preserving or improving strong-model performance.}
    \caption{
    The rightmost panel shows $\Delta$ GSC $-$ SC. Smaller models (\eg, LLaMA 3.1 8B) show accuracy drops (up to $-9.0\%$ on MedQA), suggesting that SC can favor answers supported by inconsistent reasoning. The gap narrows for stronger models and becomes positive for DeepSeek R1.
    }
     \label{fig:faithful_accuracy_heatmap}
 \end{figure}

% ============================================================
% Preamble
% ============================================================
% These widths are relative to each half of the full figure.
\newcommand{\entropylabelwidth}{0.075\linewidth}
\newcommand{\entropywidth}{0.295\linewidth}
\newcommand{\entropyheaders}{
    \begin{minipage}[c]{\entropylabelwidth}
        \mbox{}
    \end{minipage}%
    \hspace{2mm}%
    \begin{minipage}[c]{\entropywidth}
        \centering
        \scriptsize\textbf{Llama-3.1 8B}
    \end{minipage}%
    \hfill
    \begin{minipage}[c]{\entropywidth}
        \centering
        \scriptsize\textbf{Phi-4}
    \end{minipage}%
    \hfill
    \begin{minipage}[c]{\entropywidth}
        \centering
        \scriptsize\textbf{DeepSeek R1}
    \end{minipage}
    \vspace{1mm}
}
% #1 = dataset name
% #2 = Llama plot
% #3 = Phi plot
% #4 = DeepSeek plot
\newcommand{\entropyrow}[4]{
    \begin{minipage}[c]{\entropylabelwidth}
        \centering
        % origin=c pivots the rotation around the text's CENTER instead of its
        % bottom-left corner. Without this, longer labels (StrategyQA, AIME '26)
        % rotate around a wider unrotated box and end up visually shifted further
        % left than shorter labels (GSM8K, BoolQ) - which is why row 3 looked
        % offset relative to the others.
        \rotatebox[origin=c]{90}{\scriptsize\textbf{#1}}
    \end{minipage}%
    \hspace{2mm}%
    \begin{minipage}[c]{\entropywidth}
        \centering
        \includegraphics[width=\linewidth]{#2}
    \end{minipage}%
    \hfill
    \begin{minipage}[c]{\entropywidth}
        \centering
        \includegraphics[width=\linewidth]{#3}
    \end{minipage}%
    \hfill
    \begin{minipage}[c]{\entropywidth}
        \centering
        \includegraphics[width=\linewidth]{#4}
    \end{minipage}
    \vspace{2mm}
}
\begin{figure*}[t]
    \centering
    % ========================================================
    % LEFT HALF
    % ========================================================
    \begin{minipage}[t]{0.495\textwidth}
        \centering
        \entropyrow
            {GSM8K}
            {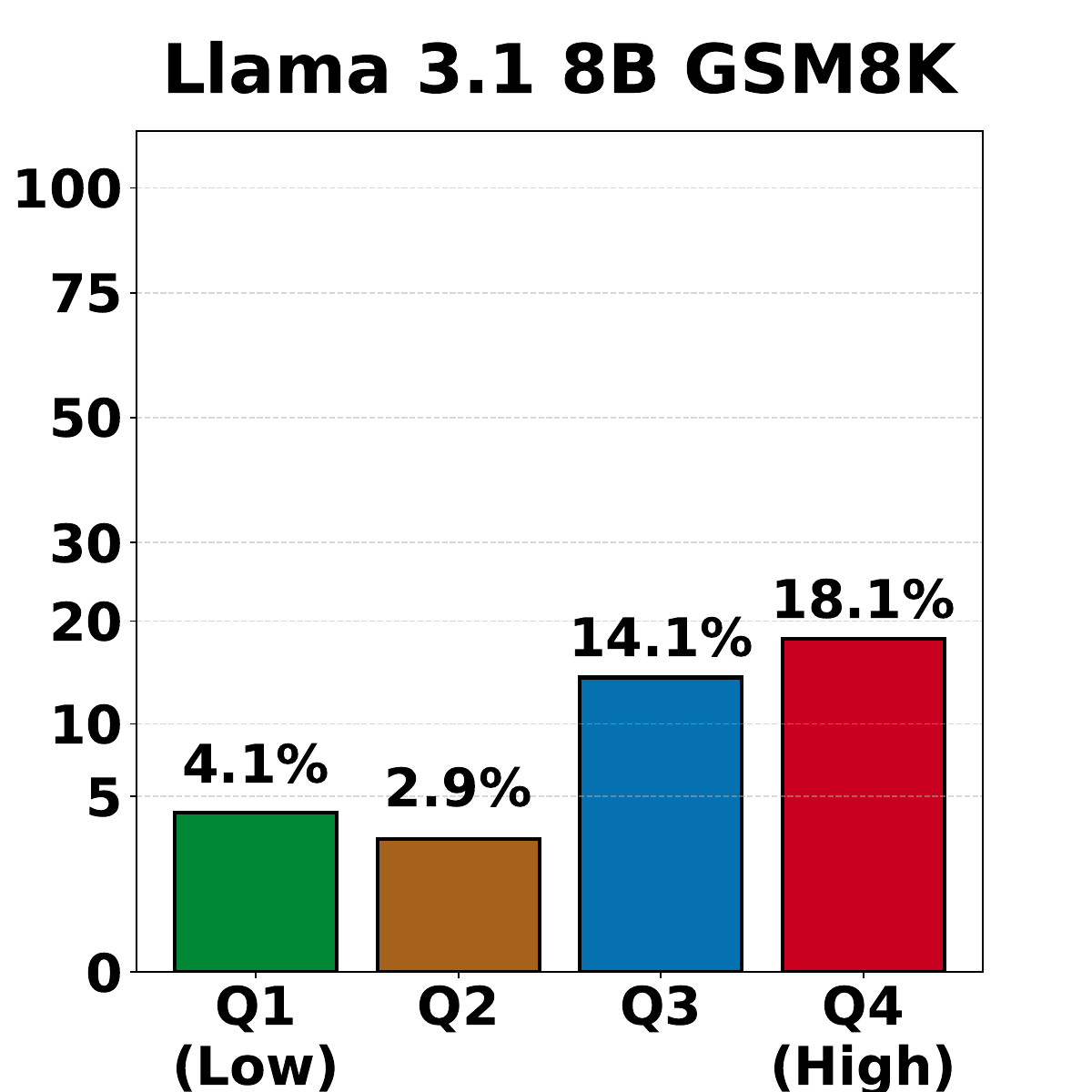}
            {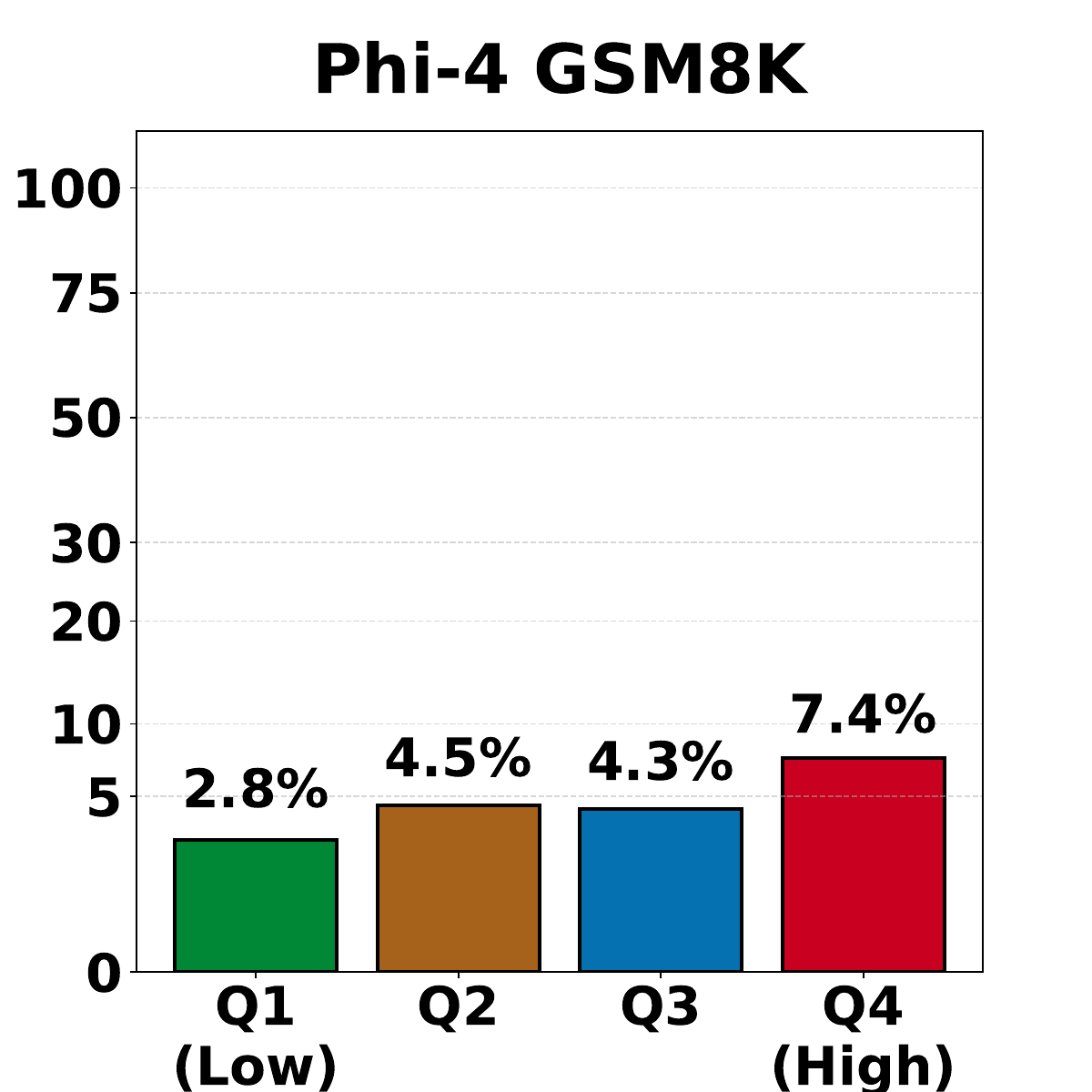}
            {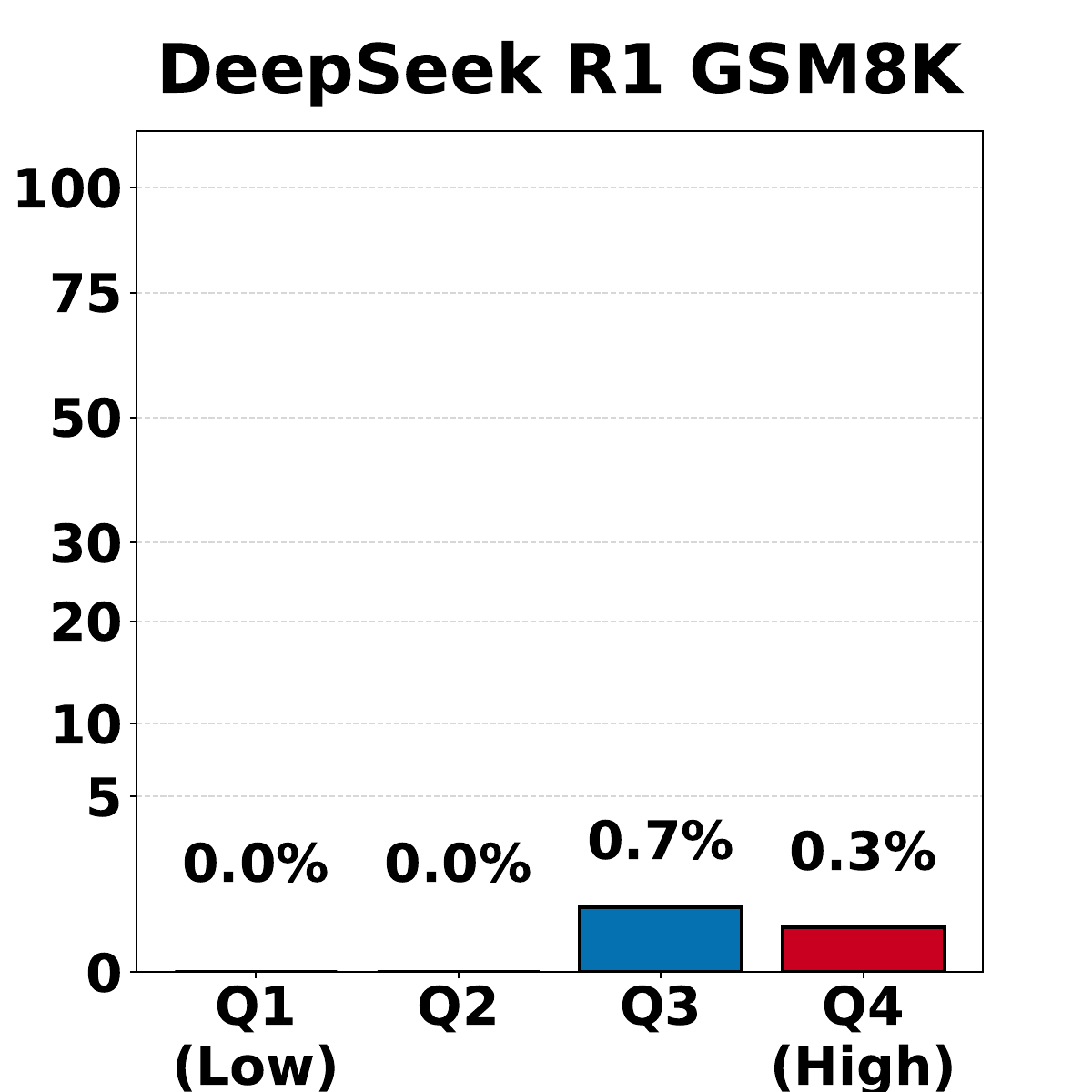}
        \entropyrow
            {BoolQ}
            {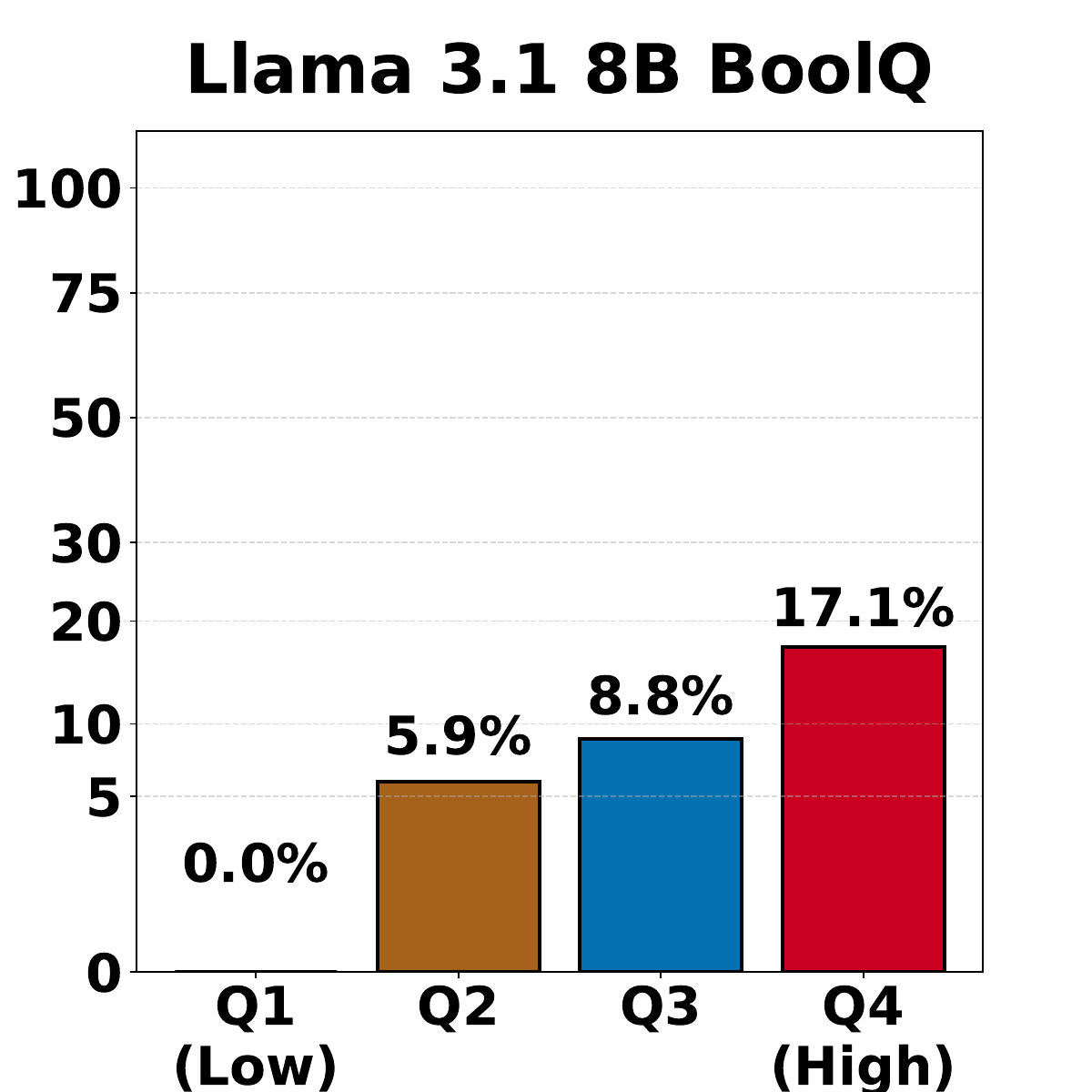}
            {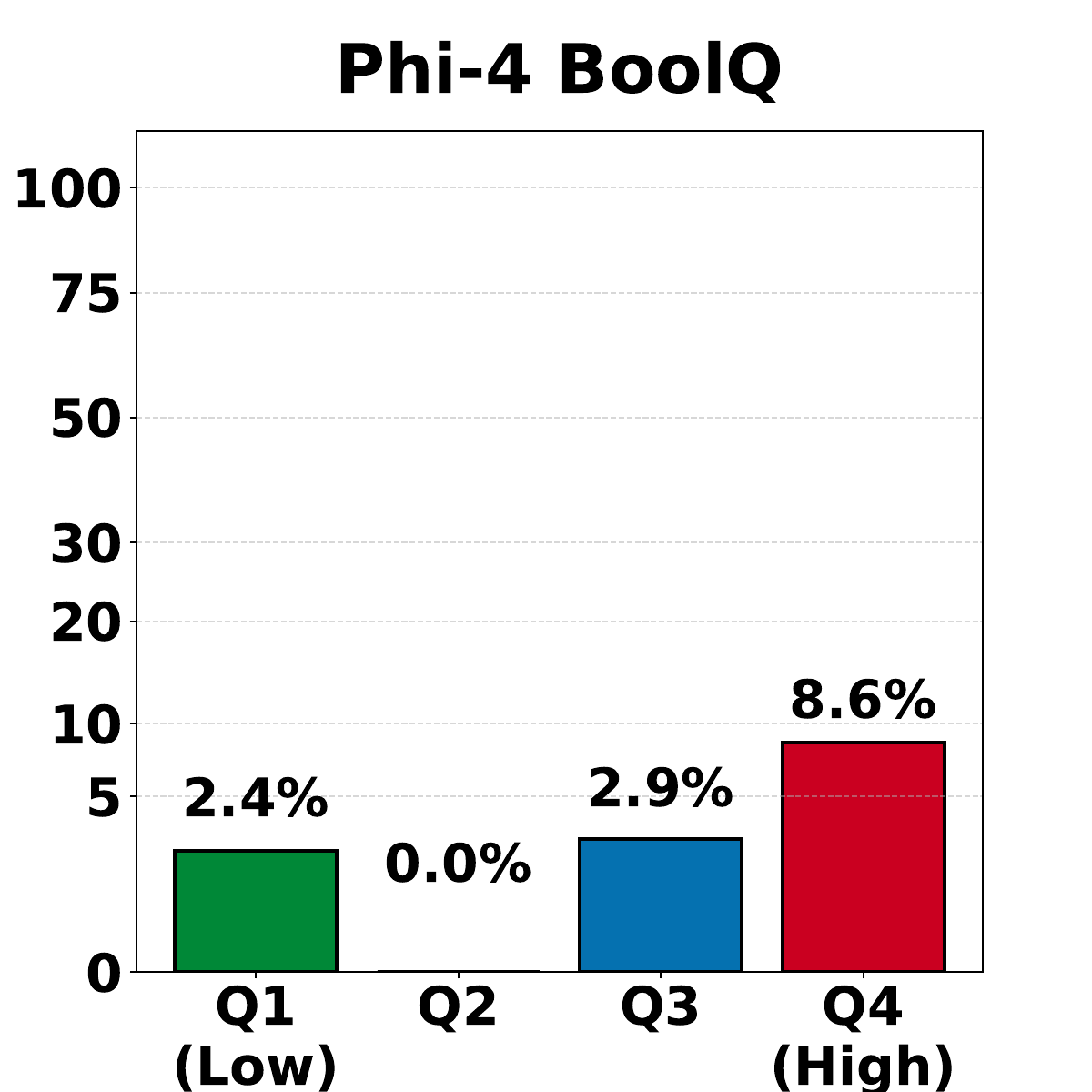}
            {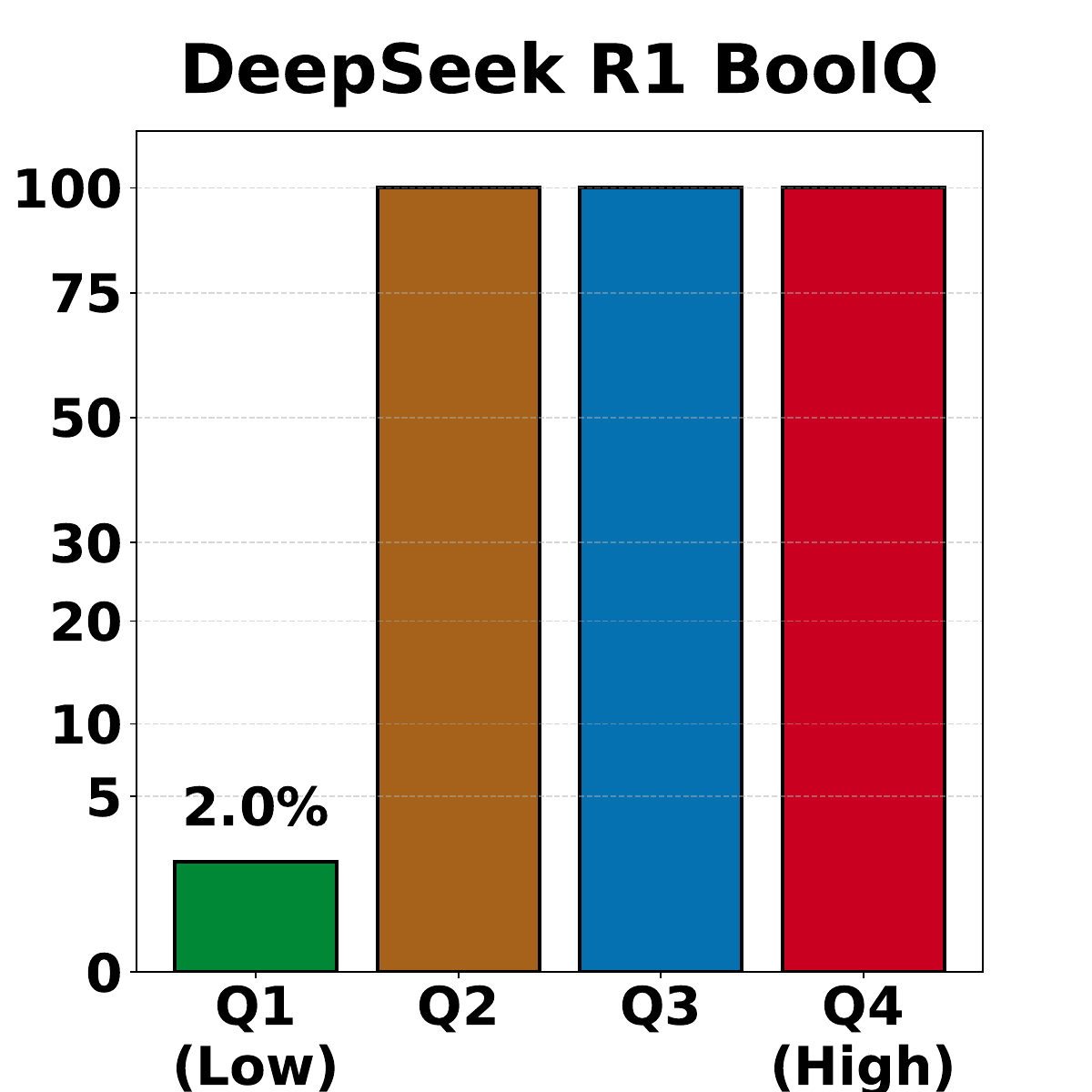}
        \entropyrow
            {StratQA}
            {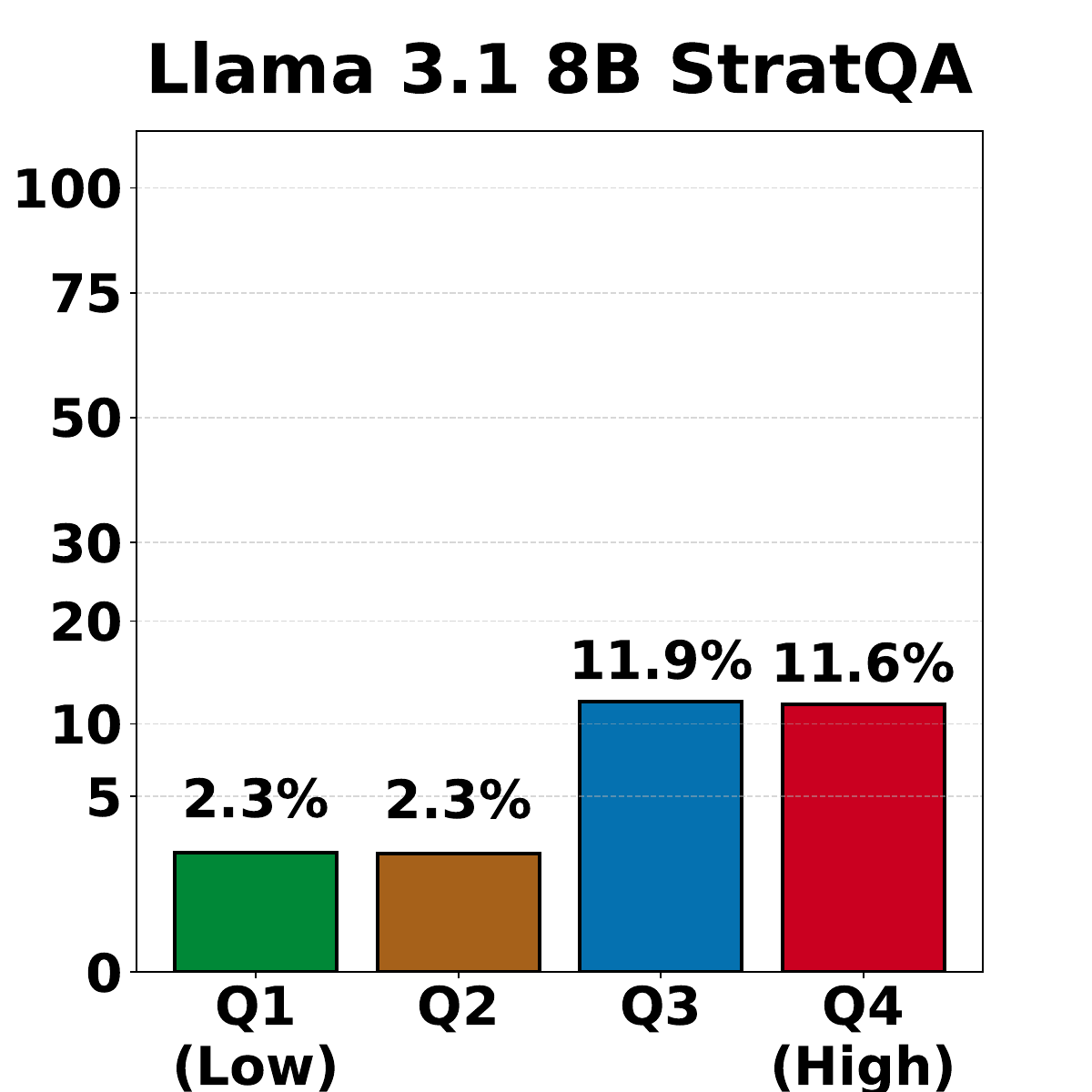}
            {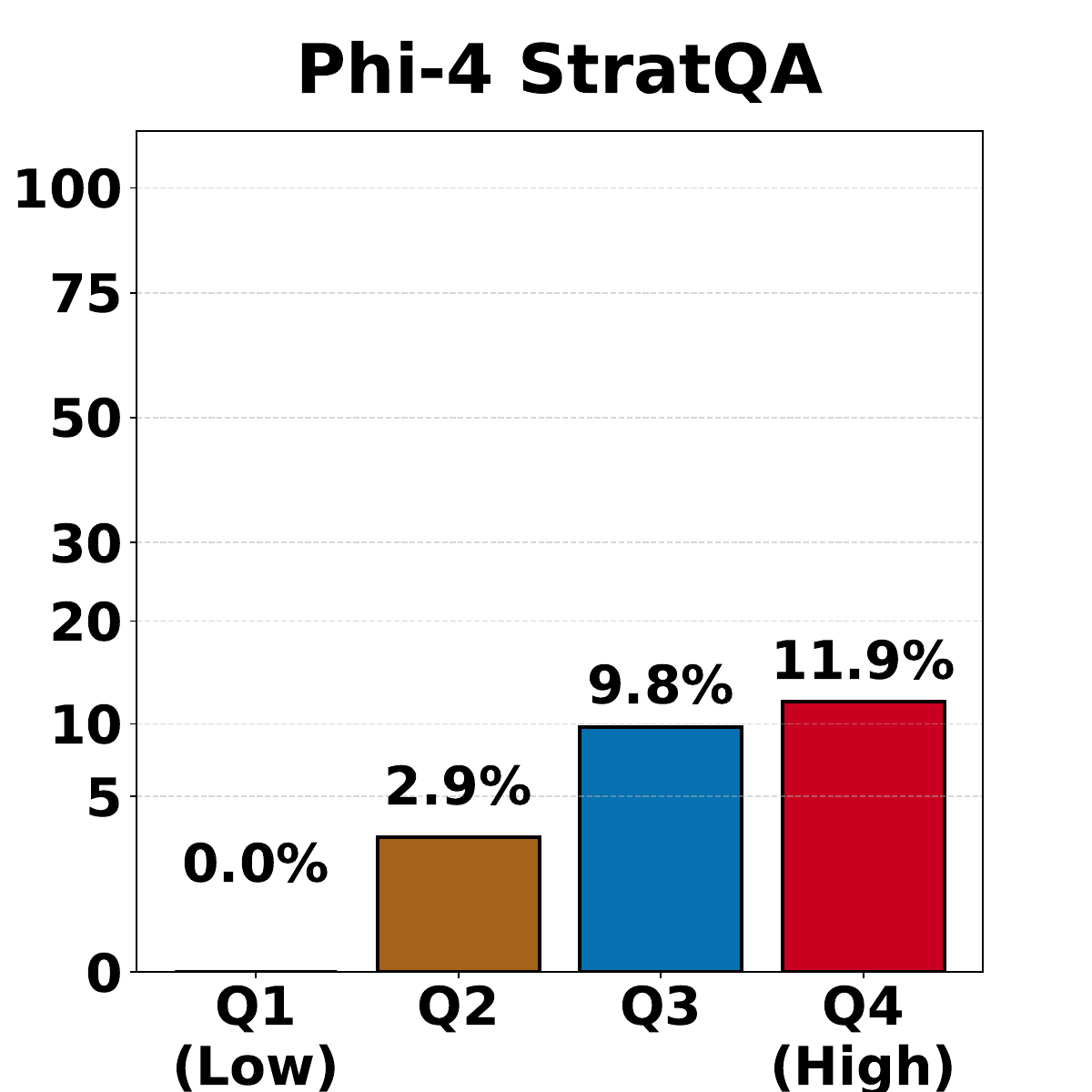}
            {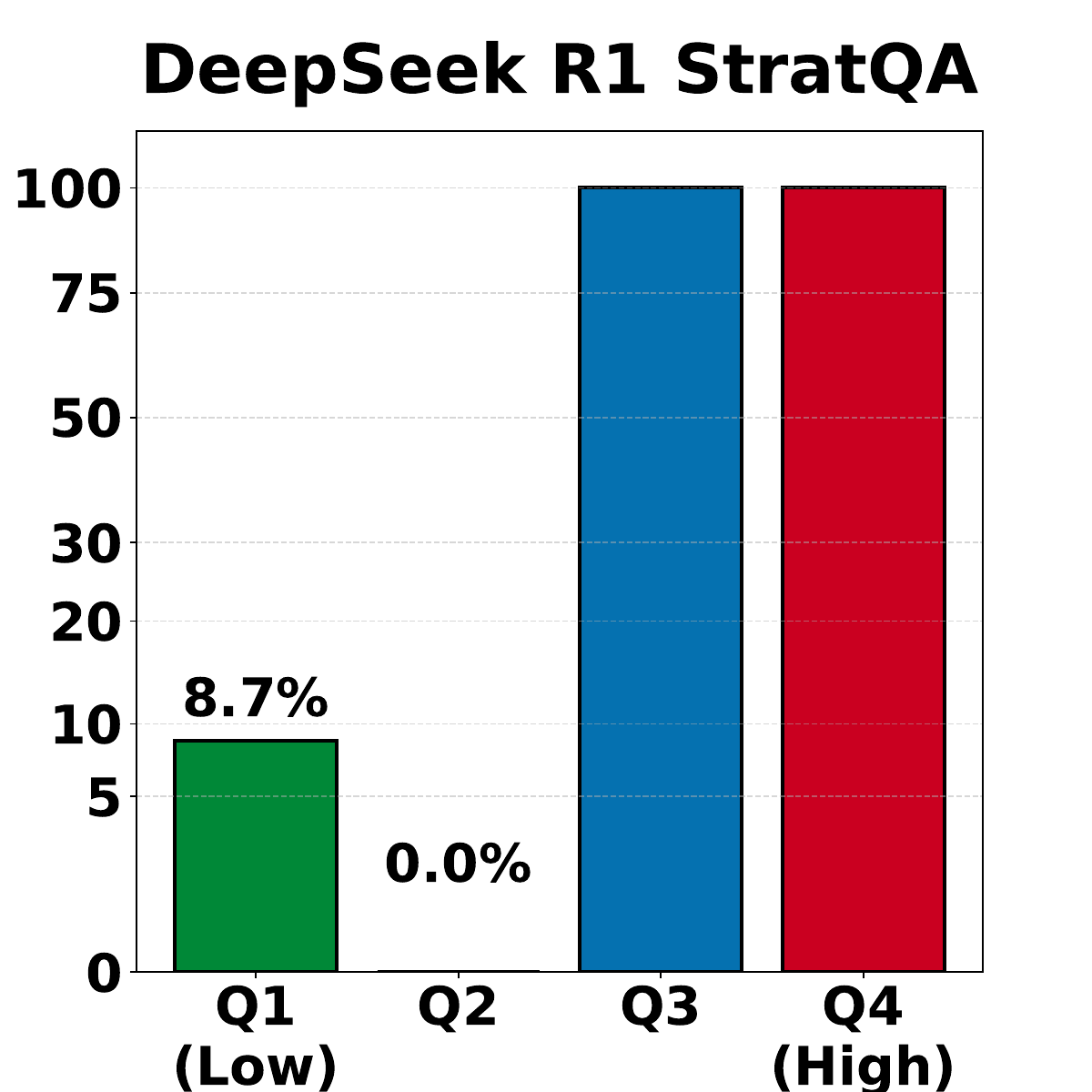}
        \vspace{-1mm}
    \end{minipage}%
    \hfill
    % ========================================================
    % RIGHT HALF
    % ========================================================
    \begin{minipage}[t]{0.495\textwidth}
        \centering
        \entropyrow
            {MedQA}
            {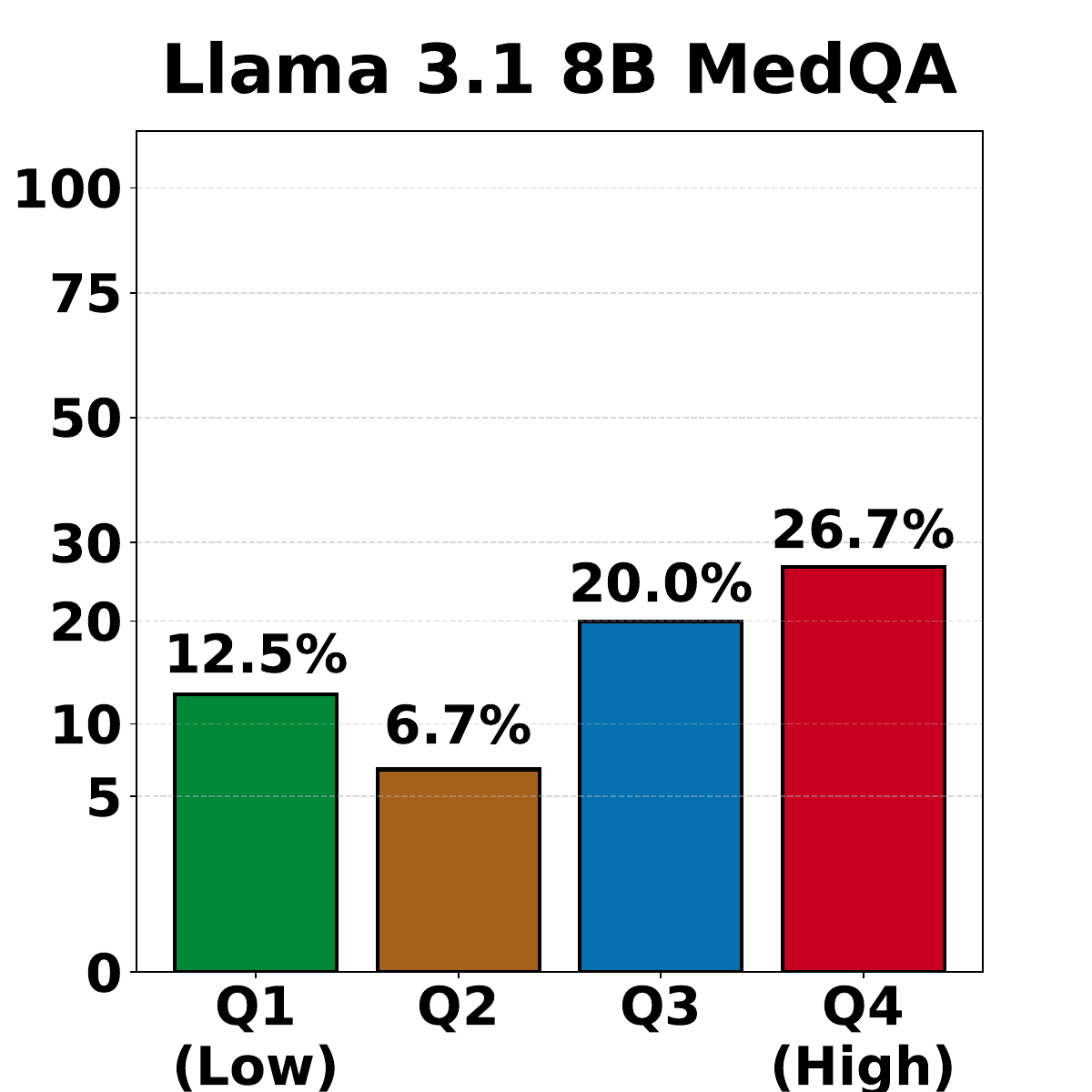}
            {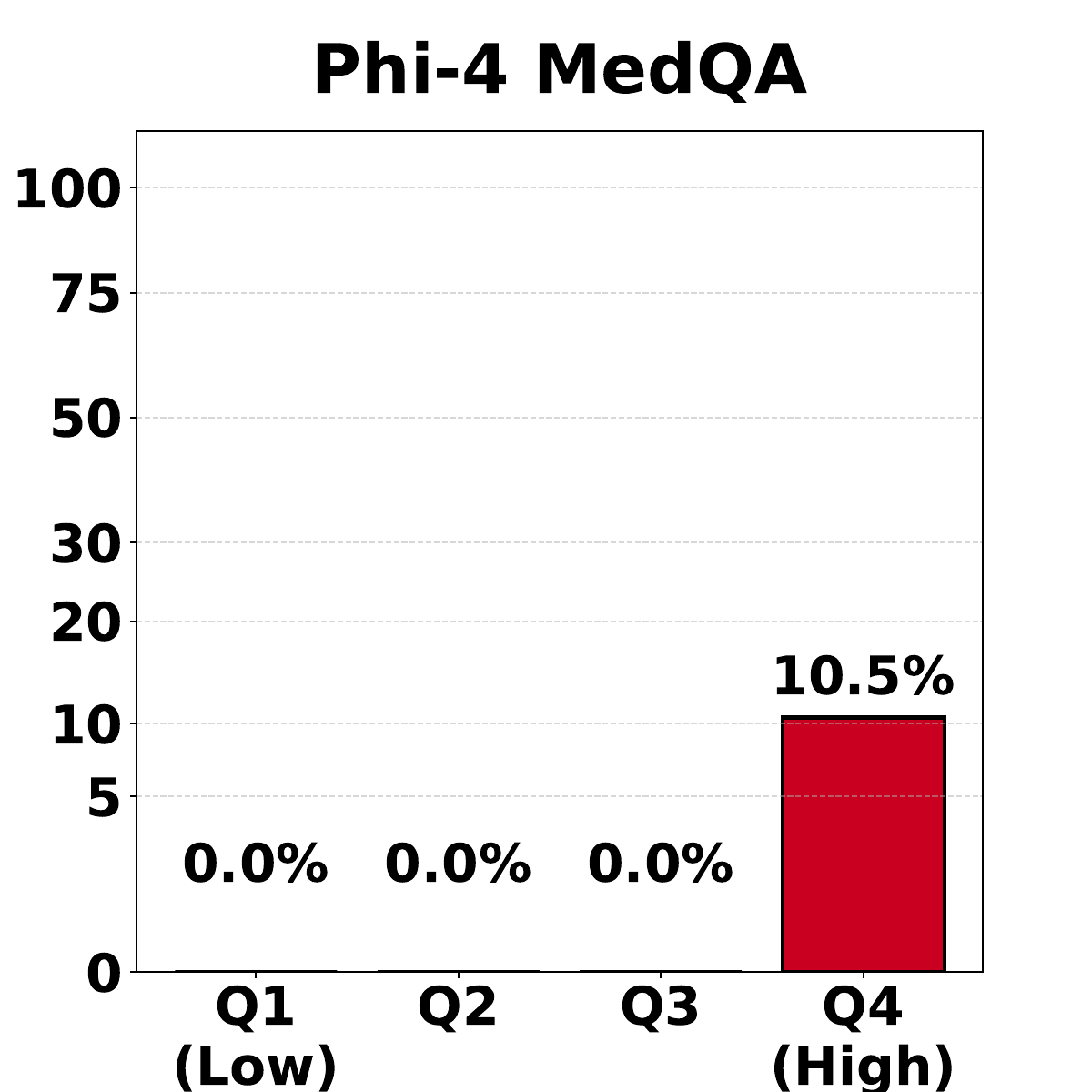}
            {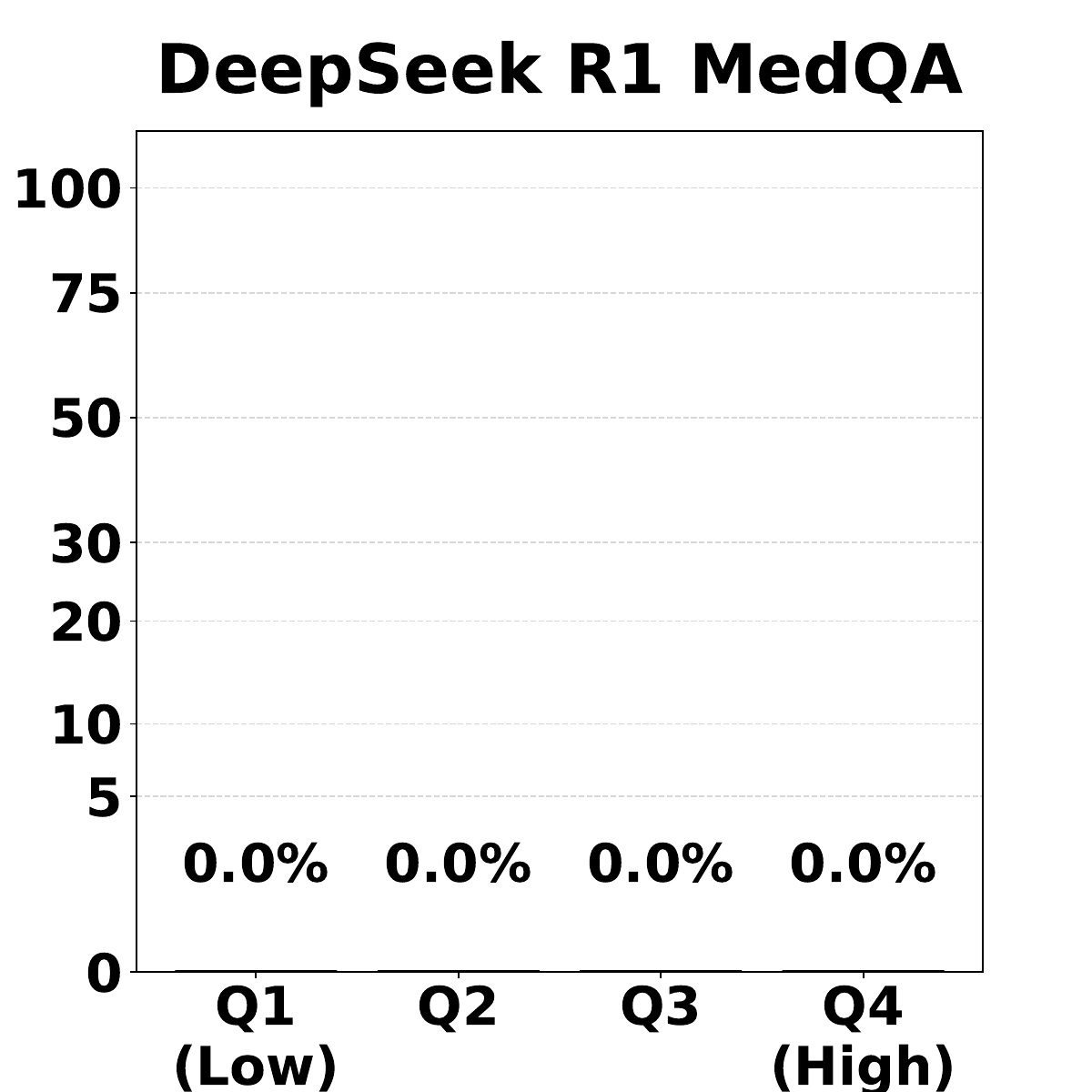}
        \entropyrow
            {GPQA}
            {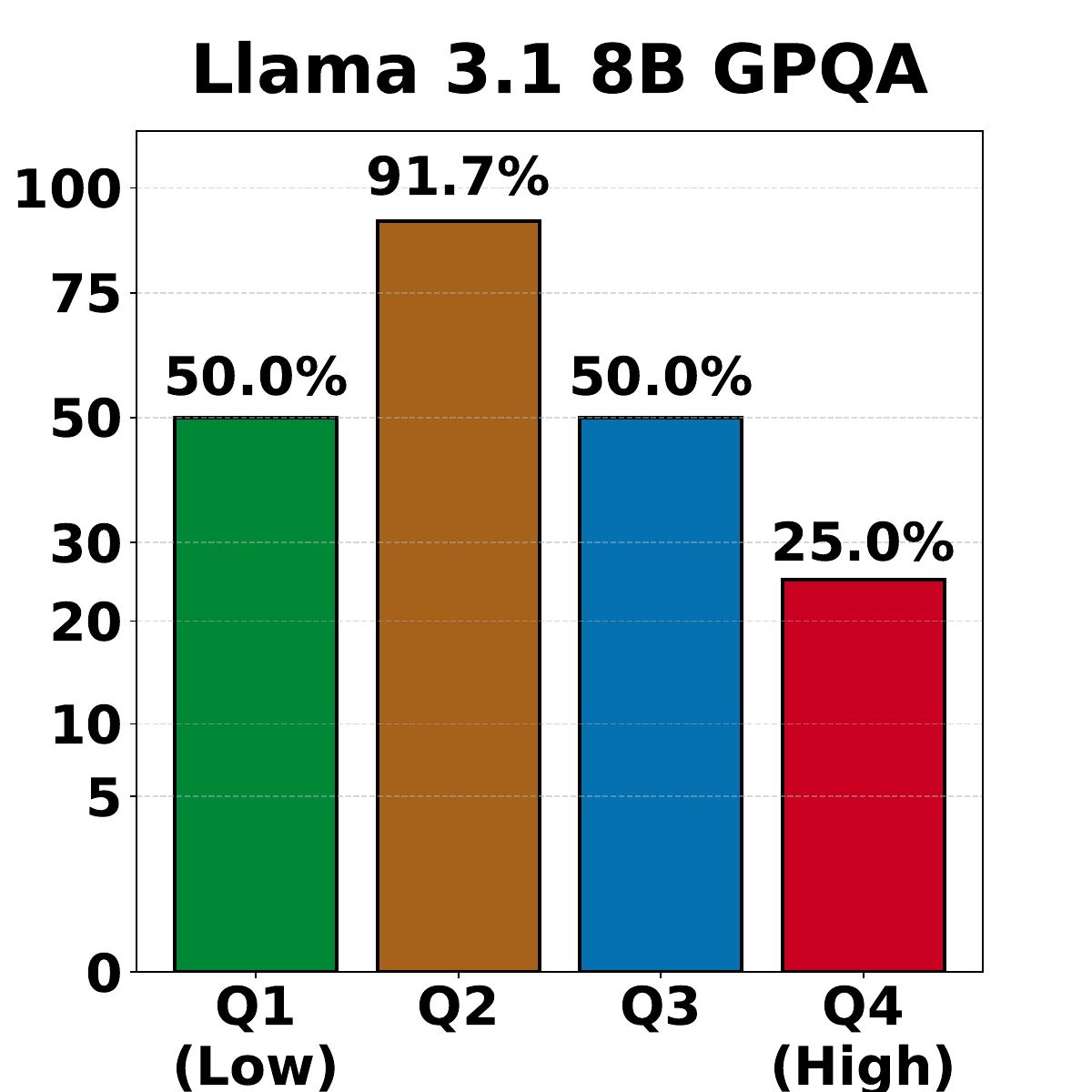}
            {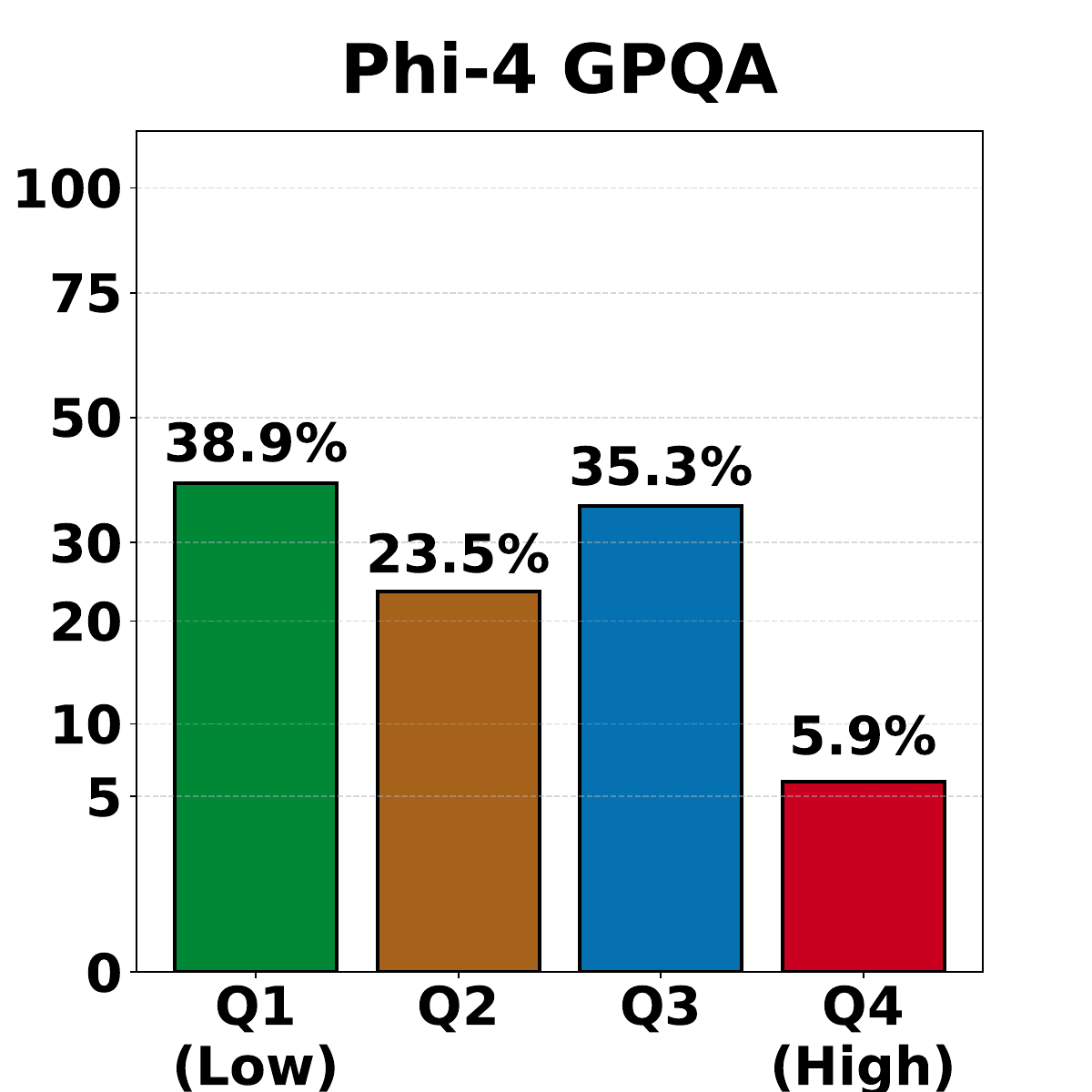}
            {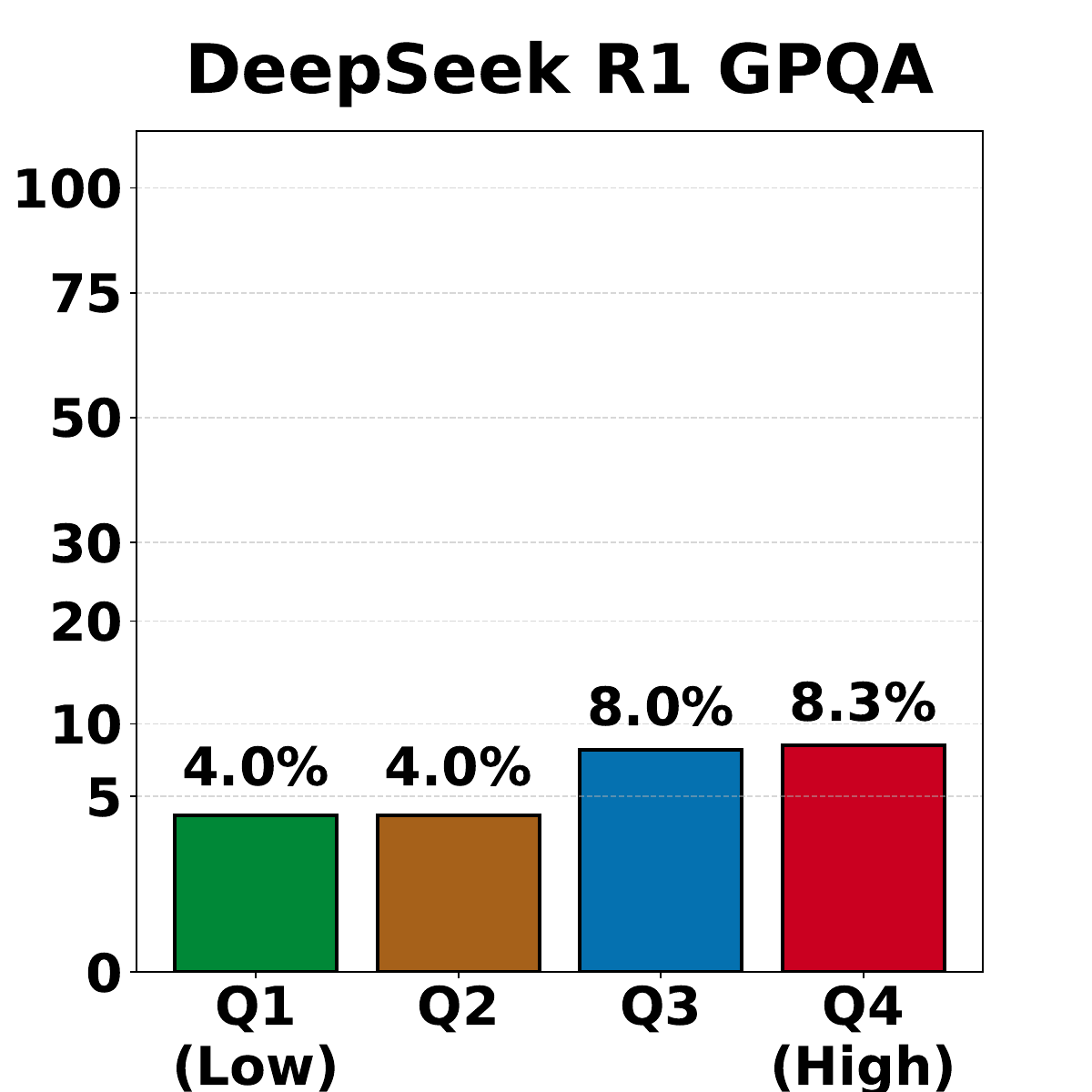}
        \entropyrow
            {AIME '26}
            {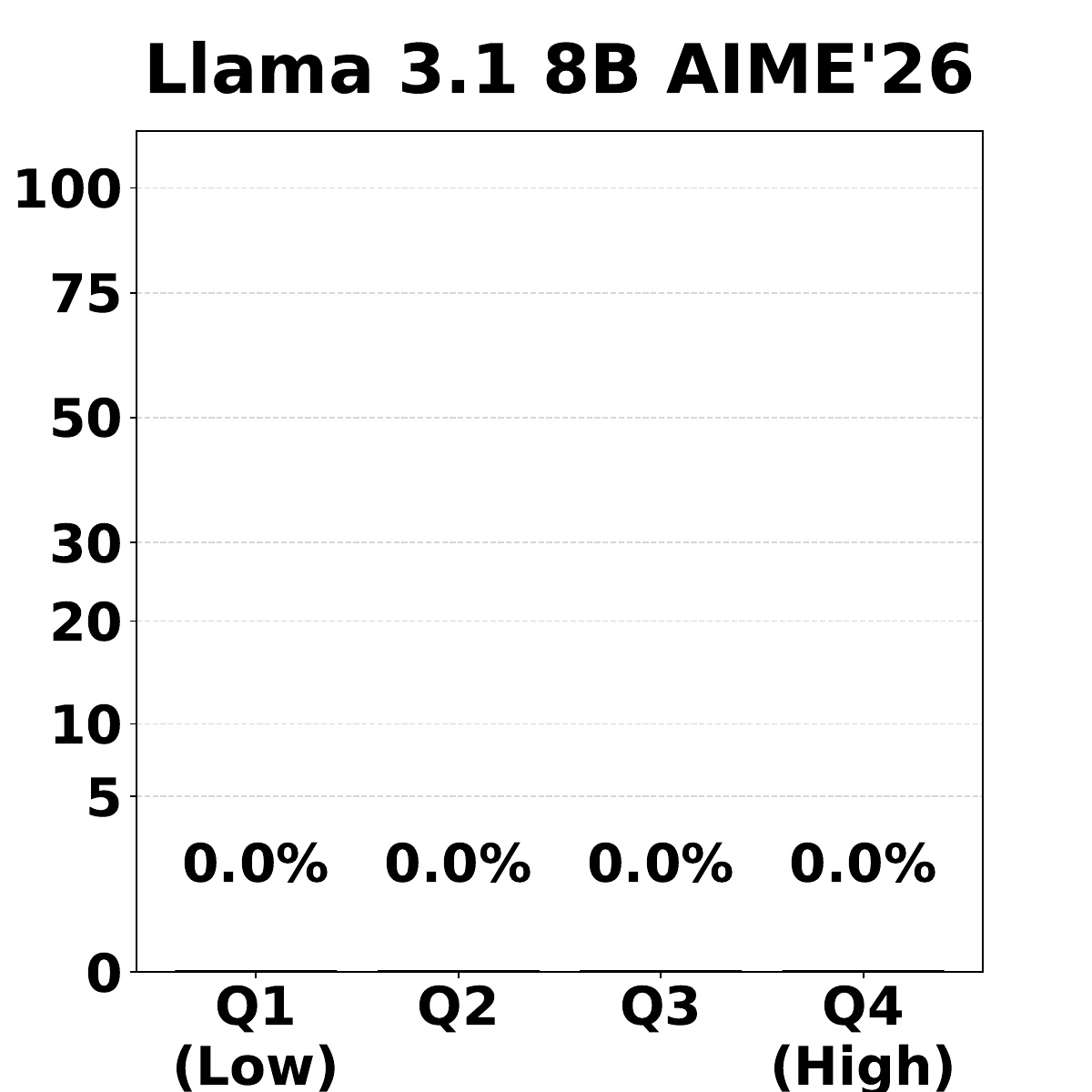}
            {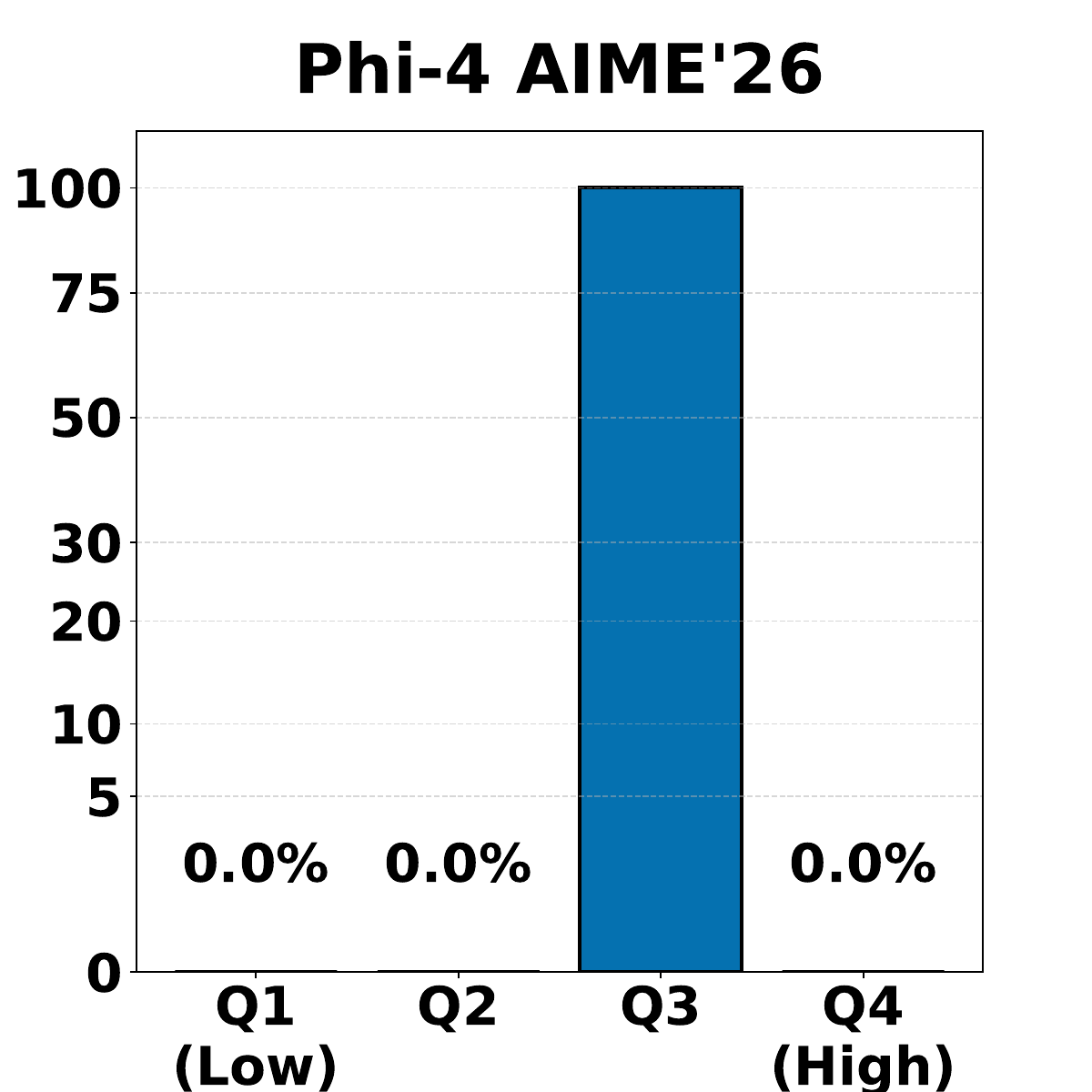}
            {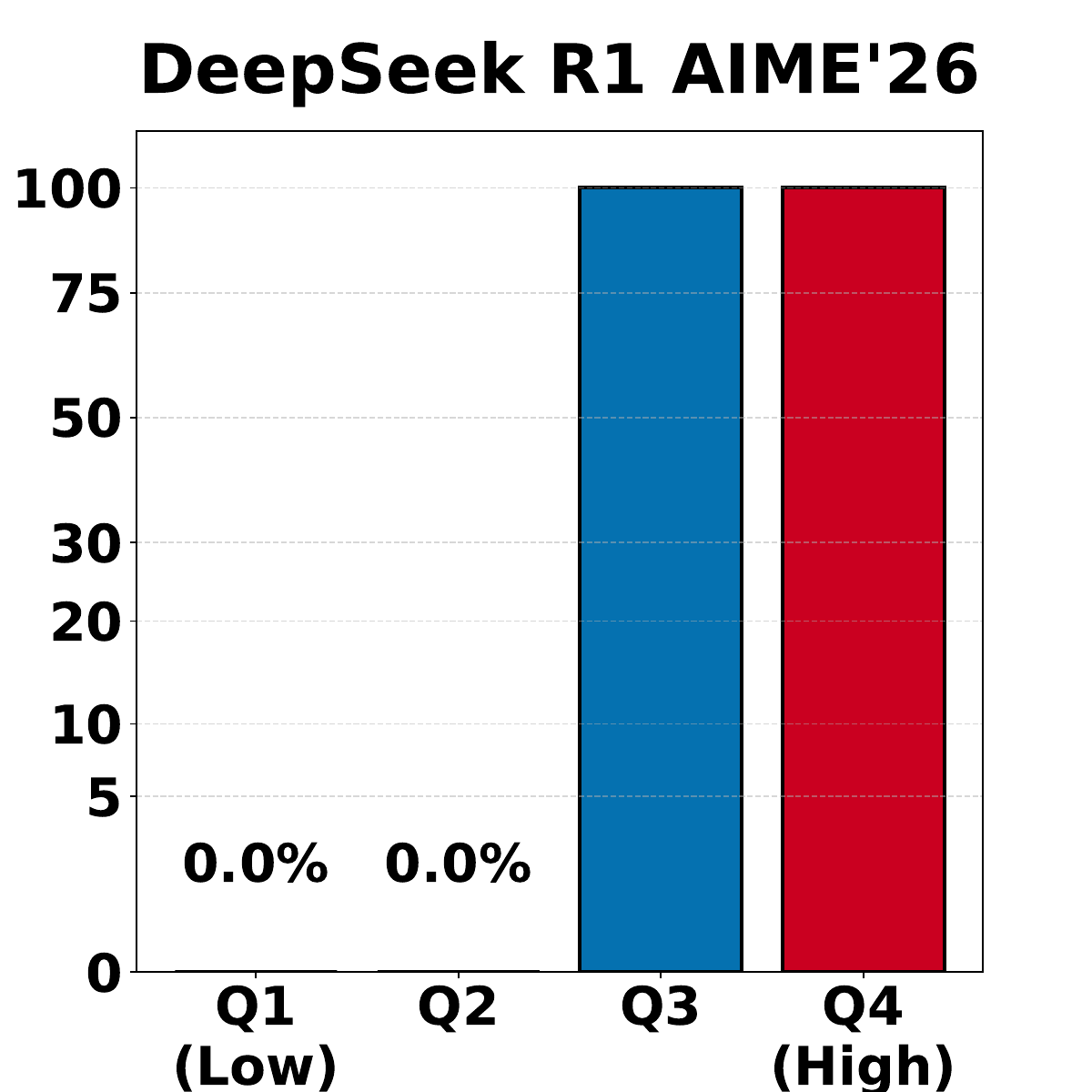}
        %\vspace{2mm}
    \end{minipage}
    \caption{
        \textbf{GSC discard rate by reasoning uncertainty.}
        Each bar chart reports, for a given GLLM and dataset, the fraction of SC CoTs that GSC discards, divided into quartiles of conditional reasoning entropy retained by each reasoning graph,
        $H(G \mid Y)$. %over the semantic-structural clusters of those chains. 
        Higher entropy quartiles
        indicate less structural agreement among the reasoning traces that reach the majority answer. The Y axis uses a monotonic sqrt scale to keep the lower Discard Rates readable without letting the few $100\%$ outliers flatten everything.
        %The number of underlying questions reflects our experimental setup (\Cref{sec:exp_setup}). 
        %For
        %AIME '26, quartiles are computed over fewer points because the dataset is composed of just $30$
        %questions. When a model has $0\%$ self-consistency accuracy on a dataset ($\eg$, Llama-3.1 8B on AIME '26),
        %no SC chains exist to condition on, so the corresponding panel is %shown 
        %empty.
        The panel DeepSeek~R1 MedQA is empty because GSC retains 100\% of answers.  The panel Llama-3.1~8B on AIME'26 is empty because the GLLM has 0\% SC accuracy on AIME'26, so no SC chains exist to condition on.
    }
    \label{fig:entropy_retention_all_datasets}
\end{figure*}

\subsection{%Shifting Decoding from SC to GSC
Comparison of GSC and SC in Accuracy and Uncertainty
%Testing for Real and Faithful Accuracy Leveraging Structural and Causal Awareness
}
\label{sec_rq2_all}

%\emph{To assess the impact of enforcing semantic–structural coherence during decoding}, 
\emph{To assess the impact of shifting %the decoding paradigm from 
decoding from outcome-based frequency to semantic-structural reasoning consensus,} we compare Self-Consistency (SC) with Graph Self-Consistency (GSC). \Cref{fig:faithful_accuracy_heatmap} %shows 
reports the outcome-based %nominal 
SC accuracy, SS-GED-based GSC accuracy, and %of both methods and 
%the performance 
their difference ($\Delta$(GSC $-$ SC)). % across the same GLLMs and datasets of increasing complexity %used in 
%(\cf,~\Cref{sec:exp_setup}). 
%\emph{These results are crucial for %both in the %It is crucial to frame these results not only as a 
%pursuit of 
%raw accuracy maximization and %to %, but also as a 
%filter for reasoning fidelity in mission critical settings}. %Standard SC selects answers based only on output frequency, which can reward “lucky guesses,” where flawed reasoning happens to produce the correct answer. In contrast, GSC routes through the medoid reasoning graph, penalizing these unfaithful shortcuts and favoring coherent reasoning.
%Standard 
%These results are crucial for framing evaluation not merely as the pursuit of raw accuracy maximization, but as a strict filter for \emph{reasoning fidelity} in mission-critical settings.
The objective of this experiment is to move beyond raw accuracy and test GSC as a filter for \emph{reasoning fidelity} in mission-critical settings.
%SC selects the most frequent answer, which can reward cases where incorrect or unfaithful reasoning paths happen to converge on the correct output. GSC instead routes through the medoid reasoning graph, favoring answers supported by the structurally central and semantically coherent CoT. %We can observe that the tradeoff between nominal SC accuracy and reasoning fidelity is especially visible 
The tradeoff between SC accuracy and reasoning fidelity is most pronounced for the smallest model, Llama 3.1 8B, which shows significant drops across all datasets, including -$7.6$ points on GSM8K and -$9.0$ on MedQA ($p<0.05$, McNemar's test~\cite{mcnemar1947}).
%As we show in \Cref{sec:decoding_res}, %this steep drop exposes the \emph{illusion of accuracy in smaller LLMs}: many SC “successes” come from chaotic, unfaithful reasoning paths.
This gap suggests that many apparent SC successes in smaller models are supported by unstable or unfaithful reasoning paths. GSC removes these cases, revealing a lower but more faithful measure of the model’s reasoning ability. Interestingly, this penalty largely disappears for more capable models. %For Phi-4, drops are minimal on StratQA (-0.5) and GPQA (-2.0), and performance even improves on BoolQ (+2.5) and MedQA (+2.0). This pattern continues with DeepSeek R1, which %successfully 
%maintains or slightly improves %nominal 
%accuracy across all datasets. 
For Phi-4, drops are minimal, and accuracy even improves with respect to SC on BoolQ (+$2.5$) and MedQA (+$2.0$). This pattern continues with DeepSeek R1, which reports 5-out-of-6 accuracy improvements relative to SC.
% but GPQA, which is the %hardest one and for which the model is more prone to produce hallucinations.
%most demanding benchmark. %Since powerful models are naturally less prone to hallucinations, the structurally central medoid path selected by GSC more confidently aligns with the correct answer. 
These behaviors suggest that for more capable models, GSC is more likely to select a %structurally 
central path that also leads to the correct answer, reducing the trade-off between faithfulness and accuracy.

\emph{To empirically prove that the accuracy reductions for smaller GLLMs are a rejection of unfaithful reasoning}, we %must 
analyze the conditional reasoning %entropy 
uncertainty retained in GSC graphs, $H(G\mid Y)$. %\rreva{NO! HERE PUT THE ENTROPY OF G GIVEN Y FOLLOWING NEW THEOREM FORMULATION, REMOVE ALLA HATS}$\hat{H}(Z\vert{}Y)$. 
%As per %theoretically proved in 
As per~\Cref{thm:supremacy} (\Cref{sec:gsc_proof}), %SC is blind to this conditional reasoning entropy, whereas GSC captures it to filter out unfaithful predictions. %We mainly focus this analysis on GSM8K and MedQA, as these datasets exhibited the highest accuracy drops for Llama 3.1 8B in \Cref{fig:exp3_fig}. 
SC is insensitive to
within-answer reasoning variation, whereas GSC operates on
$G=\Gamma(Z)$ and uses the conditional information retained in
$H(G\mid Y)$, where
$0\leq H(G\mid Y)\leq H(Z\mid Y)$.
\Cref{fig:entropy_retention_all_datasets} plots the percentage of 
SC answers that GSC %keeps 
discards (``Discard Rate"), grouped %by their conditional reasoning entropy ($\hat{H}(Z\vert{}Y) = -\sum p(z\vert{}y) \log p(z\vert{}y)$) which measures the divergence of reasoning paths ($Z$) leading to the same final answer ($Y$) (\Cref{sec:gsc_proof}). Answers, and their reasoning paths, are divided into quartiles (Q1 to Q4) of this entropy. Because entropy measures %structural 
into
quartiles %according to their 
of increasing conditional graph-level uncertainty, from Q1 (lowest) into Q4 (highest).
%uncertainty, 
%Q1 reflects low reasoning uncertainty, while Q4 reflects high uncertainty.%and unfaithful guessing. 
%, where high inconsistency is a sign of unfaithful guessing rather than valid alternative solutions. The results reveal a clear filtering trend. 
High uncertainty is symptomatic of unfaithful guessing rather than valid alternative solutions. %\bem{GSC discards the highest percentage of answers in the highest-entropy quartiles, where uncertainty peaks}. %As entropy increases in less capable GLLMs, GSC discards significantly more uncertain answers ($\eg$, %dropping Llama 3.1 8B's retention from 95.9\% (GSM8K) and 87.5\% (MedQA) in Q1 to 81.9\% (GSM8K) and 73.3\% (MedQA) in Q4) 
%18.1\% in Q4-GSM8K and 26.7\% in Q4-MedQA for Llama 3.1 8B)
The results reveal a prominent filtering trend.
GSC discards the largest proportion of
answers in the highest-uncertainty quartiles.
%For example, 
For Llama 3.1 8B, the Q4
Discard Rate reaches $\mathbf{18.1}$\% on GSM8K and $\mathbf{26.7}$\% on MedQA. %,
%whereas SC retains $100$\% of these predictions across all quartiles, remaining blind to this conditional uncertainty. 
SC retains these predictions by
construction because its decision is invariant to within-answer
reasoning variation. This behavior is not restricted to less capable GLLMs. DeepSeek R1 exhibits near-zero Discard Rate %when its correct answers are supported by low-entropy reasoning, as on MedQA and GPQA,
for low-uncertainty
answers on MedQA and GPQA,
but GSC discards \textbf{100\%} of SC answers in %multiple high-entropy 
several high-uncertainty quartiles on BoolQ, StratQA, and AIME'26 datasets.
\bem{These results %indicate that %the
suggest that
%performance differences %in \Cref{fig:faithful_accuracy_heatmap} 
%arise %primarily 
%because 
GSC
%rejects answer agreement supported by structurally inconsistent reasoning,
%while generally preserving correct answers backed by coherent reasoning
%paths.}
rejects answers associated with higher uncertainty, while preserving those %backed 
substantiated by consistent reasoning paths}.

\begin{figure*}
    \centering
    \includegraphics[width=\linewidth]{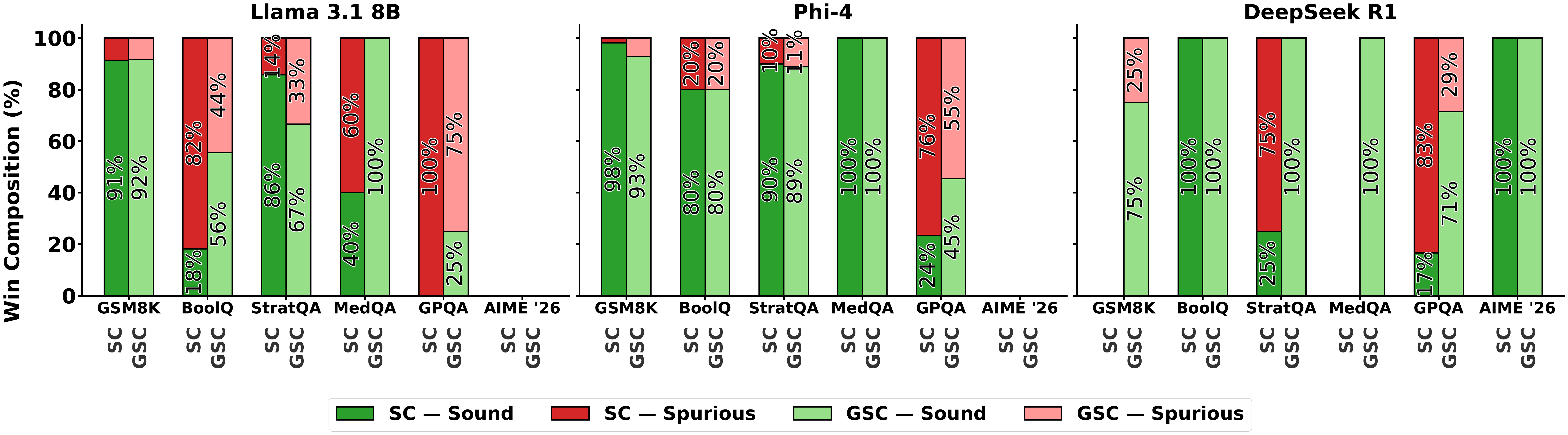}
    \caption{\textbf{Deductive Soundness of GSC-SC Disagreements} 
    Each stacked bar shows the fraction %share 
    of \textit{sound} (green) and \textit{spurious} (red) wins within that disagreement cell.
    GSC-only wins are usually more sound than SC-only wins. %, though the gap varies by model and dataset. 
    %On LLaMA 3.1 8B, GSC raises the faithful share from 40\% to 100\% on MedQA, while SC-only wins on GPQA are entirely spurious. On Phi-4, the largest gap appears on GPQA (45\% vs.\ 24\% faithful in favor of GSC). On DeepSeek R1, GSC-only wins are fully faithful on StrategyQA and MedQA, while SC-only wins are much less faithful on StrategyQA (25\%) and GPQA (17\%). 
    Overall, across 18 comparable disagreement settings, %(\ie, with both SC-only and GSC-only wins), 
    GSC has a higher soundness share in $9$ ($\mathbf{50\%}$), ties in $6$, and is lower in $3$.}
    \label{fig:gsc_sc_verification_rate}

    %\vspace{-6mm}
\end{figure*}

\subsection{Deductive Soundness of GSC-SC Disagreements}\label{sec:decoding_res}

%\emph{To rigorously validate that GSC filters unfaithful reasoning rather than simply acting as a noisier decoding strategy}, we conduct a hybrid qualitative audit on the subset of questions where SC and GSC disagree. Specifically, we focus on two complementary disagreement cells: SC-only wins (SC correct, GSC wrong) and GSC-only wins (GSC correct, SC wrong). By auditing both directions of disagreement symmetrically, we can test whether the divergence between the two methods is systematically meaningful, that is, whether GSC's rejections correspond to unfaithful reasoning and GSC's unique recoveries correspond to genuinely sound reasoning. As detailed in \Cref{sec:audit_methodology}, we randomly sampled up to N=50 paths for each pair of (model, dataset), and then anonymized and shuffled them prior annotation to mitigate author bias. Along with the human annotator, we also used a Gemini 3.1 Pro as LLM-as-judge. 
%To validate this approach, we calculated the inter-rater reliability between the human annotator and the LLM on a blind subset of 50 reasoning paths. The human and the LLM achieved a high Cohen's kappa coefficient $\kappa = \mathbf{0.913}$, demonstrating that the automated judge highly aligns with human reasoning in distinguishing robust deduction from unfaithful, spurious successes.
%\emph{To further determine whether GSC improves reasoning fidelity and serves as an %is 
%effective %as a 
%decoding filter}, 
\emph{To evaluate the deductive soundness of SC--GSC disagreements}, %we audit the subset of questions on which SC and GSC disagree, 
%partitioning them into \textit{SC-only wins} and \textit{GSC-only wins}. A SC prediction is classified as \emph{faithful} %if the paths supporting its majority vote are predominantly logically sound. 
we audit questions for which the two decoding strategies %rules select different
select different answers, partitioned into \textit{SC-only} and \textit{GSC-only wins}.
%if the majority of paths supporting its vote are logically sound. A GSC prediction is \emph{faithful} if the selected medoid is logically sound. This disagreement set is especially informative because it captures the precise decisions introduced by the decoding rule itself. %(see Appendices~\ref{sec:decoding_res_path_appendix} and \ref{sec:cases_appendix} for path level analysis and qualitative case studies). %For a more granular audit at the \emph{path level}, refer to Appendix~\ref{sec:decoding_res_path_appendix}. 
%Following \Cref{sec:audit_methodology}, 
Using DS ($\cf$, \Cref{sec:exp_setup}), an SC
prediction is sound if the majority of CoTs supporting its selected
answer satisfy $\operatorname{DS}(z)=1$. A GSC prediction is sound if
its selected medoid $z^\star$ satisfies
$\operatorname{DS}(z^\star)=1$. Predictions that do not satisfy these
conditions are classified as spurious.
%We annotate sampled disagreement traces with both a human reviewer and Gemini 3 Pro~\cite{gemini2025family} as LLM judge~\cite{gu2026llmjudge}. The two annotators achieve strong agreement, with Cohen's $\kappa = \mathbf{0.913}$ ~\cite{cohen1960kappa}, supporting the reliability of the audit %and the %full 
%disagreement analysis 
%performed with the %validated 
%LLM judge. 
A human annotator and Gemini~3~Pro~\cite{gemini2025family}, used as
an LLM-as-judge, independently evaluate the
disagreement traces. Their strong agreement
($\kappa=0.913$) supports the reliability of the audit~\cite{cohen1960kappa}.
%\Cref{fig:gsc_sc_verification_rate} visualizes the percentage of wins that are \textit{faithful} versus \textit{spurious} for each disagreement cell: green %segments 
%denotes faithful wins and red %segments 
%denotes spurious wins. %, and the percentages shown inside each bar report the composition of that cell. %A clear pattern emerges, dependent on model \emph{scale} and task \emph{difficulty}. 
\Cref{fig:gsc_sc_verification_rate} reports the proportion of deductively sound (green) and spurious (red) wins in each disagreement cell.
For Llama 3.1 8B, many SC-only wins are %not supported by faithful reasoning, 
spurious, especially on harder tasks. On MedQA, only $40$\% of SC-only wins are %faithful, 
sound, while $60$\% are spurious. In
contrast, $100$\% of GSC-only wins are %faithful. 
sound. On GPQA, SC-only wins are entirely spurious, whereas GSC recovers $25$\% soundness. %BoolQ also shows severe SC spuriousness at $18$\%, which GSC significantly improves to $56$\%. This suggests that a substantial fraction of SC's apparent advantage in smaller models comes from answer-level success without reasoning fidelity. 
On BoolQ,
the soundness share increases from $18\%$ for SC to $56\%$ for GSC. This suggests that part of SC's apparent advantage on smaller GLLMs arises from
correct answers reached through deductively unsound reasoning.
For Phi-4, SC %, 
and GSC %the two methods 
are %much 
closer on easier benchmarks, indicating that as reasoning capacity %quality 
improves, the gap between spuriousness % nominal 
and soundness %correctness 
%accuracy 
narrows. %However, there is again a clear distinction on harder tasks. 
%On GPQA, the faithful share rises from $24$\% for SC-only wins to $45$\% for GSC-only wins, reducing the spurious fraction from $76$\% to $55$\%. 
%The same trend is even sharper 
%For DeepSeek R1, on StratQA, only $25$\% of SC-only wins are faithful, compared with $100$\% of GSC-only wins. %On GSM8K, GSC-only wins are 75\% faithful, with no SC-only wins. On MedQA, all audited GSC-only wins are faithful, again with no SC-only wins. In other words, for GSM8K and MedQA, every disagreement between the two decoding rules favors GSC. 
%For GSM8K and MedQA, SC produces no unique wins, so the comparison reduces to whether GSC's unique wins are faithful, which they are at $75$\% and $100$\%, respectively.
On GPQA,
however, the soundness share increases from $24\%$ for SC-only wins to
$45\%$ for GSC-only wins. For DeepSeek~R1 on StratQA, only
$25\%$ of SC-only wins are sound, compared with $100\%$ of
GSC-only wins. On GSM8K and MedQA, SC produces no unique wins,
while $75\%$ and $100\%$ of GSC-only wins are sound, respectively.
BoolQ is an exception in which both Phi-4 and DeepSeek R1 achieve comparable soundness, consistent with the task being less demanding for larger models.\ %bem{These results indicate that when the two decoding rules disagree, GSC typically selects the more trustworthy answer, particularly in more complex domains (MedQA, GPQA), where unfaithful majority voting is most costly}.
%These findings align with GSC’s design. 
%By %selecting the medoid 
%using structural centrality %$C(G_i)$ 
%and path consensus %$P(G_i)$ 
%(\Cref{sec:methodology}), GSC favors reasoning paths that are more structurally and semantically consistent %with the rest of the sample 
%along the complete reasoning process. %As shown in Appendix~\ref{sec:cases_appendix}, this may help reduce cases in which SC selects correct answers supported by inconsistent intermediate reasoning. %Since SC relies only on final-answer frequency, it does not directly use signals about reasoning consistency.
\bem{These results %indicate 
suggest that when the two decoding %rules 
strategies disagree, GSC 
%typically %selects the more trustworthy answer, particularly in more complex 
%domains (MedQA, GPQA), where unfaithful majority voting is most costly}. 
favors paths that are more likely to satisfy
deductive soundness}.
%Averaged 
Averaged across all disagreement cells, GSC achieves $71.8$\% %faithful 
sound wins versus $46.9$\% for SC (\textbf{GSC wins by +$\mathbf{24.9}$ percentage points}, $p < 0.05$, Wilcoxon signed-rank test~\cite{wilcoxon1945}),  and performs at least as well in
$15/18$ cells ($83.3\%$), with $9$ wins, $6$ ties, and $3$ losses.

\newcommand{\risklow}[1]{\cellcolor{red!8}#1}
\newcommand{\riskmed}[1]{\cellcolor{red!16}\textbf{#1}}
\newcommand{\riskhigh}[1]{\cellcolor{red!28}\textbf{#1}}

\begin{table}[h]
    \centering
    \caption{
    %\rreva{put aime results when done}
        \textbf{Ablating the GSC medoid degrades faithfulness}
        %Percentage of answers for which accuracy remains unchanged while
        %reasoning faithfulness decreases, i.e.,
        %$\Delta\mathrm{Acc}=0$ and $\Delta\mathrm{Faith}<0$.
        %Both columns 
        The columns report cases in which final-answer accuracy remains unchanged after medoid ablation ($\Delta\mathrm{Acc}=0$).
        The first reports the percentage of all evaluated answers for which
        faithfulness decreases ($\Delta\mathrm{Faith}<0$). The second reports the percentage of all answers
        suffering a faithfulness drop of $\geq$ 50\%.
        Light-red cells indicate degradation rates of $\geq$ 20\%, while
        dark-red cells indicate rates of $\geq$ 40\%.
    }
    \label{tab:silent_faithfulness_degradation}
    \resizebox{0.85\columnwidth}{!}{%
    \begin{tabular}{llcc}
        \toprule
        \textbf{GLLM}
        & \textbf{Dataset}
        & \shortstack{\textbf{$\Delta\text{Faith} < 0$}\\
                      \textbf{(\% All Answers)}}
        & \shortstack{\textbf{Faith Drops $\geq$ 50\%}\\
                      \textbf{(\% All Answers)}} \\
        \midrule

        \multirow{6}{*}{\textbf{Llama 3.1 8B}}
        & GSM8K
        & \risklow{22.49\%}
        & \textit{2.00\%} \\

        & BoolQ
        & \risklow{37.50\%}
        & 2.00\% \\

        & StratQA
        & \riskhigh{43.23\%}
        & 7.86\% \\

        & MedQA
        & \risklow{34.00\%}
        & 16.00\% \\

        & GPQA
        & \risklow{35.03\%}
        & 16.75\% \\

        & AIME '26
        & 16.78\%
        & 0.00\% \\
        \midrule

        \multirow{6}{*}{\textbf{Phi-4}}
        & GSM8K
        & \risklow{35.89\%}
        & \textit{7.70\%} \\

        & BoolQ
        & \risklow{20.50\%}
        & 2.10\% \\

        & StratQA
        & \riskhigh{54.95\%}
        & \risklow{21.29\%}\% \\

        & MedQA
        & \risklow{24.00\%}
        & 1.00\% \\

        & GPQA
        & \risklow{30.96\%}
        & 9.14\% \\

        & AIME '26
        & \risklow{26.78\%}
        & \risklow{23.30\%} \\
        \midrule

        \multirow{6}{*}{\textbf{DeepSeek R1}}
        & GSM8K
        & \riskhigh{46.89\%}
        & \textit{4.80\%} \\

        & BoolQ
        & \risklow{21.50\%}
        & 1.50\% \\

        & StratQA
        & 10.92\%
        & 2.18\% \\

        & MedQA
        & 16.13\%
        & 6.45\% \\

        & GPQA
        & \riskhigh{46.29\%}
        & \risklow{24.00\%} \\

        & AIME '26
        & \risklow{26.78\%}
        & \risklow{23.30\%} \\
        \bottomrule
    \end{tabular}%

    }
    \vspace{-4mm}
\end{table}

\subsection{GSC-selected Medoid Ablation Exposes Faithfulness Loss
%\dpal{START HERE}
}\label{sec:sec_rq3_broad_ablation}

%\emph{To assess whether the structural fragility exposed by the narrow stress test generalizes %beyond the marginal GSC-wins subset 
%to the full reasoning population}
\emph{To assess the %population-level 
robustness of the GSC-selected medoid}, we conduct an adversarial medoid ablation across all questions of all datasets ($\cf$, \Cref{sec:methodology}). %, as shown in \Cref{fig:scatter_all_datasets}. %Due to the page limit, 
This ablation reveals substantial faithfulness degradation even when final-answer accuracy remains unchanged.
On StratQA, for example, this degradation occurs %in 
for $\mathbf{43.2}$\% of %cases 
answers for Llama 3.1 8B and $\mathbf{55.0}$\% for Phi-4 (with up to $21.3$\% of %cases 
answers in which it drops by at least $50$\%). On GPQA, this occurs in $\mathbf{46.3}$\% of %cases 
answers for DeepSeek R1 (in which it drops by $\geq$ 50\% in %almost one out of four cases
up to $21.3$\% of answers) and $\mathbf{35}$\% for Llama $3.1$ 8B. \Cref{tab:silent_faithfulness_degradation} shows that this phenomenon is present across \textbf{all} (GLLM, dataset) configurations, and degrades faithfulness %of $11$\% to $55$\% of answers, depending on the configuration.
for up to $55\%$ (average: $30.59\%$) of answers
\Cref{fig:scatter_all_datasets}  visually quantifies these margins of faithfulness drops across GLLMs and datasets.
%\rreva{7/20: continue here}
Furthermore, %\Cref{fig:scatter_all_datasets} 
it shows that 
%The high negative slope %is downward slope 
%indicates that 
instances of %nominal 
accuracy \emph{improvement} under ablation ($\Delta \text{Accuracy}>0$) are coupled with \emph{drops} in reasoning faithfulness ($\Delta \text{Faithfulness}<0$, points placed in the $4^{th}$ quadrant). This behavior emerges in all charts %, with 
that show a regression line having a negative slope ($11$ charts out of $18$), and is prominent for Llama 3.1 8B and Phi-4. %and empirically captures the ``lucky guess'' phenomenon. 
This empirically exposes the ``lucky guess'' phenomenon: when forced off its primary load-bearing path, the GLLM may still reach the correct answer, but mainly through flawed or hallucinated intermediate reasoning rather than a valid alternative deduction.
\bem{These results show that the GSC-selected medoid is often essential to preserve faithful reasoning, especially in smaller GLLMs, where accuracy gains can coincide with reduced faithfulness.}
%Many questions preserve the final answer but lose faithfulness, and in smaller models, accuracy gains often mask reasoning collapse, revealing %unfaithful 
%lucky guesses.}
% ============================================================
% Preamble
% ============================================================

% These widths are relative to each half of the full figure.
\newcommand{\scatterlabelwidth}{0.075\linewidth}
\newcommand{\scatterwidth}{0.295\linewidth}

\newcommand{\scatterheaders}{
    \begin{minipage}[c]{\scatterlabelwidth}
        \mbox{}
    \end{minipage}%
    \begin{minipage}[c]{\scatterwidth}
        \centering
        \scriptsize\textbf{Llama-3.1 8B}
    \end{minipage}%
    \hfill
    \begin{minipage}[c]{\scatterwidth}
        \centering
        \scriptsize\textbf{Phi-4}
    \end{minipage}%
    \hfill
    \begin{minipage}[c]{\scatterwidth}
        \centering
        \scriptsize\textbf{DeepSeek R1}
    \end{minipage}

    \vspace{1mm}
}

% #1 = dataset name
% #2 = Llama plot
% #3 = Phi plot
% #4 = DeepSeek plot
\newcommand{\scatterrow}[4]{
    \begin{minipage}[c]{\scatterlabelwidth}
        \centering
        \rotatebox{90}{\scriptsize\textbf{#1}}
    \end{minipage}%
    \begin{minipage}[c]{\scatterwidth}
        \centering
        \includegraphics[width=\linewidth]{#2}
    \end{minipage}%
    \hfill
    \begin{minipage}[c]{\scatterwidth}
        \centering
        \includegraphics[width=\linewidth]{#3}
    \end{minipage}%
    \hfill
    \begin{minipage}[c]{\scatterwidth}
        \centering
        \includegraphics[width=\linewidth]{#4}
    \end{minipage}

    \vspace{-2mm}
}

\begin{figure*}
    \centering

    % ========================================================
    % LEFT HALF
    % ========================================================
    \begin{minipage}{0.495\textwidth}
        \centering

        \scatterrow
            {GSM8K}
            {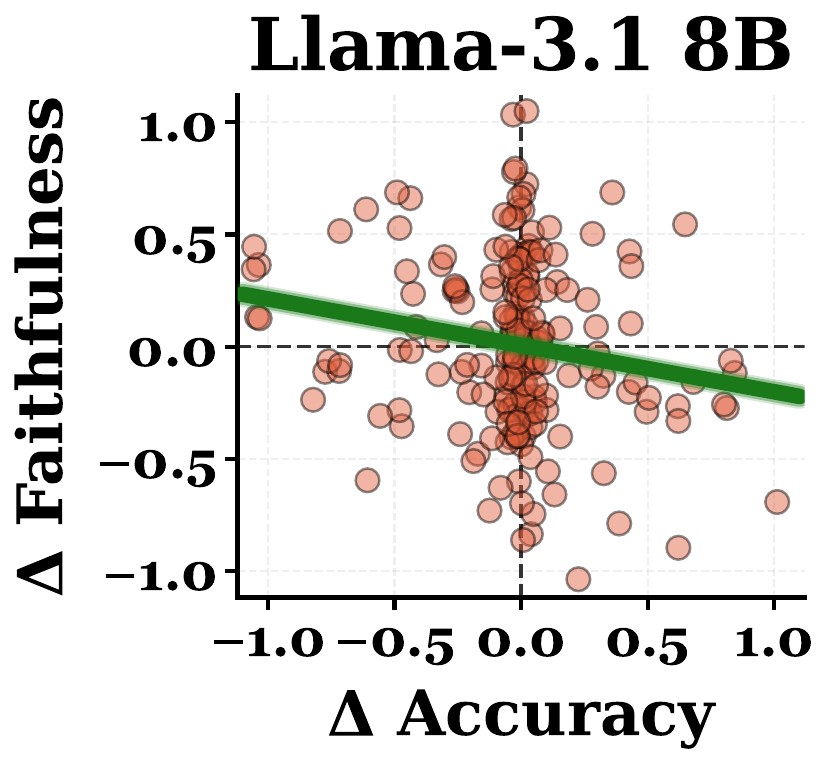}
            {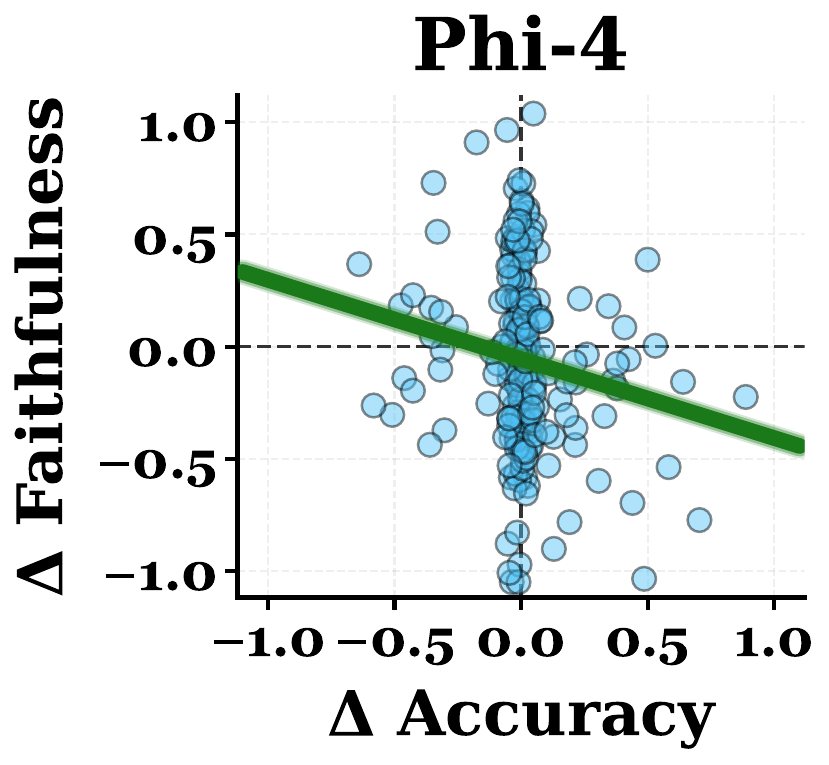}
            {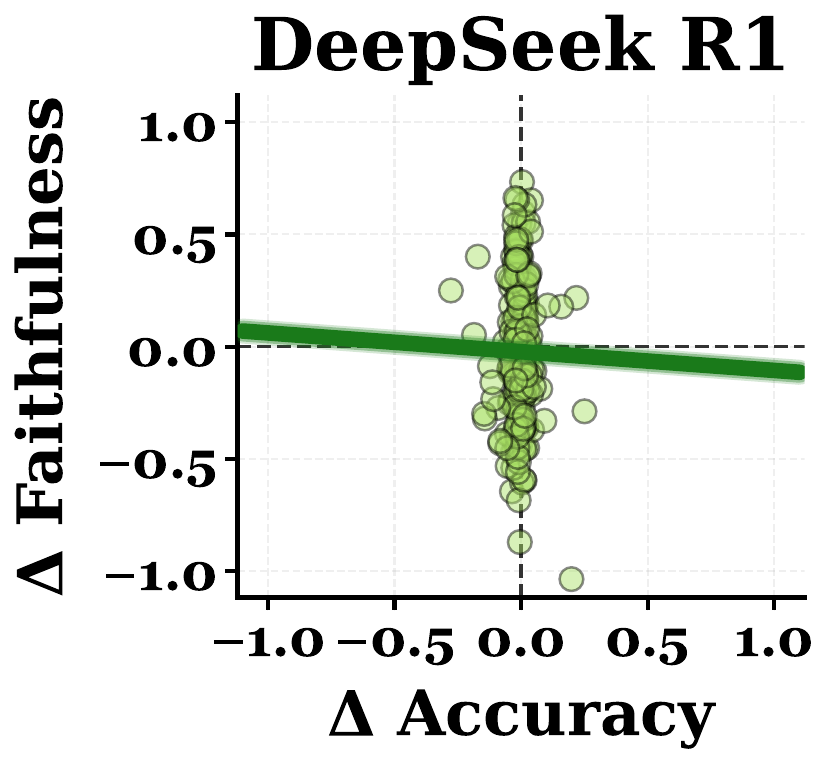}

        \scatterrow
            {BoolQ}
            {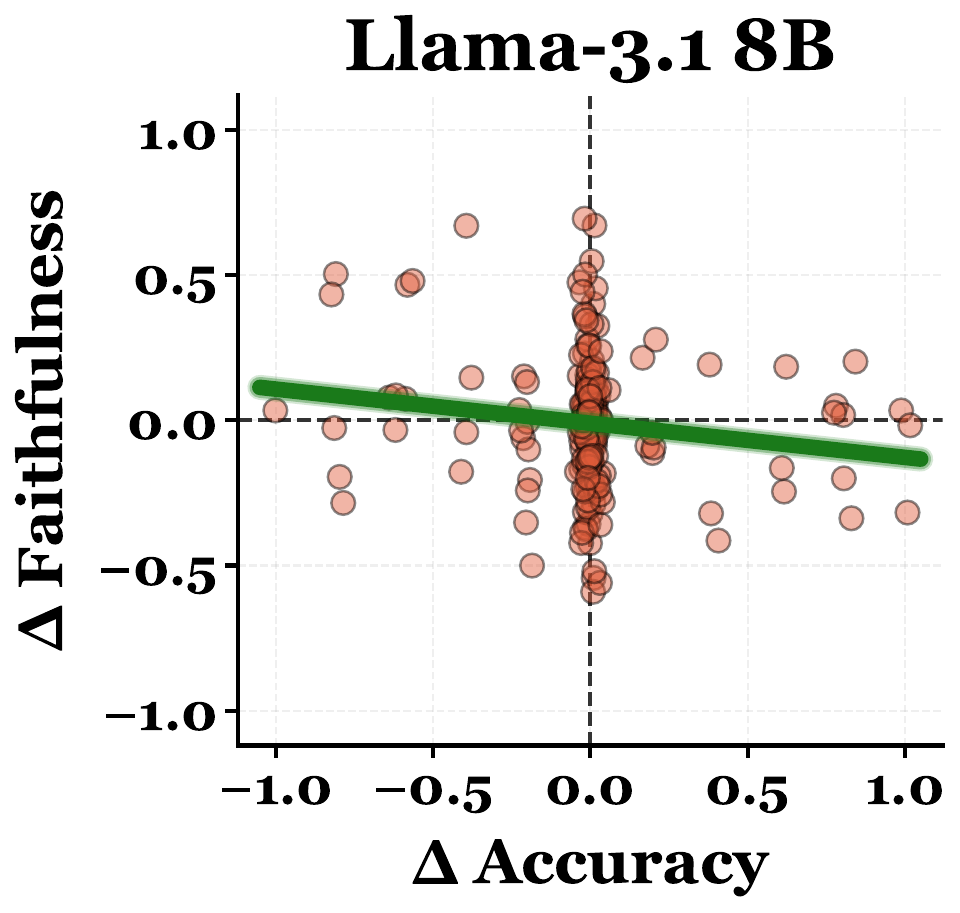}
            {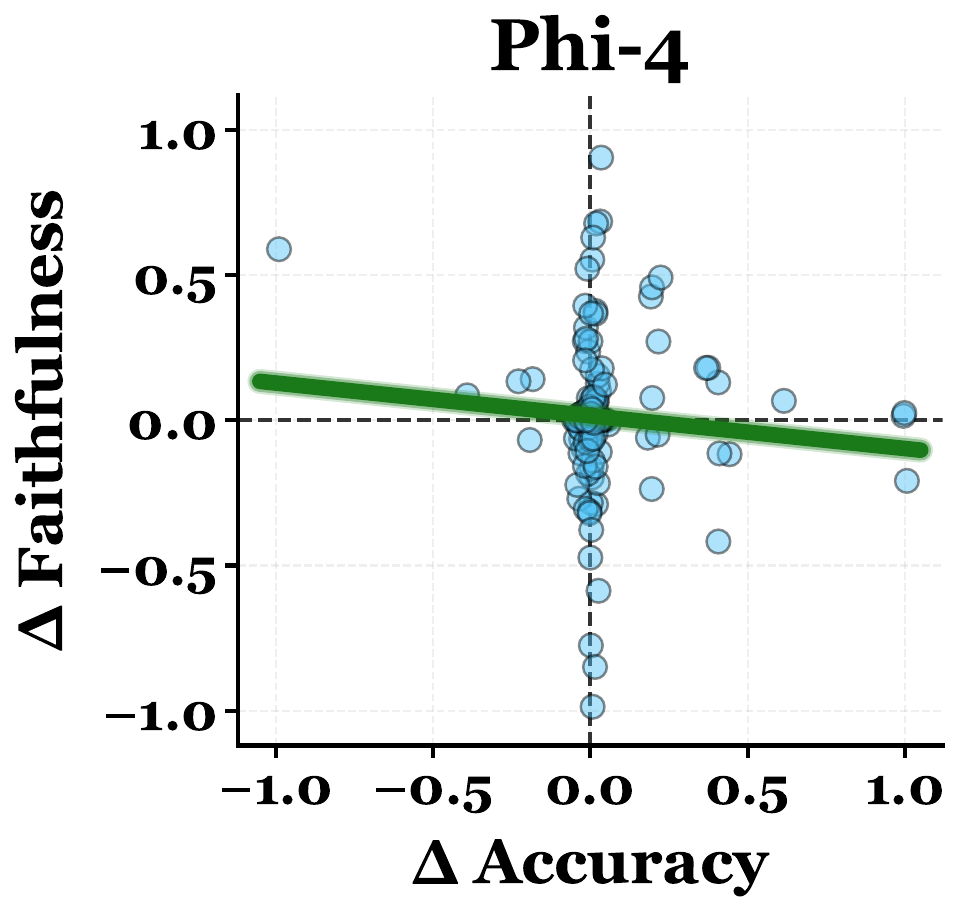}
            {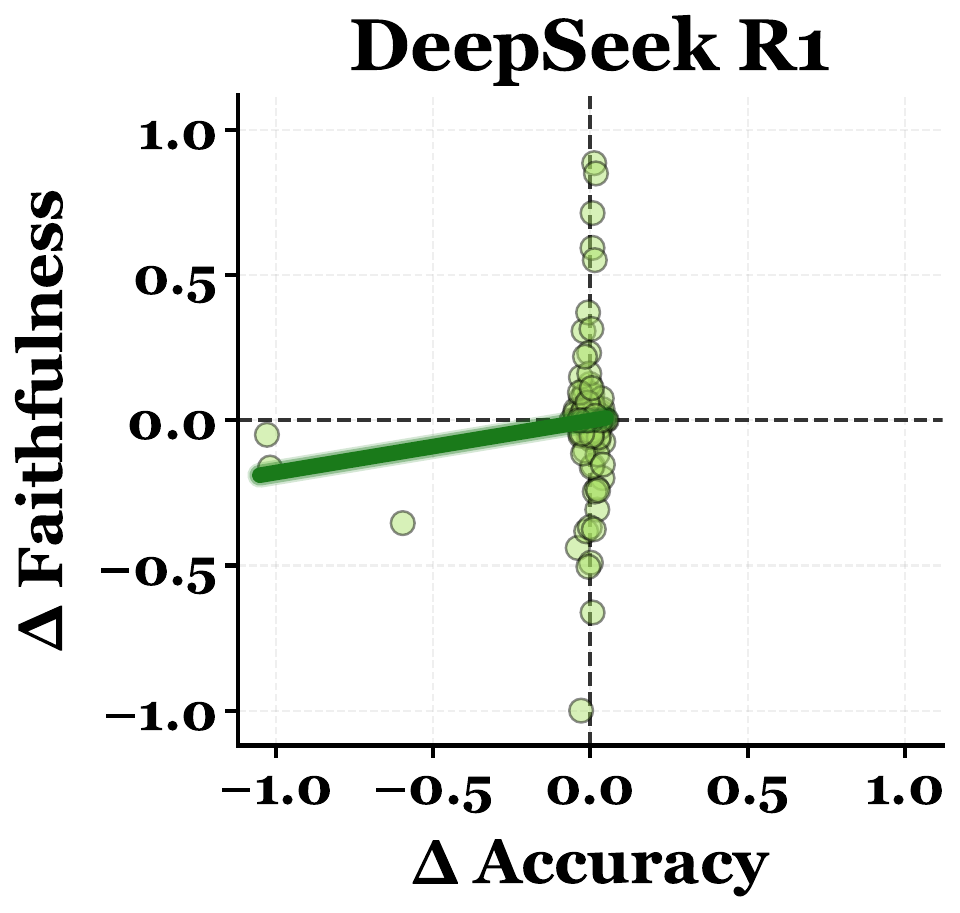}

        \scatterrow
            {StratQA}
            {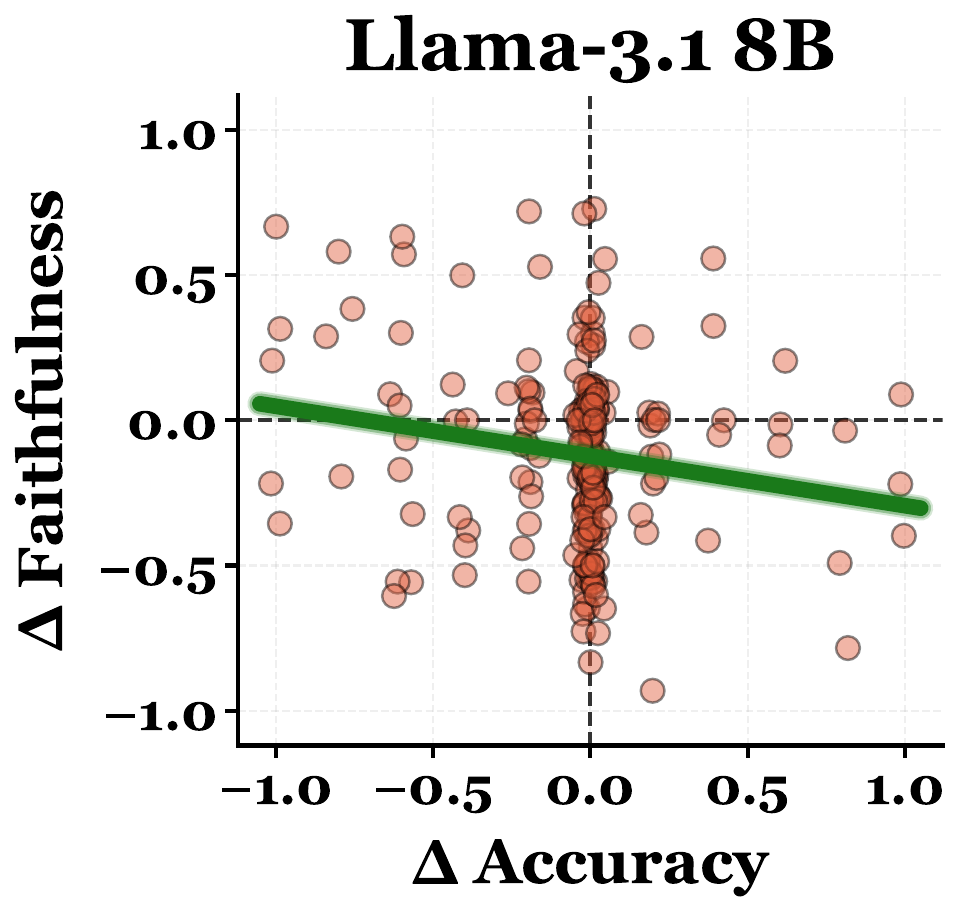}
            {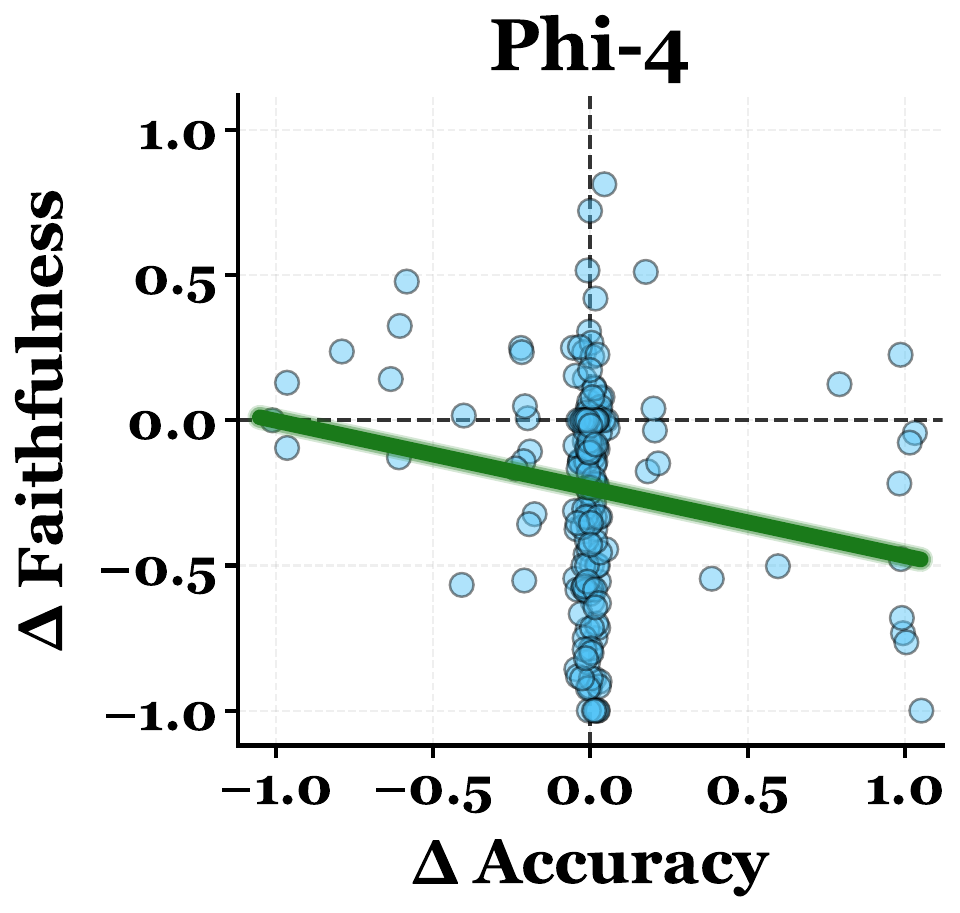}
            {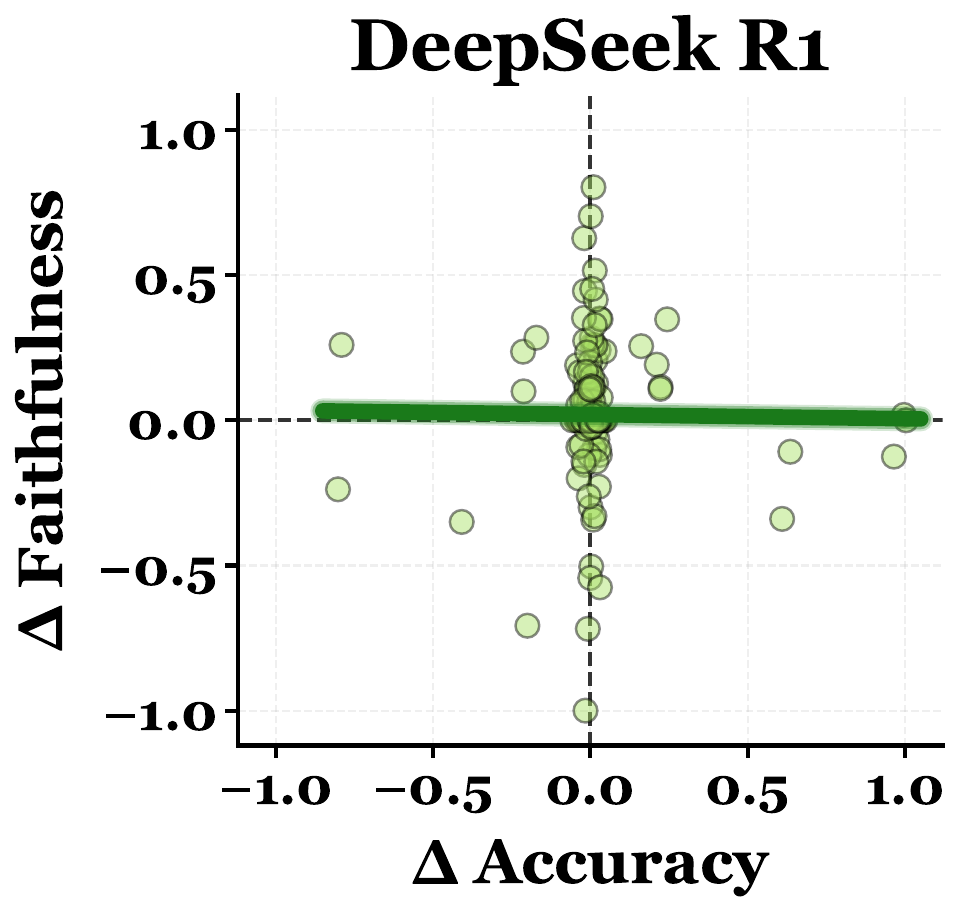}

        \vspace{-1mm}

    \end{minipage}%
    \hfill
    % ========================================================
    % RIGHT HALF
    % ========================================================
    \begin{minipage}{0.495\textwidth}
        \centering

        \scatterrow
            {MedQA}
            {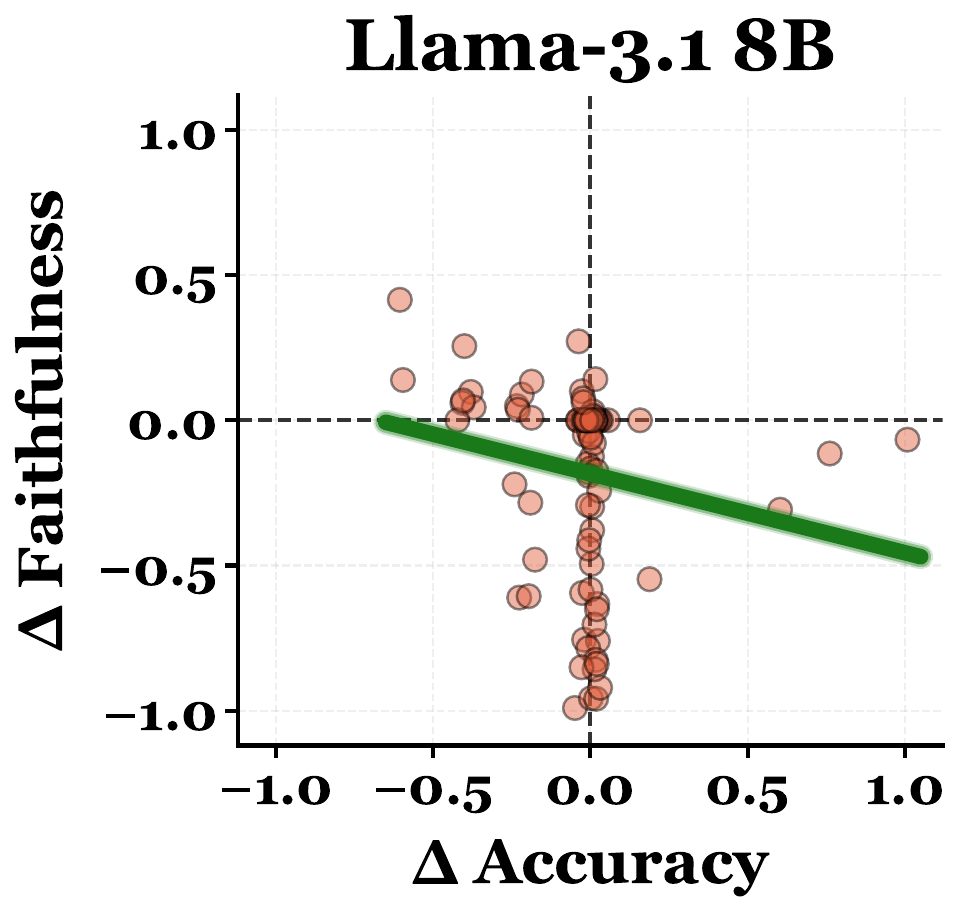}
            {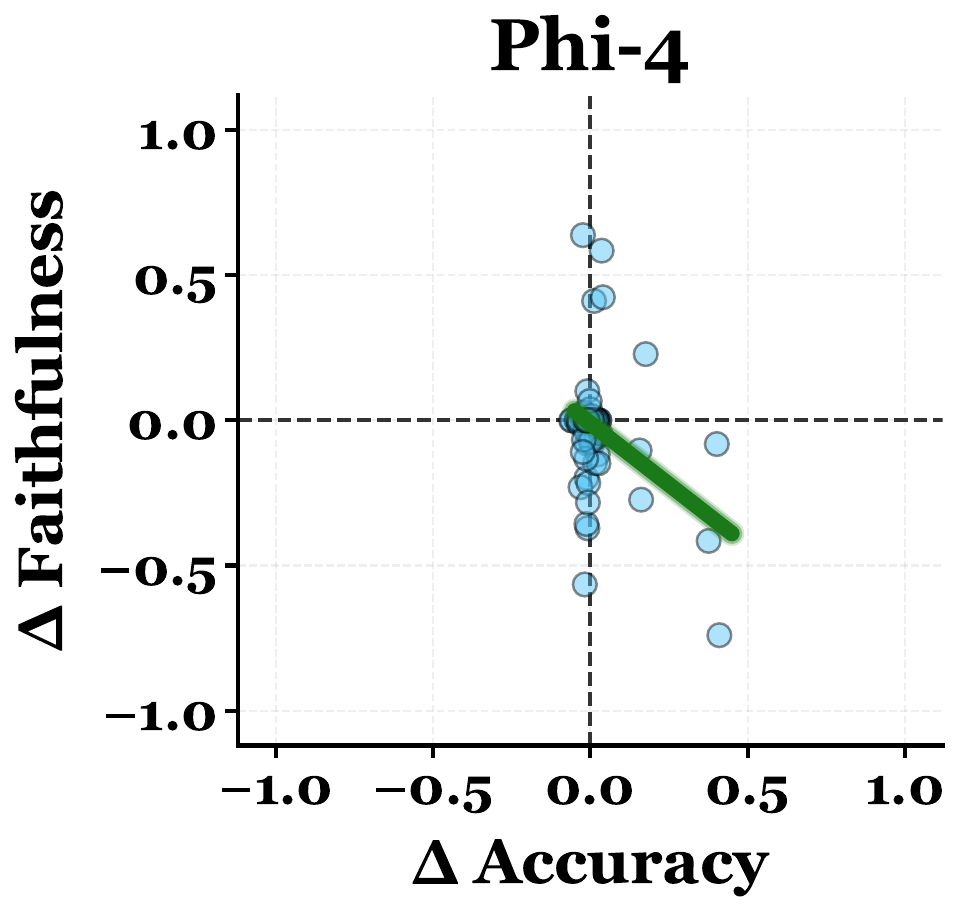}
            {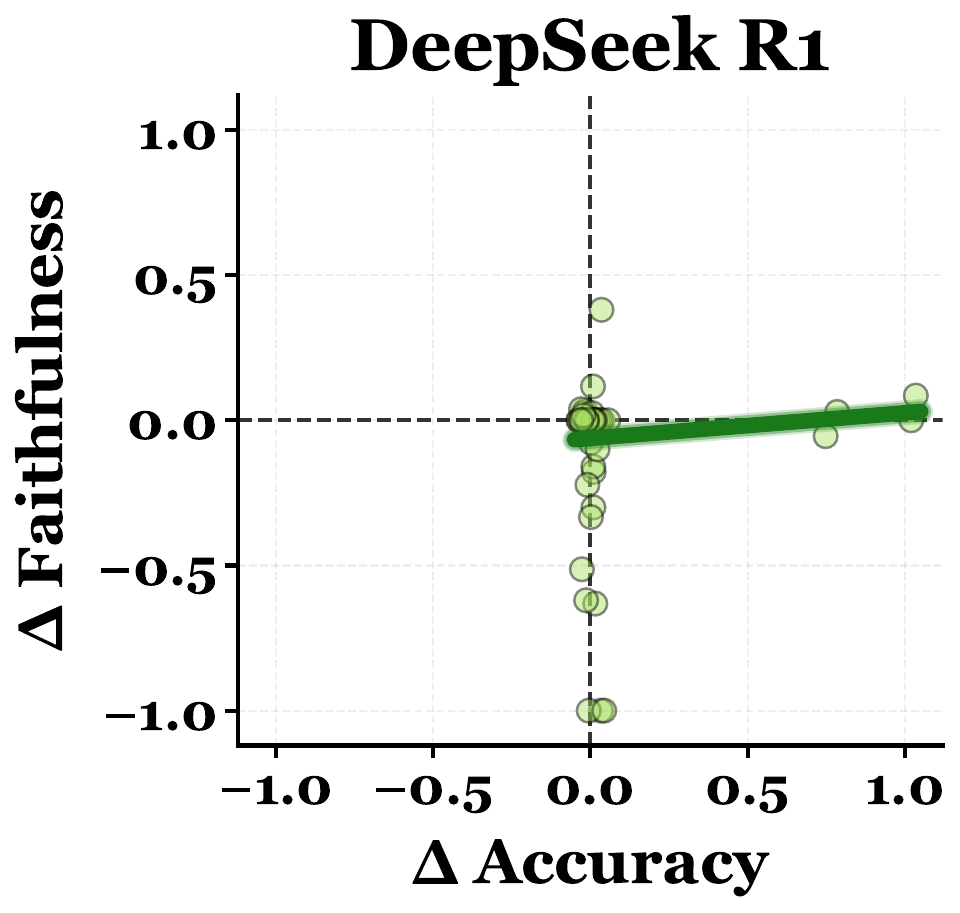}

        \scatterrow
            {GPQA}
            {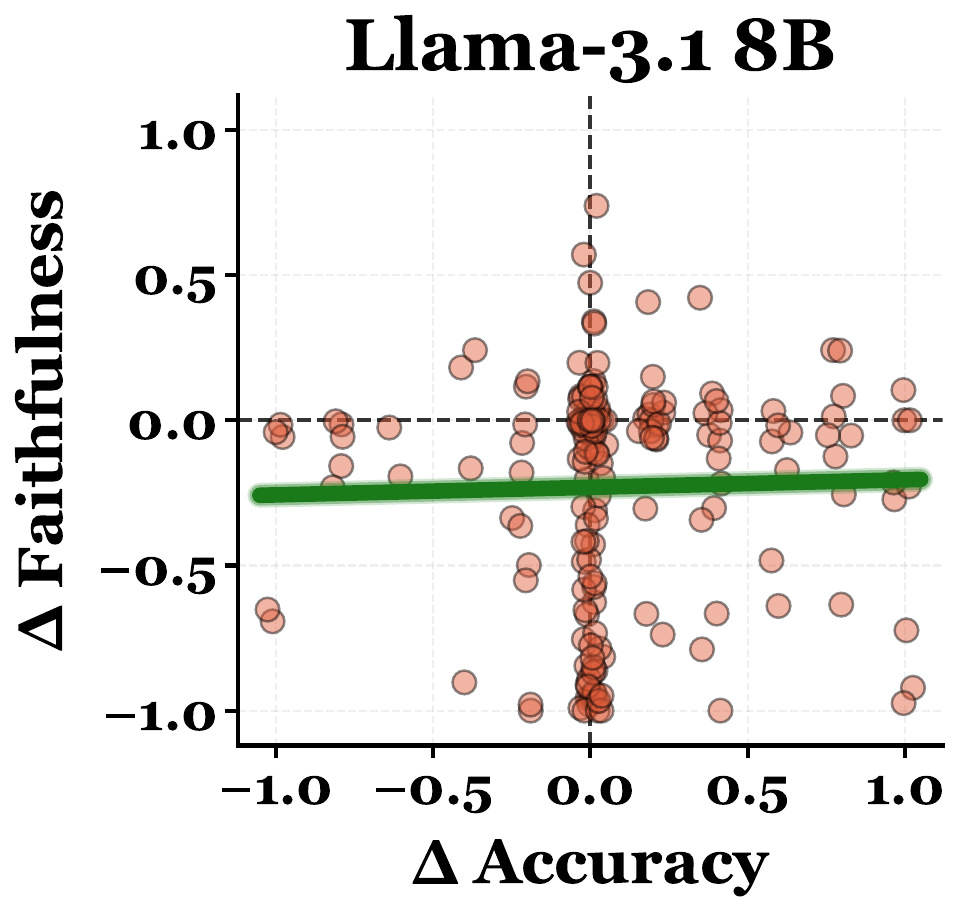}
            {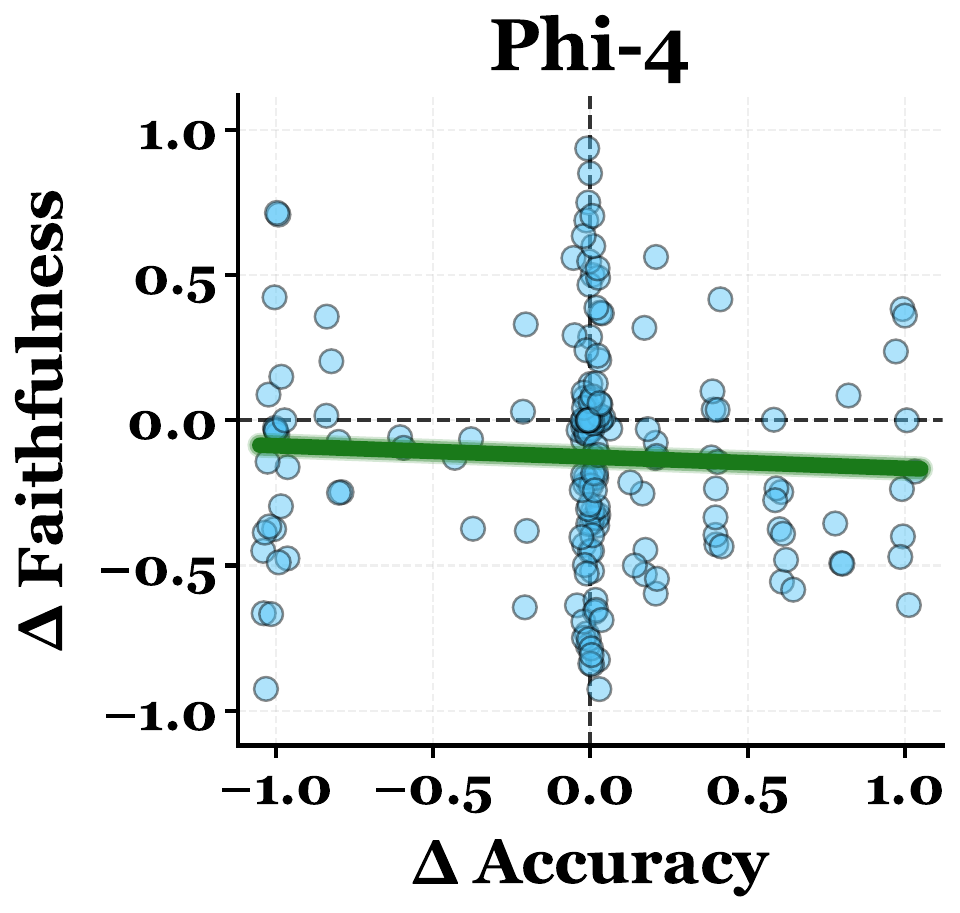}
            {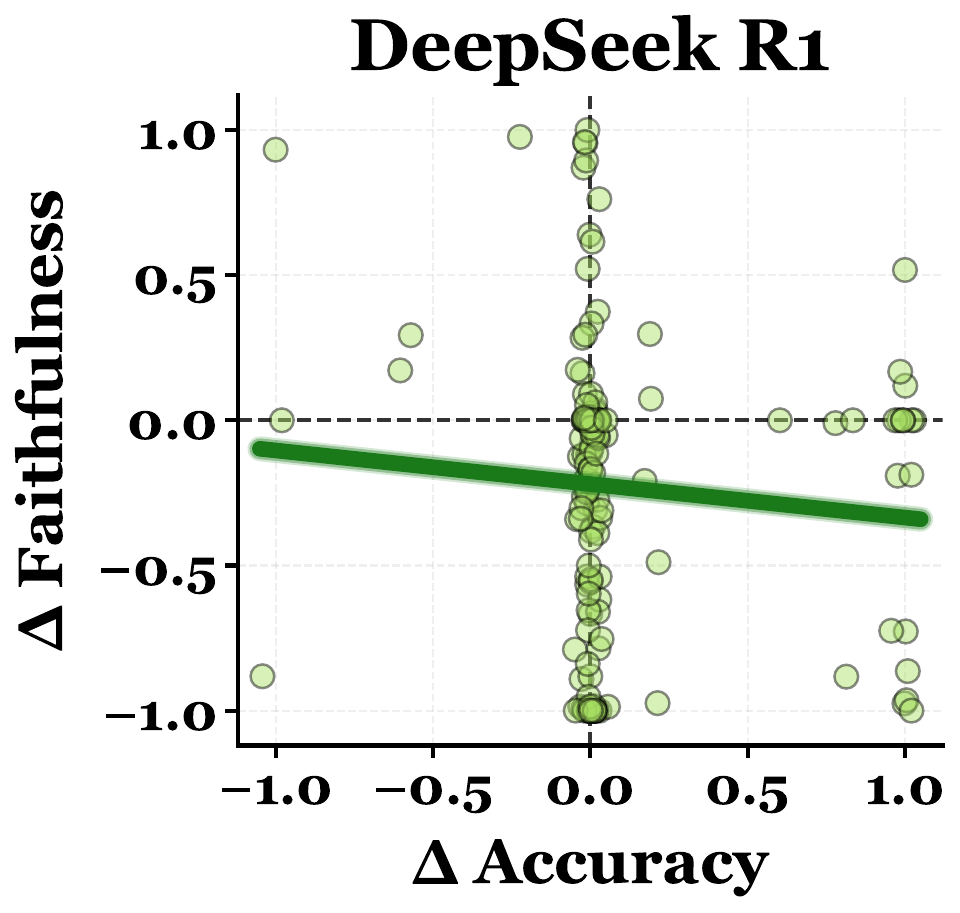}

        \scatterrow
            {AIME '26}
            {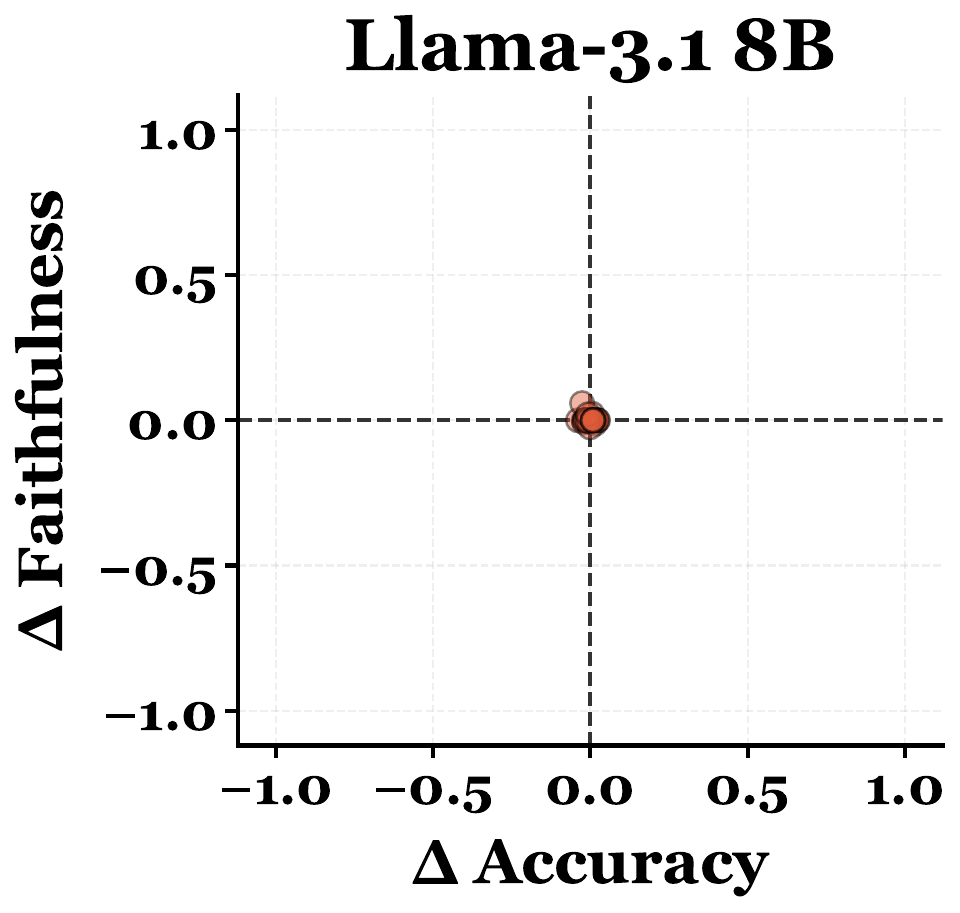}
            {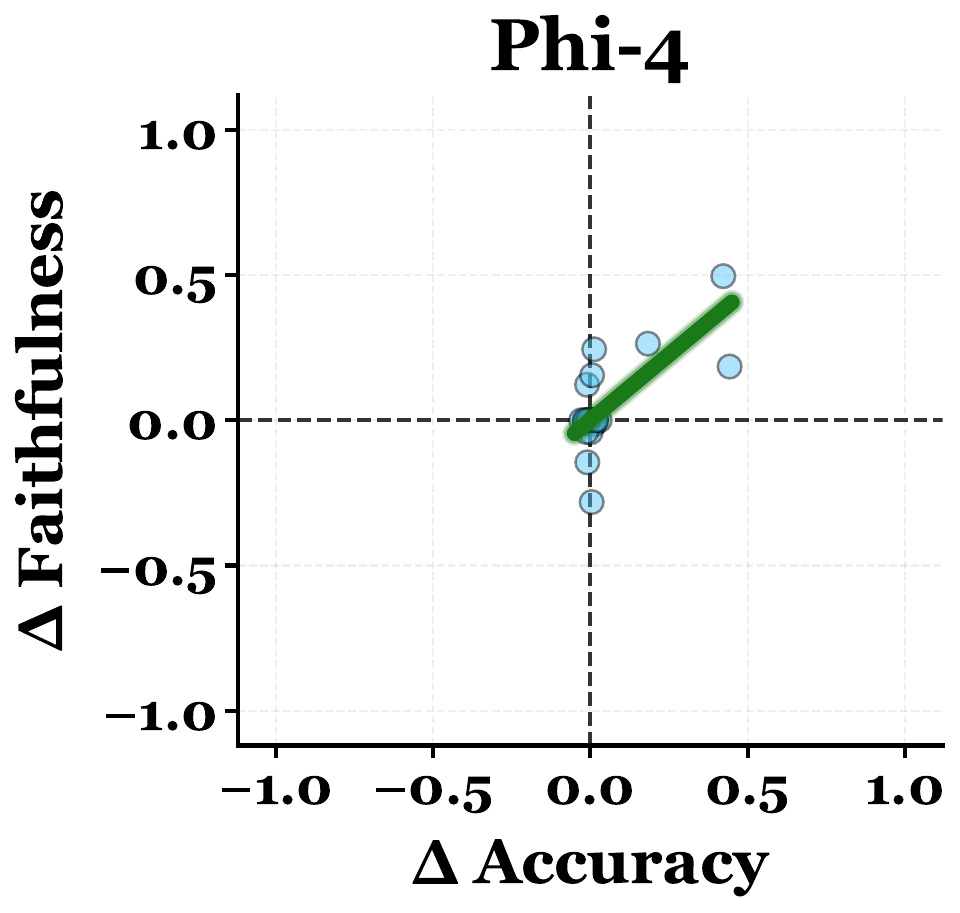}
            {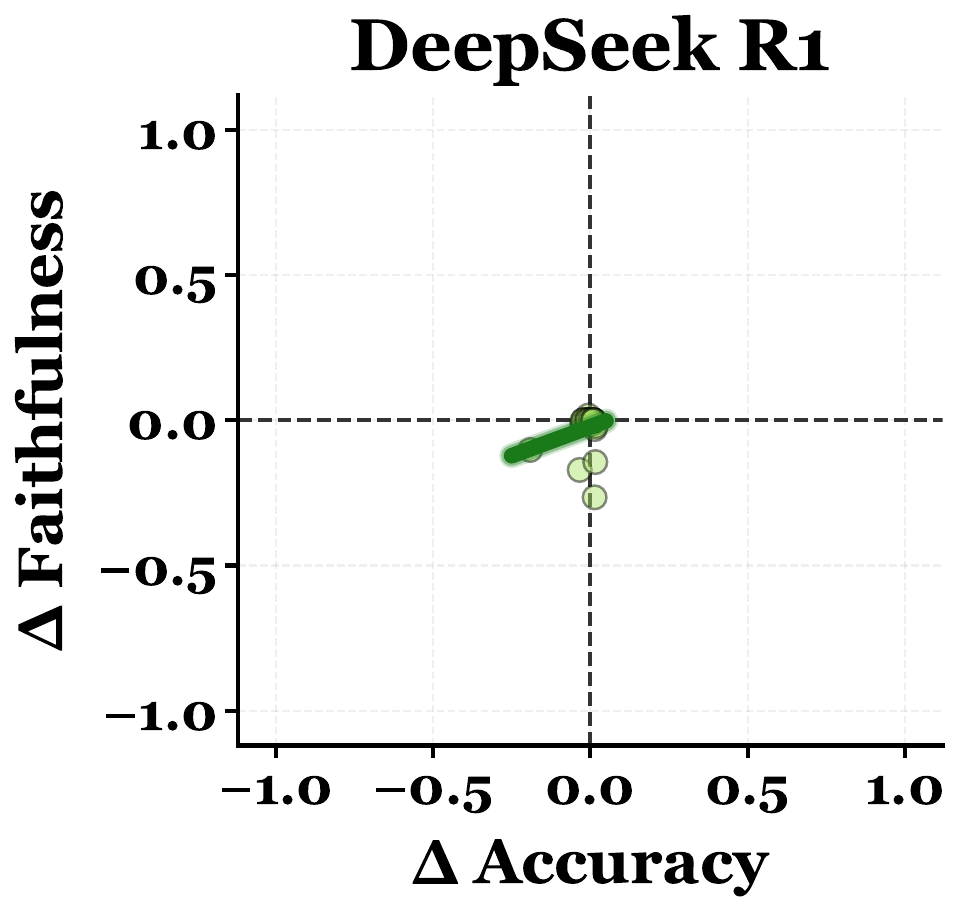}

        %\vspace{-1mm}

    \end{minipage}

    \caption{
        \textbf{GSC medoid ablation}
        Each point represents a GLLM's answer and reports the change in accuracy and reasoning faithfulness
        following the adversarial intervention which ablated the GSC selected medoid. The green line denotes the
        linear regression trend. The number of points reflects our experimental setup (\Cref{sec:exp_setup}). For AIME '26, we plot less points because the dataset is composed of just 30 questions. Therefore, AIME '26 regression trends are not statistically significant ($p > 0.05$).
    }
    \label{fig:scatter_all_datasets}
\end{figure*}

%\input{pic/broad_attack}
%\rreva{PUT DLLM ABLATION EXP FOR SURE}
%\rreva{PUT OTHER RESULTS FROM APPENDIX HERE, IN MAIN TEXT, BECAUSE WE HAVE SPACE + NEW SECTION ON COMPUTATIONAL COST!!! RESOURCES UTILIZATION, VS TOPO-UQ, SC ETC}
%\begin{figure}[h]

%\setlength{\intextsep}{0pt} % Default is usually larger
%\begin{wrapfigure}[18]{r}{0.3\textwidth}
%    \centering
%    \includegraphics[width=\linewidth]
    %\includegraphics[scale=0.18]
%    {pic/accuracy_combined_heatmap.pdf} 
%    \caption{\textbf{Testing for Real, Faithful Accuracy}. 
    %\textit{Methodology}: We compare the nominal accuracy of standard SC against our GSC framework using the same datasets and models of \Cref{sec_rq1_all}. 
%    The rightmost panel displays the %absolute 
%    performance difference ($\Delta$ GSC $-$ SC). 
    %\textit{Results}: 
%    Smaller models (\eg, LLaMA 3.1 8B) show accuracy drops %(up to $-9.0$ pp on MedQA), 
%    revealing reliance on SC-rewarded lucky guesses. As reasoning capability increases (Phi-4, DeepSeek R1), the gap narrows and turns positive with DeepSeek R1 improving in all settings.} %GSC thus filters hallucinations while preserving or improving strong-model performance.}
    \label{fig:exp3_fig}
    %\vspace{-5mm}
%\end{figure}
%\end{wrapfigure}

\begin{figure}
    \centering
    \begin{minipage}[c]{0.56\linewidth}
        \centering
        \includegraphics[width=\linewidth]{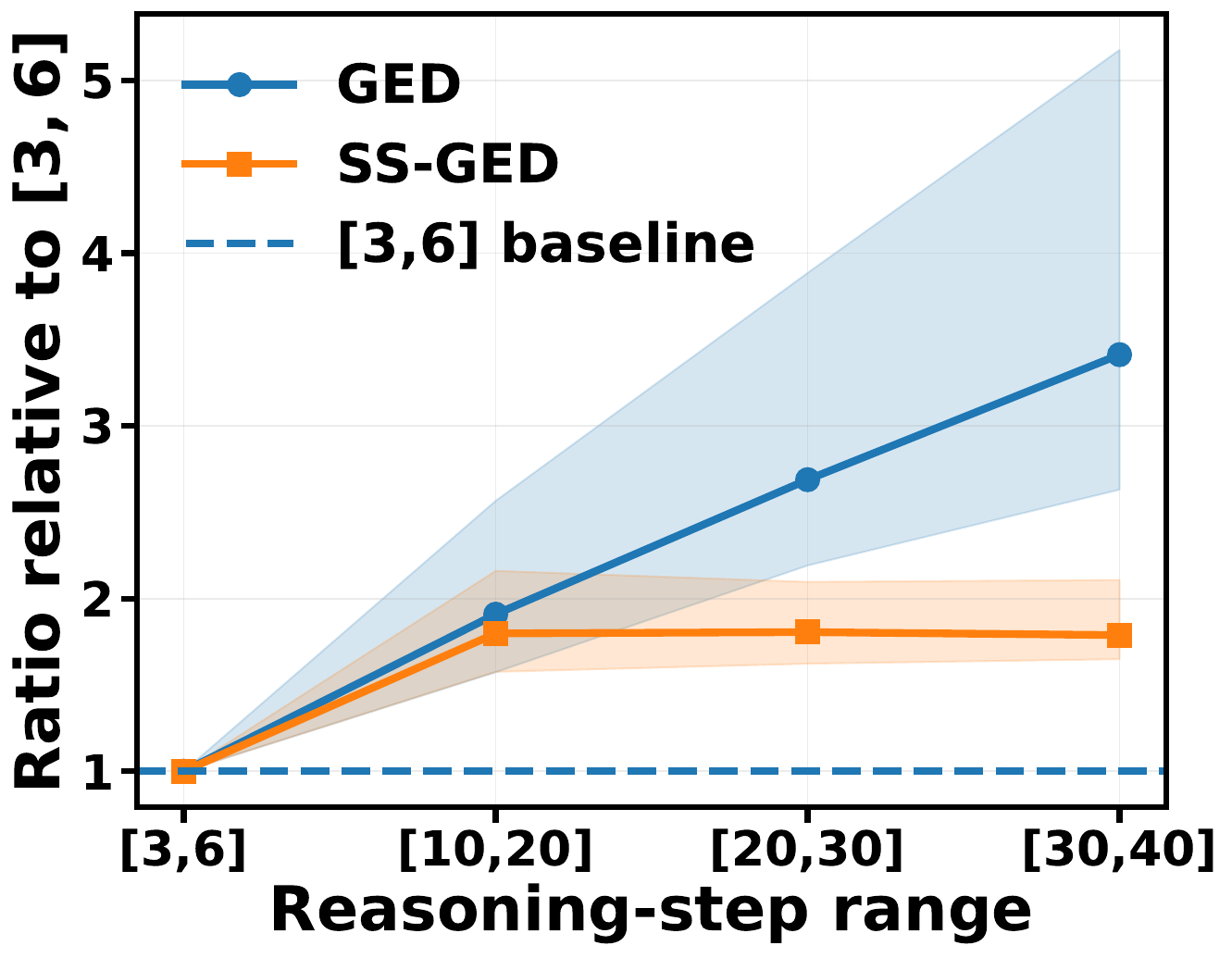}
    \end{minipage}
    \hfill
    \begin{minipage}[c]{0.40\linewidth}
        \caption{\textbf{GED and SS-GED scaling with resoning steps.}
        Values are normalized to the $[3,6]$ baseline (dashed). Light bands show the distributions, dark lines show the geometric averages across dataset--GLLM configurations}
        \label{fig:ged_ss_ged_scalabaility}
    \end{minipage}
\end{figure}

\subsection{Scalability to Longer Reasoning Paths}\label{sec:scalability_resources}
% Shorthand for aligned numerical columns
\newcolumntype{N}[1]{%
    S[
        table-format=#1,
        table-number-alignment=center,
        detect-weight=true,
        detect-inline-weight=math
    ]%
}

\definecolor{grcsrow}{gray}{0.93}

\definecolor{stepBase}{RGB}{248,250,253}
\definecolor{step10}{RGB}{232,241,250}
\definecolor{step20}{RGB}{209,225,243}
\definecolor{step30}{RGB}{184,207,233}

\newcommand{\basecell}{\cellcolor{stepBase}}
\newcommand{\stepXcell}{\cellcolor{step10}}
\newcommand{\stepXXcell}{\cellcolor{step20}}
\newcommand{\stepXXXcell}{\cellcolor{step30}}

\begin{table*}
    \centering
    \scriptsize
    \setlength{\tabcolsep}{3pt}
    \renewcommand{\arraystretch}{0.95}

    \caption{
        \textbf{Resource scaling of $\mathcal{U}_{\mathrm{GRCS}}$ with reasoning steps.}
        %GPU runtime is reported per individual CoT and per set of
        %20 CoTs. Energy consumption, peak GPU memory, and peak GPU
        %utilization are reported per set of 20 CoTs.
        The table reports the same measures of \Cref{tab:computational_cost}.
        The baseline $[3,6]$ values are reported from
        \Cref{tab:computational_cost}. %For longer traces,
        %each metric is first averaged across questions within each
        %independent run, and the reported mean $\pm$ standard
        %deviation is then computed across all available model--run
        %observations.
        Measurements %are
        collected on the same GPU over complete datasets and $5$
        independent runs, with results reported as mean $\pm$ standard
        deviation.
        Lower values indicate lower resource cost.
    }
    \label{tab:scalability_resources}

    \begin{tabular}{
        C{1.25cm}
        C{1.05cm}
        N{1.4} @{\,${}\pm{}$\,} N{1.4}
        N{2.2} @{\,${}\pm{}$\,} N{2.2}
        N{4.2} @{\,${}\pm{}$\,} N{4.2}
        N{4.2} @{\,${}\pm{}$\,} N{4.2}
        N{3.2} @{\,${}\pm{}$\,} N{3.2}
    }
        \toprule
        \textbf{Dataset}
        & \textbf{Steps}
        & \multicolumn{2}{c}{
            \shortstack{\textbf{GPU Runtime (s)}\\\textbf{per 1 CoT}}
        }
        & \multicolumn{2}{c}{
            \shortstack{\textbf{GPU Runtime (s)}\\\textbf{per 20 CoTs}}
        }
        & \multicolumn{2}{c}{
            \shortstack{\textbf{Energy (J)}\\\textbf{per 20 CoTs}}
        }
        & \multicolumn{2}{c}{
            \shortstack{\textbf{Peak GPU Mem. (MB)}\\\textbf{per 20 CoTs}}
        }
        & \multicolumn{2}{c}{
            \shortstack{\textbf{Peak GPU Usage (\%)}\\\textbf{per 20 CoTs}}
        } \\

        \cmidrule(lr){3-4}
        \cmidrule(lr){5-6}
        \cmidrule(lr){7-8}
        \cmidrule(lr){9-10}
        \cmidrule(lr){11-12}

        \multirow{4}{*}{GSM8K}
        & \basecell $\mathbf{[3,6]}$
        & \basecell 0.0381 & \basecell 0.0034
        & \basecell 0.75 & \basecell 0.07
        & \basecell 72.04 & \basecell 4.86
        & \basecell 1252.79 & \basecell 0.00
        & \basecell 8.25 & \basecell 2.75 \\

        & \stepXcell $\mathbf{[10,20]}$
        & \stepXcell 0.0296 & \stepXcell 0.0001
        & \stepXcell 0.59 & \stepXcell 0.00
        & \stepXcell 79.67 & \stepXcell 2.91
        & \stepXcell 1296.40 & \stepXcell 0.00
        & \stepXcell 38.16 & \stepXcell 0.77 \\

        & \stepXXcell $\mathbf{[20,30]}$
        & \stepXXcell 0.0344 & \stepXXcell 0.0001
        & \stepXXcell 0.69 & \stepXXcell 0.00
        & \stepXXcell 117.04 & \stepXXcell 1.56
        & \stepXXcell 1299.92 & \stepXXcell 0.00
        & \stepXXcell 38.23 & \stepXXcell 3.94 \\

        & \stepXXXcell $\mathbf{[30,40]}$
        & \stepXXXcell 0.0404 & \stepXXXcell 0.0001
        & \stepXXXcell 0.81 & \stepXXXcell 0.00
        & \stepXXXcell 135.80 & \stepXXXcell 1.72
        & \stepXXXcell 1318.46 & \stepXXXcell 0.00
        & \stepXXXcell 32.84 & \stepXXXcell 4.05 \\

        \midrule

        \multirow{4}{*}{BoolQ}
        & \basecell $\mathbf{[3,6]}$
        & \basecell 0.0300 & \basecell 0.0033
        & \basecell 0.60 & \basecell 0.07
        & \basecell 60.28 & \basecell 2.32
        & \basecell 1273.67 & \basecell 0.00
        & \basecell 11.19 & \basecell 5.53 \\

        & \stepXcell $\mathbf{[10,20]}$
        & \stepXcell 0.0360 & \stepXcell 0.0001
        & \stepXcell 0.72 & \stepXcell 0.00
        & \stepXcell 117.78 & \stepXcell 3.72
        & \stepXcell 1373.51 & \stepXcell 0.00
        & \stepXcell 37.45 & \stepXcell 9.08 \\

        & \stepXXcell $\mathbf{[20,30]}$
        & \stepXXcell 0.0454 & \stepXXcell 0.0001
        & \stepXXcell 0.91 & \stepXXcell 0.00
        & \stepXXcell 161.55 & \stepXXcell 1.58
        & \stepXXcell 1489.32 & \stepXXcell 0.00
        & \stepXXcell 34.68 & \stepXXcell 3.52 \\

        & \stepXXXcell $\mathbf{[30,40]}$
        & \stepXXXcell 0.0569 & \stepXXXcell 0.0002
        & \stepXXXcell 1.14 & \stepXXXcell 0.00
        & \stepXXXcell 204.53 & \stepXXXcell 2.99
        & \stepXXXcell 1492.01 & \stepXXXcell 0.00
        & \stepXXXcell 29.52 & \stepXXXcell 4.47 \\

        \midrule

        \multirow{4}{*}{StratQA}
        & \basecell $\mathbf{[3,6]}$
        & \basecell 0.0315 & \basecell 0.0006
        & \basecell 0.63 & \basecell 0.01
        & \basecell 62.71 & \basecell 5.69
        & \basecell 1286.92 & \basecell 0.00
        & \basecell 11.19 & \basecell 5.53 \\

        & \stepXcell $\mathbf{[10,20]}$
        & \stepXcell 0.0334 & \stepXcell 0.0001
        & \stepXcell 0.67 & \stepXcell 0.00
        & \stepXcell 97.89 & \stepXcell 3.75
        & \stepXcell 1275.78 & \stepXcell 0.00
        & \stepXcell 40.26 & \stepXcell 2.32 \\

        & \stepXXcell $\mathbf{[20,30]}$
        & \stepXXcell 0.0412 & \stepXXcell 0.0001
        & \stepXXcell 0.82 & \stepXXcell 0.00
        & \stepXXcell 140.28 & \stepXXcell 2.45
        & \stepXXcell 1307.53 & \stepXXcell 0.00
        & \stepXXcell 36.51 & \stepXXcell 2.02 \\

        & \stepXXXcell $\mathbf{[30,40]}$
        & \stepXXXcell 0.0536 & \stepXXXcell 0.0001
        & \stepXXXcell 1.07 & \stepXXXcell 0.00
        & \stepXXXcell 185.85 & \stepXXXcell 2.79
        & \stepXXXcell 1333.15 & \stepXXXcell 0.00
        & \stepXXXcell 25.55 & \stepXXXcell 7.34 \\

        \midrule

        \multirow{4}{*}{MedQA}
        & \basecell $\mathbf{[3,6]}$
        & \basecell 0.0349 & \basecell 0.0018
        & \basecell 0.70 & \basecell 0.04
        & \basecell 64.79 & \basecell 3.23
        & \basecell 1281.63 & \basecell 0.00
        & \basecell 5.17 & \basecell 1.23 \\

        & \stepXcell $\mathbf{[10,20]}$
        & \stepXcell 0.0417 & \stepXcell 0.0001
        & \stepXcell 0.83 & \stepXcell 0.00
        & \stepXcell 141.30 & \stepXcell 4.52
        & \stepXcell 2169.67 & \stepXcell 0.00
        & \stepXcell 35.85 & \stepXcell 3.22 \\

        & \stepXXcell $\mathbf{[20,30]}$
        & \stepXXcell 0.0506 & \stepXXcell 0.0002
        & \stepXXcell 1.01 & \stepXXcell 0.00
        & \stepXXcell 181.18 & \stepXXcell 1.99
        & \stepXXcell 2205.36 & \stepXXcell 0.00
        & \stepXXcell 27.30 & \stepXXcell 7.28 \\

        & \stepXXXcell $\mathbf{[30,40]}$
        & \stepXXXcell 0.0568 & \stepXXXcell 0.0002
        & \stepXXXcell 1.14 & \stepXXXcell 0.00
        & \stepXXXcell 206.95 & \stepXXXcell 2.50
        & \stepXXXcell 2211.55 & \stepXXXcell 0.00
        & \stepXXXcell 27.13 & \stepXXXcell 6.08 \\

        \midrule

        \multirow{4}{*}{GPQA}
        & \basecell $\mathbf{[3,6]}$
        & \basecell 0.0378 & \basecell 0.0007
        & \basecell 0.74 & \basecell 0.01
        & \basecell 72.38 & \basecell 1.49
        & \basecell 1343.50 & \basecell 0.00
        & \basecell 8.23 & \basecell 4.21 \\

        & \stepXcell $\mathbf{[10,20]}$
        & \stepXcell 0.0497 & \stepXcell 0.0008
        & \stepXcell 0.99 & \stepXcell 0.02
        & \stepXcell 184.03 & \stepXcell 5.42
        & \stepXcell 1460.00 & \stepXcell 0.02
        & \stepXcell 43.31 & \stepXcell 9.39 \\

        & \stepXXcell $\mathbf{[20,30]}$
        & \stepXXcell 0.0571 & \stepXXcell 0.0009
        & \stepXXcell 1.14 & \stepXXcell 0.02
        & \stepXXcell 219.85 & \stepXXcell 3.32
        & \stepXXcell 1480.13 & \stepXXcell 0.02
        & \stepXXcell 38.13 & \stepXXcell 8.69 \\

        & \stepXXXcell $\mathbf{[30,40]}$
        & \stepXXXcell 0.0719 & \stepXXXcell 0.0019
        & \stepXXXcell 1.44 & \stepXXXcell 0.04
        & \stepXXXcell 282.75 & \stepXXXcell 2.57
        & \stepXXXcell 1557.66 & \stepXXXcell 0.00
        & \stepXXXcell 30.50 & \stepXXXcell 6.20 \\

        \midrule

        \multirow{4}{*}{AIME'26}
        & \basecell $\mathbf{[3,6]}$
        & \basecell 0.0674 & \basecell 0.0004
        & \basecell 1.35 & \basecell 0.01
        & \basecell 34.60 & \basecell 0.24
        & \basecell 1405.09 & \basecell 0.00
        & \basecell 17.23 & \basecell 2.50 \\

        & \stepXcell $\mathbf{[10,20]}$
        & \stepXcell 0.0672 & \stepXcell 0.0006
        & \stepXcell 1.34 & \stepXcell 0.01
        & \stepXcell 293.84 & \stepXcell 7.07
        & \stepXcell 2334.28 & \stepXcell 0.01
        & \stepXcell 46.79 & \stepXcell 7.91 \\

        & \stepXXcell $\mathbf{[20,30]}$
        & \stepXXcell 0.0827 & \stepXXcell 0.0002
        & \stepXXcell 1.65 & \stepXXcell 0.00
        & \stepXXcell 367.48 & \stepXXcell 2.43
        & \stepXXcell 2687.79 & \stepXXcell 0.00
        & \stepXXcell 42.75 & \stepXXcell 5.22 \\

        & \stepXXXcell $\mathbf{[30,40]}$
        & \stepXXXcell 0.0953 & \stepXXXcell 0.0004
        & \stepXXXcell 1.91 & \stepXXXcell 0.01
        & \stepXXXcell 412.71 & \stepXXXcell 2.88
        & \stepXXXcell 2289.75 & \stepXXXcell 0.01
        & \stepXXXcell 33.55 & \stepXXXcell 7.36 \\

        \bottomrule
    \end{tabular}
\end{table*}
\emph{To test $\pname$ scalability to longer reasoning paths}, we increase the $[n_{l}, n_{h}]$ declarative steps required for each CoT from $[3,6]$ ($\cf$, \Cref{sec:exp_setup}) to $[10,20]$, $[20,30]$, and $[30,40]$ steps.
These settings increase the geometric-mean size of reasoning graphs across datasets and GLLMs 
from
$14.43$ nodes at $[3,6]$ to $26.19$, $36.65$, and $44.65$ nodes, corresponding to increases of 
$1.82\times$, $2.54\times$, and $3.09\times$, %increases, 
respectively.
\Cref{fig:ged_ss_ged_scalabaility} compares GED with %our 
SS-GED metric, %used throughout $\pname$'s pipeline, with 
both normalized to the $[3,6]$-step baseline and
geometrically averaged across dataset--GLLM configurations.
GED grows steadily with graph
size, and reaches on average $1.91\times$, $2.69\times$, and $3.41\times$ of its baseline
value. In contrast, %our 
SS-GED reaches on average $1.80\times$ at
$[10,20]$ steps and remains \textbf{stable} as reasoning steps further increase to $[20,30]$ and $[30,40]$. % steps.

\Cref{tab:scalability_resources} reports the resource efficiency of $\mathcal{U}_{\mathrm{GRCS}}$ for $[3,6]$ (baseline configuration, % exactly the same results %shown 
the exact same results
reported in \Cref{tab:computational_cost}), $[10,20]$, $[20,30]$, and $[30,40]$ declarative steps per CoT. Relative to $[3,6]$ declarative steps, the average GPU runtime per 20 CoTs at $[30, 40]$ declarative steps increases by only $1.4\times$ for AIME'26, $1.95\times$ for GPQA and $1.62\times$ for MedQA, the %hardest 
most challenging datasets. Energy consumption and peak GPU usage also increase because longer CoTs produce more graph nodes, thus graph construction must encode and connect more facts, and pairwise SS-GED computations operate on larger nodes/edges matrices. Nevertheless, even at $[30,40]$ steps, our pipeline consumes less energy and
reaches lower peak GPU utilization than Topo-UQ and CoTA evaluated at
$[3,6]$ steps on every dataset (compare 
Energy and GPU usage results %from 
in~\Cref{tab:scalability_resources} with Energy and GPU usage results for Topo-UQ, CoTA from \Cref{tab:computational_cost}).
\bem{These results show that SS-GED remains stable, %as reasoning %graphs
%steps grow, 
and $\pname$ remains computationally tractable %for CoTs containing
%when reasoning steps in CoTs grow
%up to $[30,40]$.} % declarative steps.} 
on longer reasoning tasks.}
\subsection{Stability of $\pname$ %stability 
across decomposition}\label{sec:dllm_ablation_sec}

%\StQ{Inconsistent naming of StratQA: Table III uses StratQA while the text uses StratQA. Pick one convention or define de abbreviation.}

%\emph{To %address concerns regarding 
%test the consistency of the factual decomposition step and potential over-reliance on a single DLLM,} 
\emph{To assess the %methodological 
stability of $\pname$ across decomposition and rule out dependence on a single DLLM},
we conducted an ablation study using multiple different DLLMs. We compared our primary DLLM (GPT 4o~\cite{openai2024gpt4o}) with two open-weight, frontier-class LLMs (Mistral-Large-2512~\cite{mistral2025introducing} and GPT-OSS-120b~\cite{openai2025gptoss120bgptoss20bmodel}) and one commercial frontier-class LLM (Google Gemini 3 Flash~\cite{gemini2025family}). One of the authors manually evaluated 50 randomly sampled reasoning graphs (seed $42$) for %GPQA, MedQA, and StratQA 
each of the dataset ($\cf$, \Cref{sec:exp_setup})
%on three criteria: Atomicity (\emph{ability of the DLLM to isolate independent claims}), Absence of Hallucinations, and Dependency Validity (\emph{accurate preservation and mapping of causal topological edges}). 
for \emph{Atomicity} (measures the fraction of facts correctly
decomposed into independent unit), \emph{No Hallucination} (measures the absence of invented facts) and \emph{Dependency Validity} (measures the correctness of preserved causal edges). 
We then evaluated this same subset using Gemini 3 Pro~\cite{gemini2025family} as LLM-judge, which achieved a strong agreement ($\kappa = 0.82$~\cite{cohen1960kappa}) with the human evaluation. %(we used a different LLM than the DLLMs under study to avoid self-evaluation biases). 
We then extended the automated evaluation to $1{,}500$ randomly sampled reasoning graphs. %across GPQA, MedQA, and StategyQA (seed $42$). 
Because the goal is to assess decomposition quality rather than dataset-level task performance, \Cref{tab:dllm_evaluation} reports the averaged results across datasets. %We observe that the pipeline is robust and sufficiently decomposer-agnostic. The absence of Hallucinations is consistently greater than $95$\% across the evaluated DLLMs. Atomicity consistently exceeds $88$\%. This confirms that the evaluated DLLMs are suitable to break reasoning traces into atomic, independent facts. The Dependency Validity scores are lower because occasionally, DLLMs tend to skip transitive logical edges in complex answers. Furthermore, fixing the GLLM  to Llama-3.1 8B, 
All DLLMs achieve over $88\%$ atomicity and $95\%$ absence of hallucinations,
while dependency validity ranges from $68.8\%$ to $77.0\%$.
%\Cref{fig:dllm_agreement_pearson} reports the Pearson Correlation Coefficient (PCC) between the %final 
%GRCS score assigned using the baseline GPT-4o DLLM and the GRCS scores assigned by the three additional DLLMs.  
\Cref{fig:dllm_agreement_pearson} further compares the $\mathcal{U}_{\mathrm{GRCS}}$ %scores 
estimates produced by %each 
the same 
alternative DLLM with those obtained using GPT-4o, fixing the GLLM
to Llama~3.1~8B.
%These high correlations \StQ{entail - suggest} that the pipeline is stable across different DLLMs. 
Correlations range from $0.66$ to $0.94$ (average: $\mathbf{0.80}$), showing
that $\mathcal{U}_{\mathrm{GRCS}}$ remains stable across DLLMs.
The lowest agreement occurs for GPT-OSS-120B on StratQA (PCC = $0.66$), indicating %that StratQA 
it is the most challenging dataset for atomic decomposition. Its open-domain commonsense questions rely on informal reasoning that is harder to decompose into clean Subject-Verb-Object atomic facts than those in BoolQ,  MedQA, %and 
or GPQA, which are more structured. BoolQ and MedQA consistently yields the highest correlations (up to $0.94$, average: $0.94$ for BoolQ; $0.89$ for MedQA), likely because reading comprehension and clinical reasoning follow a %highly 
more structured format ($\eg$, symptoms, diagnosis, treatment for the latter) that %systematically 
more easily maps onto atomic Subject-Verb-Object facts with clear causal dependencies. %In practice, 
We find that more capable frontier-class DLLMs tend to yield %cleaner decompositions and 
more reliable graph decompositions. \textbf{The results show that %the pipeline 
$\pname$ is robust to the choice of DLLM.}

%\StQ{Table IV’s caption uses column names like Atomicity, No Hallucination, and Dependency Validity without a hint of what they mean. Add a short explanation in the caption: Atomicity measures the fraction of facts correctly decomposed into independent units; No Hallucination measures the absence of invented facts; Dependency Validity measures the correctness of preserved causal edges.}

\begin{table}%[t]
    \centering
    \caption{Evaluation of DLLMs for graph decomposition. \emph{Atomicity} measures the fraction of facts correctly decomposed into independent units; \emph{No Hallucination} measures the absence of invented facts; \emph{Dependency Validity} measures the correctness of preserved causal dependencies.}
    \label{tab:dllm_evaluation}
    \resizebox{\columnwidth}{!}{%
    \begin{tabular}{lccc}
        \toprule
        \textbf{DLLM} &
        \makecell{\textbf{Atomicity}} &
        \makecell{\textbf{No}\\\textbf{Hallucination}} &
        \makecell{\textbf{Dependency}\\\textbf{Validity}} \\
        \midrule
        GPT-4o             & 94.2\% & 95.6\% & 74.1\% \\
        Mistral-Large-2512 & 97.8\% & 95.8\% & 77.0\% \\
        GPT-OSS-120b       & 88.0\% & 97.5\% & 73.5\% \\
        Gemini 3 Flash     & 98.7\% & 96.0\% & 68.8\% \\
        \bottomrule
    \end{tabular}%
    }
\end{table}

\definecolor{cMistral}{HTML}{3182bd}
\definecolor{cGPTOSS}{HTML}{de2d26}
\definecolor{cGemini}{HTML}{31a354}
 
\begin{figure}
\centering
\includegraphics[width=0.8\linewidth]{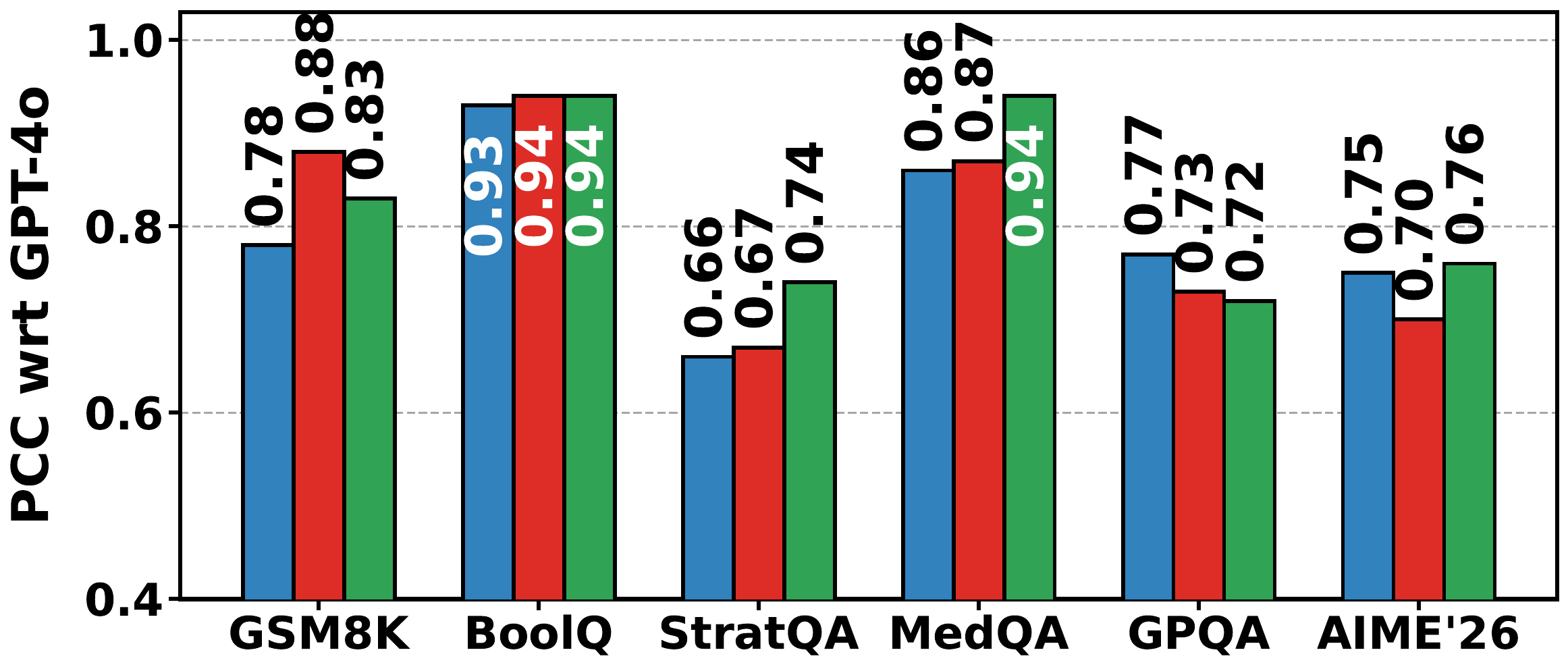}\\[-1.2mm]
\resizebox{\linewidth}{!}{%
\begin{tikzpicture}[baseline, every node/.style={font=\footnotesize\bfseries}]
    \node[draw=black, line width=0.6pt, fill=cMistral,
          minimum width=0.9em, minimum height=0.9em] (s1) at (0,0) {};
    \node[right=0.3em of s1, anchor=west] (l1) {Mistral-Large-2512};
 
    \node[draw=black, line width=0.6pt, fill=cGPTOSS,
          minimum width=0.9em, minimum height=0.9em,
          right=1.2em of l1.east, anchor=west] (s2) {};
    \node[right=0.3em of s2, anchor=west] (l2) {GPT-OSS-120B};
 
    \node[draw=black, line width=0.6pt, fill=cGemini,
          minimum width=0.9em, minimum height=0.9em,
          right=1.2em of l2.east, anchor=west] (s3) {};
    \node[right=0.3em of s3, anchor=west] (l3) {Gemini 3 Flash};
\end{tikzpicture}%
}
\caption{
  \label{fig:dllm_agreement_pearson}
  \textbf{Robustness of $\mathcal{U}_{\mathrm{GRCS}}$ across different DLLMs}. PCC between $\mathcal{U}_{\mathrm{GRCS}}$ scores obtained using GPT-4o and three alternative DLLMs. High agreement across datasets indicates that the proposed reasoning-graph construction pipeline yields stable uncertainty estimates regardless of the chosen DLLM.
}
%\vspace{-6mm}
\end{figure}

\section{Limitations}
\label{sec:limitations}

We should consider the following limitations when interpreting our results.
For UQ, $\mathcal{U}_{\mathrm{GRCS}}$ computation has a worst-case time complexity of $\mathcal{O}(N^2 \cdot m^3)$, where $N$ is the number of sampled CoTs and $m$ is the maximum node or edge assignment size. This matches Topo-UQ's complexity~\cite{da2025understanding}.
Under the CoTA formulation introduced by Tanneru et al.~\cite{tanneru2024quantifying}, we derive that %the CoTA implementation 
it requires $\mathcal{O}(N^2s^2T_{\mathrm{CE}})$ time, where $s$ is the maximum number of reasoning steps and $T_{\mathrm{CE}}$ is the cost of one entailment cross-encoder inference.
Thus, all evaluated UQ metrics scale at least quadratically with the number of sampled CoTs.
Despite this asymptotic scalability limitation, under our evaluated settings, $\mathcal{U}_{\mathrm{GRCS}}$ is empirically faster, consumes less energy, and requires lower peak GPU utilization than both baselines (\Cref{sec:GRCS_comp_cost}).
%w%hereas CoTA requires  than SC ($\mathcal{O}(N)$), since SC requires no graph 
%construction and it's just a majority voting. However, compared to Topo-UQ and CoTA, $\pname$ is empirically the fastest and most energy-efficient 
%(\cf, \Cref{sec:GRCS_comp_cost}).
For decoding, GSC inherits the pipeline's worst-case time complexity of $\mathcal{O}(N^2 \cdot m^3)$, whereas SC, based on majority-voting, requires $\mathcal{O}(N)$. GSC, therefore, trades a higher worst-case computational cost than SC for a more trustworthy answer. % selection.
%Also, $\pname$'s strength does not scale monotonically with model size. GRCS 
%correlations for DeepSeek~R1 are weaker than Llama 3.1 8B's on most datasets, 
%yet strongest in the entire study on AIME'26 (-$0.542$). %, and 
%Medoid ablation 
%shows DeepSeek R1 with some of the largest faithfulness drops on GPQA and 
%AIME'26. Conversely, Llama 3.1 8B's correlations turn positive on GPQA and 
%AIME'26, where the task exceeds its competence and it defaults to repetitive 
%hallucinated shortcuts. $\pname$ is thus most informative when a task stresses a GLLM's reasoning without exceeding it 
%entirely. This is reached by smaller GLLMs on moderately hard tasks ($\eg$, MedQA) and by more capable models on the hardest settings ($\eg$, AIME'26).
%Finally, by limiting reasoning to at most $n_h = 6$ steps, we primarily focus on short-horizon benchmarks. Its scalability to long-context, highly compositional tasks remains an open question for future work.
%\rreva{remove this after you write the final experiment on longer steps! This should not be a limitations, because we address this in our final experiment, still in writing}
%Finally, we cap CoTs at $n_h = 6$ steps regardless of task 
%complexity. This does not prevent evaluation on long-horizon tasks 
%(\eg, AIME'26), but it may truncate %genuinely 
%longer reasoning chains. 
%Preliminary runs with higher $n_h \in \{10;20\}$ show $\pname$ scales accordingly, but a full 
%evaluation is left to future work.

Also, the current evaluation focuses on textual reasoning tasks. Preliminary tests on the visual ARC-AGI benchmark~\cite{arc_agi_chollet} yielded $0\%$ accuracy with all tested GLLMs, and undefined faithfulness scores. This prevented meaningful evaluation of $\pname$. Future work will adapt the framework to multimodal reasoning by incorporating visual information into graph construction and faithfulness evaluation.
\section{Threats to Validity}\label{sec:threats}

\noindent \hypertarget{sec:external_threats}{\textbf{External Threats}.} Our evaluation cannot cover all available GLLMs or reasoning tasks. %, %and all possible reasoning tasks. 
%To mitigate this, we select 
We mitigate this by selecting GLLMs spanning general-purpose, reasoning-specialized, and distilled RL-enhanced paradigms, with substantially different model sizes. We evaluate them across six benchmarks covering arithmetic, reading comprehension, open-domain reasoning, medicine, graduate-level science, and competition mathematics. Nevertheless, evaluating across additional GLLM types and benchmarks would further strengthen our claims.

\smallskip

\noindent \hypertarget{sec:internal_threats}{\textbf{Internal Threats}.}
The causal dependencies are directly stated by the GLLM
itself and may therefore be incorrect. We intentionally preserve %these
hallucinated 
dependencies, %rather than correcting them with the DLLM, 
since they are part
of the reasoning behavior that $\pname$ aims to evaluate. However, under full mode-collapse, these inconsistencies are not guaranteed to be detected ($\cf$, \Cref{sec:grcs}).

\smallskip

\noindent \hypertarget{sec:construction_threats}{\textbf{Construct Threats}.} A potential construct threat is whether reasoning faithfulness is captured
consistently by the chosen evaluation procedure. To mitigate this, we evaluate
faithfulness using two complementary techniques, Early Answering (EA) and
Counterfactual Editing (CFE), and verify their agreement across configurations. We further assess the reasoning selected by GSC through Deductive Soundness,
%(DS), 
which independently evaluates factual correctness and logical entailment. % of the GSC-selected medoid.

\section{Related Work
%\dpal{START HERE}
}\label{sec_rel_work}

%\rreva{JOIN THIS + REL WORK OF APPENDIX + REL WORK OF THESIS}
%In recent years, LLMs have been used in many disparate tasks. As a result, UQ has become an essential field for understanding and deploying trustworthy models, especially in safety-critical applications. 
LLMs are being deployed across a wide range of mission-critical settings, including medicine~\cite{Shen2023ChatGPTDoubleEdged}, law~\cite{Linna2026ChallengesLegalReasoning}, and financial markets~\cite{Chen2025TortLiabilityAI}.
%As LLMs are deployed across diverse tasks, UQ has become essential for trustworthy use, especially in safety-critical settings.
Therefore, UQ has become an essential step to assess LLMs' trustworthiness prior to their deployment~\cite{tanneru2024quantifying}. We can divide existing UQ approaches for LLMs into two primary categories, \emph{architecture-dependent} and \emph{architecture-agnostic}. Architecture-dependent approaches require direct access to the internal mechanisms of the model to estimate uncertainty and can involve training auxiliary classifiers on these internal representations or fine-tuning the LLM itself to explicitly output confidence scores alongside its predictions~\cite{chen2024inside,
zou2025uncertainty}. Architecture-agnostic approaches treat the LLM as a closed system, and estimate uncertainty relying solely on the final generated outputs~\cite{tanneru2024quantifying,da2025understanding}. $\pname$ belongs to the architecture-agnostic category, as it requires neither internal model access nor additional training.

%\subsection{Architecture-dependent approaches}\label{sec:rel_work_a_d}

\smallskip

\noindent \hypertarget{sec:rel-work-a-d}{\textbf{Architecture-dependent approaches}.}
Within architecture-dependent approaches, Kadavath et al.~\cite{kadavath2022language} train a  classifier on the model’s internal representations to estimate the probability that the model will produce the correct answer across repeated open-ended samples. Chen et al.~\cite{chen2024inside} introduce \emph{INSIDE} to detect hallucinations by measuring the dispersion of hidden-state representations across sampled responses. Zou et al.~\cite{zou2025uncertainty} modify a LLM with a multi-hypothesis prediction head that jointly estimates medical-image grounding boxes and their uncertainty. %\bem{In contrast, $\pname$ %models sampled CoTs as reasoning graphs and 
%measures uncertainty from the semantic and topological consistency of reasoning, rather than over hidden representations or task-specific predictions, %Unlike task-specific approaches~\cite{zou2025uncertainty}, $\pname$ 
%and can be effective on a broader range of tasks, such as mathematical, medical, and graduate-level scientific reasoning.}
In contrast, $\pname$ %measures uncertainty from semantic and topological reasoning consistency and generalizes across mathematical, medical, and scientific reasoning tasks.}
provides a unified framework for UQ, faithfulness assessment, and decoding for mathematical, medical, and scientific reasoning. %tasks.

%\subsection{Architecture-agnostic approaches}\label{sec:rel_work_a_a}

\smallskip

\noindent \hypertarget{sec:rel-work-a-a}{\textbf{Architecture-agnostic approaches}} Within architecture-agnostic approaches, Kunh et al. introduce \emph{Semantic Entropy}~\cite{KuhnGalFarquhar2023}, which estimates semantic likelihoods and whether different outputs express the same answer.
Farquhar et al.~\cite{Farquhar2024SemanticEntropy} build on this concept and show that semantic entropy is an effective tool for \emph{hallucination detection}, by clustering multiple generations according to meaning and measuring uncertainty over those semantic clusters. %, they identify cases where the model is likely to confabulate
Lin et al.~\cite{LinTrivediSun2024} %studies output uncertainty with graph Laplacian eigenvalues, 
use spectral properties of semantic similarity graphs built from sampled responses. However, all these approaches quantify uncertainty at the level of the final answer, discarding the reasoning topology leading to each answer. In contrast, $\pname$ %instead 
estimates uncertainty from the semantic and structural relationships among intermediate reasoning steps.
%Considering works 
%Research specifically studying reasoning frameworks, 
%\cite{Yao2023TreeOfThoughts} proposes \emph{Tree of Thoughts}, enabling search over multiple reasoning branches through lookahead and backtracking, while~\cite{BestaEtAl2024} argues that graph structures can express richer dependencies among intermediate thoughts than linear chains or trees. \bem{However, these structures are designed to guide reasoning at inference time, not to model sampled CoTs, quantify uncertainty across them, or use that structure for decoding.}

% \subsection{Architecture-agnostic approaches focused on reasoning}\label{sec:rel_work_a_a_reasoning}

\smallskip

\noindent \hypertarget{sec:rel-work-a-a-reasoning}{\textbf{Architecture-agnostic approaches focused on reasoning}.} To move beyond this answer-level limitation, more recent architecture-agnostic approaches %have begun to incorporate intermediate reasoning steps.
focus on reasoning steps.
%Among %reasoning frameworks
%them, 
%\emph{Tree of Thoughts}~\cite{Yao2023TreeOfThoughts} searches over multiple reasoning branches, while graph-based reasoning can capture richer dependencies than chains or trees~\cite{BestaEtAl2024}. 
\emph{Tree of Thoughts}~\cite{Yao2023TreeOfThoughts} organizes candidate intermediate thoughts into a search tree, allowing the LLM to evaluate alternative reasoning paths, look ahead, and backtrack from unpromising branches. \emph{Graph of Thoughts}~\cite{BestaEtAl2024} generalizes this structure to a %directed 
graph, where nodes are thoughts and edges are dependencies. This enables reasoning steps to be combined, refined, and reused during inference. %\bem{However, these methods guide inference rather than model sampled CoTs, quantify uncertainty across them, or decode from their structure.
%However, these methods are designed to guide inference, not to model sampled CoTs, quantify uncertainty across them, or decode from their structure.}
%\StQ{To reduce the likelihood that a reviewer says: This looks similar to Graph of Thoughts, I would add the following. Unlike Graph of Thoughts, which represents reasoning as a graph during inference to improve problem solving, GRAPHEVAL constructs graphs after sampling multiple reasoning traces to quantify uncertainty, assess reasoning faithfulness, and perform topology-aware decoding. Thus, our objective is uncertainty estimation and reasoning evaluation rather than search.}
Unlike \emph{Graph of Thoughts}, which represents reasoning as a graph during inference to improve problem solving, $\pname$ constructs graphs %after sampling multiple reasoning traces 
to quantify uncertainty, assess reasoning faithfulness, and perform %topology-aware 
decoding. Thus, our objective is uncertainty estimation and reasoning evaluation rather than search.

%Refer to Appendix~\ref{sec:rel_work_appendix} for a more detailed related work discussion.
\section{Conclusion}
\label{sec:conclusion}

%We argue that LLM trustworthiness cannot be judged from answer agreement alone. By modeling sampled CoTs as DAGs, we show that reasoning structure carries useful signals about uncertainty, faithfulness, and robustness. Our experiments indicate that GRCS is a strong uncertainty indicator, GSC is a more faithful decoding rule than SC, and the selected medoid often acts as a load-bearing reasoning path. These results highlight the value of moving from answer-level consensus to reasoning-level analysis in mission-critical settings.
%We argue that LLM trustworthiness cannot be judged by answer agreement alone. By modeling sampled CoTs as DAGs, we show that reasoning structure reveals signals of uncertainty, faithfulness, and robustness.
We argue that LLM trustworthiness requires more than answer agreement: reasoning structure reveals uncertainty, faithfulness, and robustness.
%GRCS emerges as a strong uncertainty metric, GSC as a more faithful alternative to SC, and the medoid as a load-bearing reasoning path. %hese findings motivate a shift from answer-level consensus to reasoning-level analysis in mission-critical settings
$\mathcal{U}_{\mathrm{GRCS}}$ outperforms baselines in 77.8\% of cases, GSC selects more logically sound answers in 24.9\% more disagreements, and GSC-medoid ablation reduces faithfulness in up to 55\% of answers.
These findings motivate a shift to reasoning-level analysis in mission-critical settings.

\bibliographystyle{IEEEtran}
\bibliography{bib/master,bib/reference_converted}

%\vskip -4\baselineskip

\begin{IEEEbiography}
[{\includegraphics[width=1in,height=1.25in,clip,keepaspectratio]{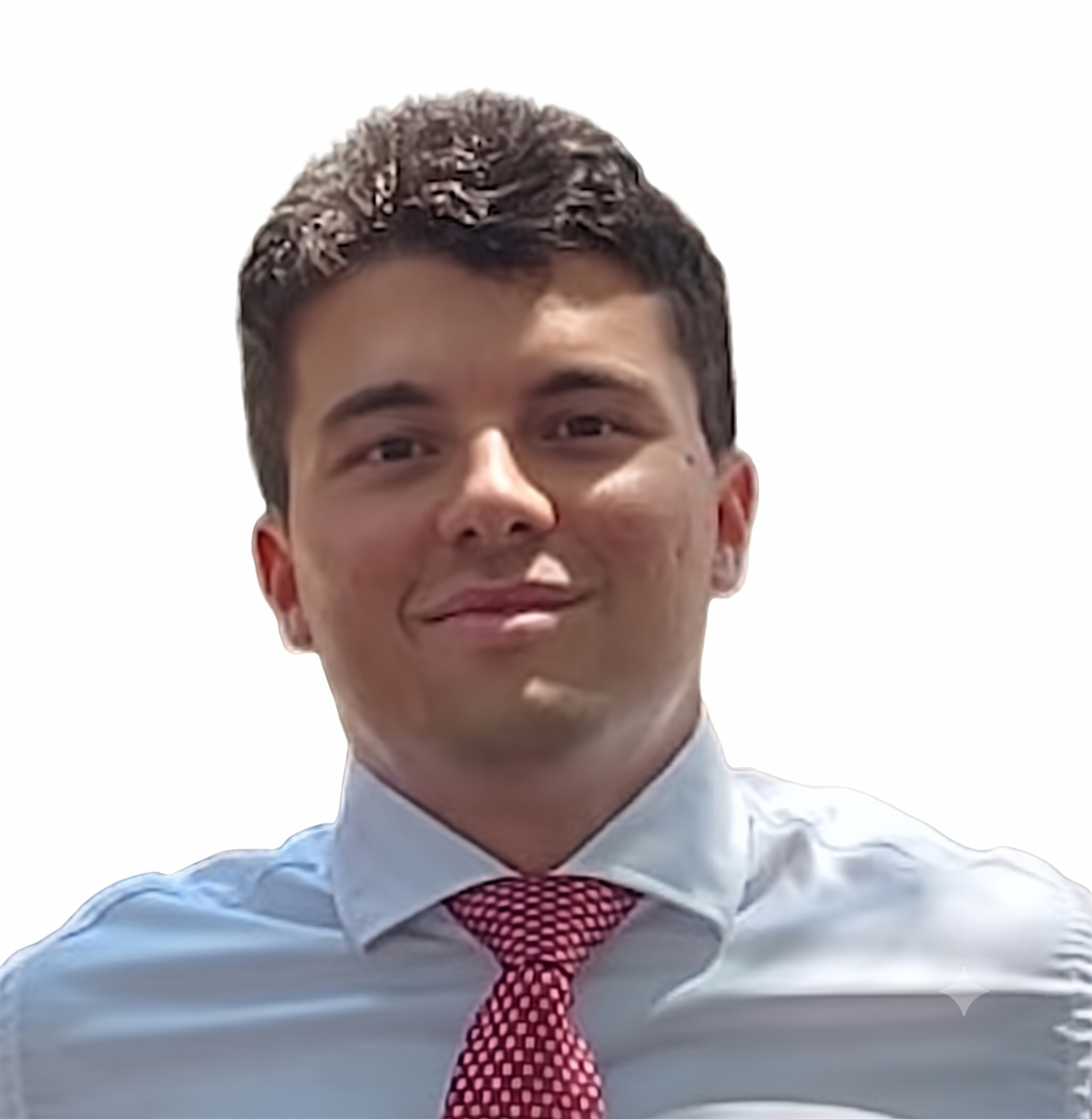}}]{Riccardo Revalor} is a Ph.D. student %from the University of Illinois Chicago. 
at the Electrical and Computer Engineering Department at the University of Illinois, Chicago. %He received a M.S. degree from the University of Illinois Chicago, a M.S. degree from Politecnico di Torino, and a B.S. degree from Politecnico di Torino. 
He received an M.S. degree in Computer Science from the University of Illinois, Chicago, and both M.S. and B.S. degrees in Computer Engineering from Politecnico di Torino, Italy.
His current research interests include neuromorphic and bio-inspired AI, AI SW/HW Codesign, AI-assisted test generation and debugging of hardware designs, and LLM uncertainty estimation.
\end{IEEEbiography}

\begin{IEEEbiography}[{\includegraphics[width=1in,height=1.25in,clip,keepaspectratio]{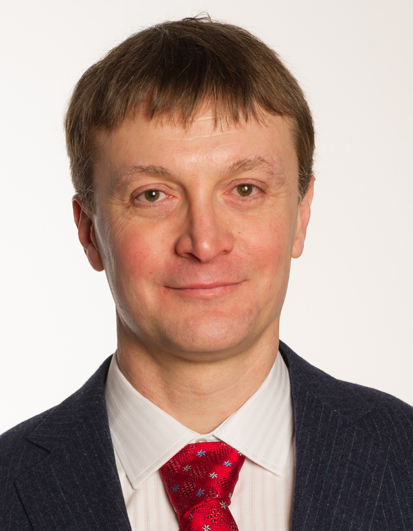}}]{Stefano Quer} received the M.S. degree in Electronic Engineering in 1991 and the Ph.D. degree in Computer Engineering in 1996. He is currently a Professor with the Department of Control and Computer Engineering at Politecnico di Torino, Italy. During his career, he has been a Visiting Faculty member at the University of California, Berkeley, and has gained industrial research experience at Digital Equipment Corporation, Intel Corporation, Synopsys Inc., and Compaq Computer Corporation, where he also served as a consultant. His research interests include computer-aided design for VLSI systems, formal methods and testing for hardware and software verification, and sequential and concurrent algorithms for solving complex problems under limited computational resources.
\end{IEEEbiography}

\begin{IEEEbiography}[{\includegraphics[width=1in,height=1.25in,clip,keepaspectratio]{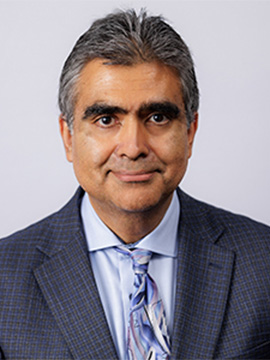}}]{Jalees Rehman} received an M.D. from the Technical University of Munich in 1991, Munich, Germany, a Dr.med from Philipps-Universit\"{a}t Marburg, Marburg, Germany in 2004, and a Fellowship in Cardiology at Indiana University in 2006. He is currently the Benjamin J. Goldberg Professor and Head in the Department of Biochemistry and Molecular Genetics at the University of Illinois Chicago. His current research interests include application of causal reasoning, counterfactual reasoning in mission critical and safety critical medical applications, computational genomics, and bioinformatics. 
\end{IEEEbiography}

\begin{IEEEbiography}[{\includegraphics[width=1in,height=1.25in,clip,keepaspectratio]{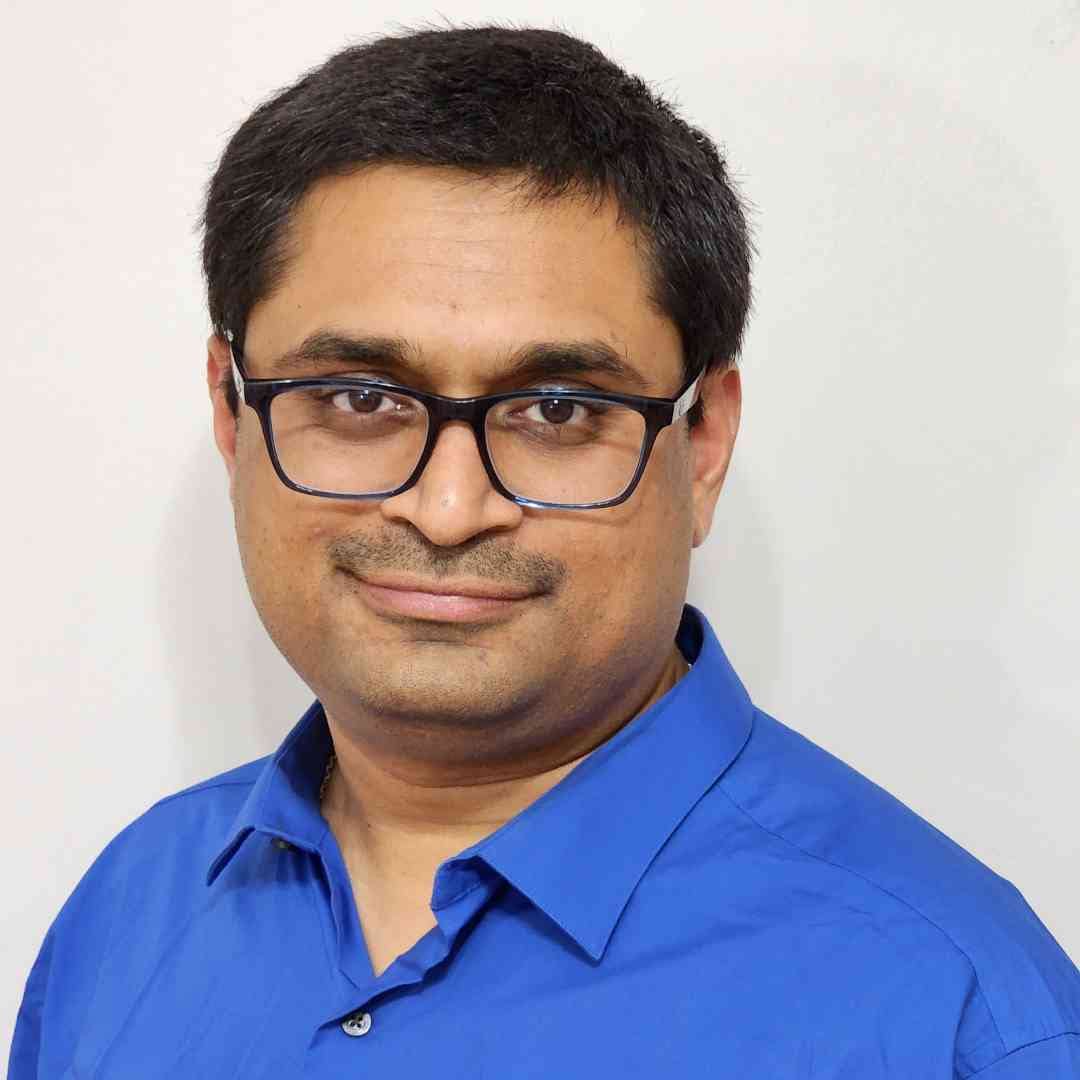}}]{Debjit Pal} received a Ph.D. from the University of Illinois at Urbana-Champaign. He received B. Tech and M.S. degrees from Jadavpur University and IIT Kharagpur, India, respectively. He is currently an assistant professor in the Department of Electrical and Computer Engineering at the University of Illinois, Chicago. His current research interests include post-silicon and pre-silicon validation and debug, analog design automation, and programming paradigms for hardware accelerators. He received best paper nominations in ICCAD 2015, DAC 2018, ASP-DAC 2019, and FPGA 2024. He is the winner of the IEEE CEDA System Validation and Debug Technology Committee (SVDTC) Student Research Award 2016.
\end{IEEEbiography}

\end{document}